\documentclass[11pt]{article}
\usepackage[preprint]{acl}

\usepackage{times}
\usepackage{latexsym}
\usepackage{amssymb}
\usepackage{amsmath}
\usepackage[T1]{fontenc}
\usepackage{enumitem} 
\usepackage{booktabs}
\usepackage{tabularx} 
\usepackage{supertabular} %
\usepackage{graphicx}
\usepackage{microtype}
\usepackage{enumitem}
\usepackage{inconsolata}
\usepackage{multirow}
\usepackage{multicol}
\usepackage{tcolorbox}
\usepackage{longtable}
\usepackage{graphicx}
\usepackage{parskip}
\usepackage{float}
\usepackage[table]{xcolor}
\usepackage{colortbl}
\usepackage[utf8]{inputenc}
\usepackage{fvextra}

\DefineVerbatimEnvironment{PromptBlock}{Verbatim}{
  fontsize=\scriptsize,
  breaklines=true,
  breakanywhere=true,
  frame=single,
  framesep=3pt
}

\title{RUMBA: Russian User Memory Benchmark}

\author{
 \textbf{Elizaveta Shevtsova\textsuperscript{1}},
 \textbf{Inna Glebkina\textsuperscript{1}},
 \textbf{Mark Baushenko\textsuperscript{1}},\\
 \textbf{Pavel Gulyaev\textsuperscript{1}},
 \textbf{Alena Fenogenova\textsuperscript{1}}
\\
 \textsuperscript{1}DAIMLD
\\
 \small{
   \textbf{Correspondence:} \href{mailto:lisabeth.shevtsova@gmail.com}{lisabeth.shevtsova@gmail.com}
 }
}

\begin{document}
\maketitle
\begin{abstract}

The ability to handle long-term memory in LLMs is becoming increasingly critical, yet existing benchmarks remain English‑centric and rely on aggregate retrieval metrics, failing to capture interactions between long‑range context, temporal information, and reasoning. To address this, we introduce \textbf{RUMBA} (Russian User Memory BenchmArk) — a new benchmark for long‑term conversational memory that provides a fine‑grained taxonomy of memory‑centric question types and a unified methodology accounting for semantic type, session scope, temporal reasoning, and the explicitness of temporal expressions. RUMBA consists of timestamped user–assistant dialogues with QA pairs requiring retrieval, combination, and reasoning across sessions. While designed for Russian, we also provide an aligned English subset under the same methodology. We evaluate contemporary memory systems and long‑context models, and show how RUMBA serves as a diagnostic tool to analyze model behavior across benchmark slices and identify strengths and failure modes of different memory mechanisms.

\end{abstract}

\section{Introduction}
\label{sec:intro}

Large language models (LLMs) are increasingly deployed as conversational assistants that must retain user-specific information across sessions and remain consistent over time. This requires more than simple retrieval: models must preserve salient details, resolve long-range references, and reason over temporal updates and multi-step dependencies.

Recent English-centric memory benchmarks ~\cite{maharana2024locomo, wu2024longmemeval, bei2026memgallery} have advanced long-term memory evaluation. However, recent work on memory-agent evaluation notes that existing memory
benchmarks are still largely recall- or retrieval-oriented and provide
incomplete coverage of memory behaviours required in realistic long-horizon interactions, including memory updates and consolidation, stale or obsolete information, and forgetting~\cite{du2026memory,hu2025memoryagentbench,uddin2026memora}.
As summarizes in Table~\ref{tab:benchmark_comparison}, existing benchmarks
also rely on synthetic dialogue construction and offer limited
diagnostic granularity through relatively coarse question categories. This
makes it difficult to isolate failure factors such as temporal dependencies, conversational span, or query semantics.

Meanwhile, Russian benchmarks such as MERA~\cite{fenogenova2024mera} and LIBRA~\cite{churin2024libra} evaluate broader model capabilities, including general language understanding and long-context processing, but not conversational memory in a targeted way. This leaves a clear need for a benchmark focused on memory, temporal awareness, and reasoning in long Russian dialogues.

To address this gap, we introduce \textbf{RUMBA} (\textbf{R}ussian \textbf{U}ser \textbf{M}emory \textbf{B}enchm\textbf{a}rk), a benchmark for evaluating long-term conversational memory in Russian. RUMBA is built around long-form dialogue scenarios and a taxonomy of question types covering contextual recall, entity tracking, temporal reasoning, and multi-step reasoning. We provide two test sets (Russian and an aligned English translation) and conduct evaluations under both RAG and full-context scenarios.

Our contributions are as follows:
\begin{itemize}[nosep]
    \item We create and release \textbf{RUMBA} as an open-source benchmark\footnote{\url{https://huggingface.co/datasets/ai-forever/RUMBA}}, the first benchmark specifically designed to evaluate long-term conversational memory in Russian, accompanied by an aligned English translation to enable cross-lingual diagnostics and reproducibility.
    
    \item We present a fine-grained taxonomy of memory-centric question types along multiple axes, enabling detailed model comparison beyond aggregate accuracy and simple retrieval.
    
    \item We implement a unified evaluation pipeline that supports both retrieval-based memory systems and full-context baselines for both languages, which standardizes the evaluation workflow while allowing each method family to operate under its natural memory-access setting\footnote{\url{https://github.com/ai-forever/RUMBA}}.

    \item We conduct an extensive empirical analysis of different families of memory solutions and a single-model case study, demonstrating RUMBA as a diagnostic tool for identifying strengths and failure modes across benchmark slices.
\end{itemize}

We hope that RUMBA will serve both as a diagnostic benchmark for evaluating memory in Russian-language conversational systems and as a practical resource for developing models that can reliably store, update, and use user-specific information over time.

\begin{figure*}[!ht]
    \centering
    \includegraphics[width=\linewidth]{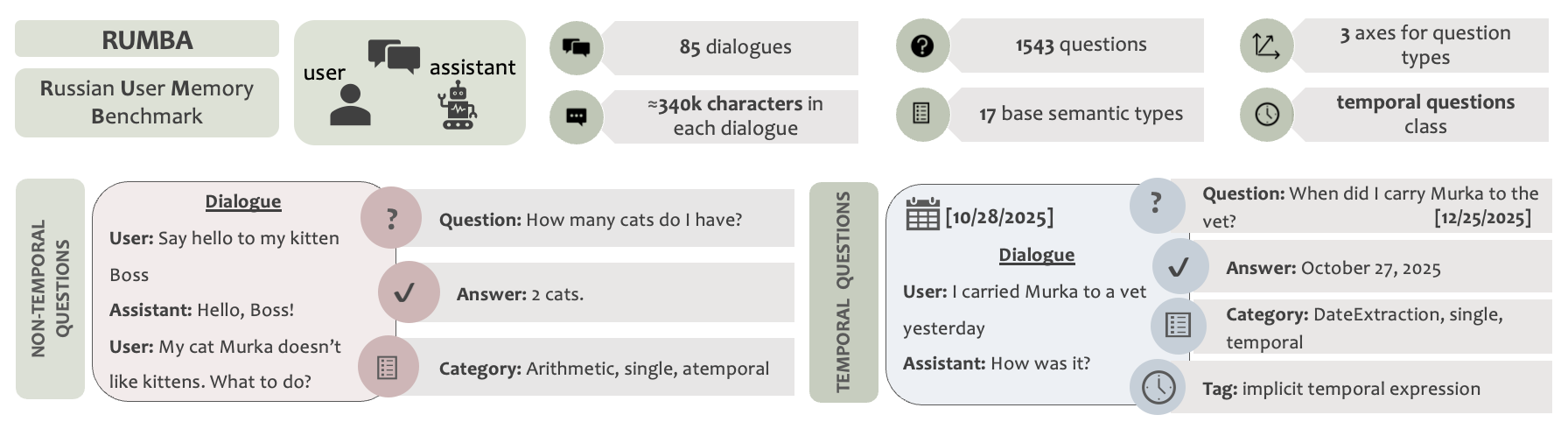}
    \caption{\small Overview of the proposed benchmark and annotation framework. The figure summarizes dataset-level statistics, representative user--assistant dialogues, example question--answer instances, and the multi-dimensional annotation scheme used to categorize memory questions.}
    \label{fig:benchmark_overview}
\end{figure*}

\section{Related work}
\label{sec:related}



Recent work on evaluating memory capabilities in language models has been shaped by benchmarks such as LoCoMo~\cite{maharana2024locomo},  LongMemEval~\cite{wu2024longmemeval}, Mem-Gallery~\cite{bei2026memgallery}. These datasets provide a structured starting point for studying long-term memory in conversational systems. In particular, LoCoMo organizes evaluation around reasoning-oriented question families (e.g., single-hop, multi-hop, temporal, commonsense, adversarial), while LongMemEval defines a set of core memory abilities (e.g., recall, personalization, multi-session reasoning, updates, temporal reasoning, abstention) and operationalizes them through specific question types. Together, these benchmarks establish a useful conceptual and methodological baseline for memory evaluation and, particularly, for the RUMBA taxonomy.

However, they have three key limitations (Table~\ref{tab:benchmark_comparison} shows the overview): (1) flat taxonomies that entangle reasoning complexity, temporal structure, and memory access within single question categories; (2) treating temporal reasoning and knowledge updates as isolated task types rather than dimensions that can interact; and (3) ignoring memory control behaviors such as intentional forgetting or validity constraints. 

Several benchmarks (e.g., TimeBench~\cite{chu-etal-2024-timebench}, temporal NLI ~\cite{vashishtha-etal-2020-temporal}, and the three‑dimensional temporal QA framework~\cite{piryani2026itshightimesurvey}) evaluate temporal reasoning in LLMs, covering event ordering, duration, and temporal commonsense. However, they are not designed for long‑context settings nor grounded in realistic user–assistant interactions, and thus fail to capture temporally dependent reasoning across extended multi‑session conversations.

The gap is even more pronounced for Russian: no dedicated benchmark for persistent long‑term memory exists. LIBRA~\cite{churin2024libra} evaluates long-context understanding, but does not test structured memory usage over time. MERA~\cite{fenogenova2024mera} focuses on general reasoning and knowledge tasks, without isolating memory-specific abilities. The only partially relevant dataset, ruTiE~\cite{chervyakov2026multimodalevaluationrussianlanguagearchitectures}, includes conversational evaluation, but operates on short dialogues and does not capture persistent, multi-session memory phenomena.

To address all of the above limitations (flat taxonomies, lack of interactive temporal dimensions, absence of forgetting scenarios, inapplicability to long‑context dialogues, and the complete gap for Russian), we propose a benchmark for Russian with an extended taxonomy covering forgetting and broader reasoning types, multi‑axis annotations for fine‑grained diagnostics, and human‑written user turns to reduce synthetic interaction bias.

\begin{table*}[!htbp]
\centering
\tiny
\setlength{\tabcolsep}{3pt}
\begin{tabular*}{\textwidth}{@{}l@{\extracolsep{\fill}}cccc@{}}
\toprule
\textbf{Feature} & \textbf{LoCoMo} & \textbf{LongMemEval} & \textbf{Mem-Gallery} & \textbf{RUMBA} \\
\midrule
Modality & Text/Image & Text & Text/Image & Text \\
Language support & English only & English only & English only & \textbf{Russian + English} \\
Context length (chars) & 44k--90k & S: 455k--514k / M: 4.5M--5.2M & 42k--86k & RU: 297k--569k / EN: 234k--528k \\
\# Annotated Q-A pairs & 1,542 & 500 & 1,711 & 1,543 \\
Question taxonomy & Flat (5 types) & Flat (5 types) & Flat (3 types) & \textbf{Orthogonal axes ($17 \times 2 \times 2$ types)} \\
User utterances origin & Generated & Generated & Mixed &
\textbf{Human-written} \\
\midrule
Temporal reasoning as independent dimension & No & No & No & \textbf{Yes} (\checkmark) \\
Explicit support for "forgetting" & No & No & No & \textbf{Yes} (\checkmark) \\
Memory scenario coverage & retention, abstention & retention, abstention, updating & retention, abstention, updating & retention, abstention, updating, \textbf{forgetting} \\
\midrule
Date for each dialogue's utterances & Yes & Yes & Yes & Yes \\
Date for questions & No & Yes & No & Yes \\
\midrule
Failure diagnosis granularity & Limited & Limited & Limited & \textbf{Finer-grained (temporal vs. atemporal)} \\
\bottomrule
\end{tabular*}
\caption{\small Comparison of memory benchmarks: LoCoMo, LongMemEval, Mem-Gallery, and RUMBA. 
\textbf{Modality}: content types present in the dialogues; 
\textbf{Language support}: languages of the dialogues and questions; 
\textbf{Context length}: dialogue length measured in characters; 
\textbf{\# Annotated Q-A pairs}: number of question-answer pairs associated with each dialogue; 
\textbf{Question taxonomy}: structure of the question taxonomy, distinguishing \textit{Flat} taxonomies, where each question is assigned a single type, from \textit{Orthogonal} axes, where each question receives multiple tags; RUMBA uses semantic $\times$ quantitative $\times$ temporal axes; 
\textbf{User utterances}: the primary source of user-side dialogue utterances, distinguishing LLM-generated, mixed-source, and human-written data; RUMBA uses manually written user utterances.
\textbf{Temporal reasoning as an independent dimension}: whether one of the assigned tags indicates that a question is temporal or atemporal; 
\textbf{Explicit support for ``forgetting''}: whether the benchmark includes scenarios involving deletion requests, such as ``delete this'' or ``forget this''; 
\textbf{Memory scenario coverage}: high-level task categories used for memory evaluation; 
\textbf{Dates for dialogue utterances and questions}: whether each utterance and question is timestamped; 
\textbf{Failure diagnosis granularity}: the ability to identify which benchmark dimension contributes to question complexity.}
\label{tab:benchmark_comparison}
\end{table*}

\section{RUMBA Methodology}
\label{sec:methodology}

\subsection{Overview}

RUMBA was designed around a set of diagnostic research questions that directly address the limitations identified in prior work. The goal is not only to produce an aggregate score but to make model evaluation informative about specific memory capabilities and failure modes. Accordingly, dialogues, questions, answers, and evidence configurations were created to support controlled comparisons along the main dimensions of long‑term conversational memory. The benchmark supports the following research questions:

\begin{enumerate}[nosep]
    \item \textbf{Memory scope and semantics.} How does performance vary with session scope (single vs. multiple sessions) and with semantic supergroups (extraction, reasoning)?
    
    \item \textbf{Temporal reasoning.} How does performance differ between temporal and atemporal questions, and between cases where temporal information is explicit vs. implicit?
    
    \item \textbf{Retrieval vs. full-context} How do retrieval-augmented memory systems, which trade off perfect recall for scalability, compare with full-context baselines that preserve all information within a finite window but cannot scale indefinitely?

\end{enumerate}

Fig.\ref{fig:benchmark_overview} shows the overall structure of the proposed benchmark.

\subsection{Taxonomy} 
\label{subsec:taxonomy}

The RUMBA taxonomy is designed to address the limitations discussed in Sec.\ref{sec:related} by moving from a flat, benchmark-specific categorization to a multi-dimensional, compositional framework for memory evaluation. Instead of assigning each question to a single class, RUMBA decomposes tasks along several orthogonal axes, including:
 \begin{itemize}[nosep]
     \item the semantic operation required (e.g., recall, update tracking, comparison),
     \item the temporal dimension (temporal vs. atemporal),
     \item the number of sessions involved,
     \item the reasoning complexity,
     \item and the memory validity state (e.g., active vs. forgotten information).
 \end{itemize}

The questions are categorized according to three axes: (1) \textit{Semantic axis} represents the classification of questions with respect to their meaning and assessment focus. Broadly, the questions are classified into Information Extraction and Reasoning classes. (2) \textit{Quantitative axis} indicates the number of sessions required to answer a given question. Questions are divided into single-session and multi-session classes. (3) \textit{Axis of temporality} denotes whether a question is temporal. Questions are divided into temporal and atemporal classes. The summary of the axes and corresponding types of questions is presented in Table~\ref{tab:taxonomy_summary}.

This axis-based design enables a more precise characterization of model behavior. For example, the same underlying operation (e.g., recall) can be evaluated in both temporal and atemporal settings, revealing different failure modes. Similarly, reasoning is treated not as a separate category, but as a property that can be combined with different types of memory access, allowing finer-grained analysis of model capabilities.

A key strength of RUMBA is its treatment of \textit{temporality} as an independent dimension. Unlike prior benchmarks, where temporal reasoning is isolated as a specific task type, RUMBA allows temporality to interact with all other axes. This makes it possible to systematically study how models handle time-dependent versus time-independent memory queries within the same semantic framework.

Another important contribution is the explicit modeling of memory control and validity. In addition to standard abilities such as recall and update tracking, RUMBA introduces evaluation of intentional forgetting, where models must correctly exclude previously stored information after explicit user instructions. The forgetting scenario is supported through user behavior and a dedicated question type, DeleteInfo. Users may trigger it with requests such as “delete this” or “forget it.” The correct response is “There is no such information” (or similar), indicating that the information has been successfully forgotten and is no longer retained by the system.

This reflects realistic requirements for memory-enabled assistants and introduces a new class of failure modes not captured in prior work.
A detailed taxonomy is provided in the Appendix~\ref{sec:taxappendix}.

The taxonomy was designed with close consideration of real-world production user behavior in memory-enabled assistant systems. 
In particular, its structure was informed by observations from a deployed proprietary conversational assistant with long-term memory functionality, including naturally occurring user behaviors related to storing, updating, correcting, and deleting memories. 
Grounding the taxonomy in authentic interaction patterns allowed us to better capture the diversity and practical characteristics of real conversational memory queries beyond synthetic or narrowly constructed benchmark settings.

\begin{table}[htbp]
\centering
\renewcommand{\arraystretch}{1.3}
\tiny
\captionsetup{font=small}
\caption{RUMBA Taxonomy Summary. Example of question's full annotation: $Question$ = What do I like about my job? $Types$ = StaticUser, multi, atemporal. Full taxonomy description is available in Table~\ref{tab:full_description} of Appendix.}
\label{tab:taxonomy_summary}
\begin{tabular}{@{}p{2cm}p{5.1cm}@{}}
\toprule
\textbf{Axis} & \textbf{Categories / Types} \\
\midrule
Semantic & Extraction class (6 types): StaticUser, UpdatingInfo, DeleteInfo, AsstQuery, OpenDomainType, DateExtraction \\
& Reasoning class (10 types): SocialRelationship, Ordering, Arithmetic, Comparison, UserQA, CalendarUnderstanding, TemporalCommonsense, TemporallyModifiedGeneralReasoning, ComplexRelations, OtherReasoning \\
& Abstention type: no answer in dialogue \\
& Types with subtypes: OtherReasoning, Arithmetic, DateExtraction, CalendarUnderstanding, TemporalCommonsense, TemporallyModifiedGeneralReasoning, ComplexRelations \\
\midrule
Quantitative & Single-session, Multi-session \\
\midrule
Temporality & Atemporal, Temporal \\
& Temporal Expression Tags: explicit / implicit / no temporal expression \\
\bottomrule
\end{tabular}
\end{table}

\subsection{Dataset characteristics}
Table \ref{tab:dataset_features} summarized the main characteristics and statistics of the Russian version of the dataset. 
The instances that comprise the dataset represent textual dialogues between an assistant (LLM) and a user, as well as questions associated with each dialogue that relate to its content. 

\paragraph{Sessions.} Each dialogue consists of timestamped sessions, each representing one day of interaction with the assistant. We distinguish \textit{evidence sessions}, which contain information relevant to at least one question, from \textit{filler sessions}, which include both conversations about the user that are irrelevant to questions and sessions with no user-related factual information. Evidence sessions constitute 52.77\% of all sessions, micro-averaged across dialogues. This proportion does not alone characterize retrieval difficulty because most content in an evidence session may be irrelevant to a particular question.

\paragraph{Questions.} As described above, the question typology is based on three axes and covers different semantic question types. The dataset consists of both open-ended and closed-ended questions. The question-level timestamp is of equal importance to the session-level timestamp, given that the correct interpretation of the former is critical to generating an accurate response.

\paragraph{Dialogues Characters.} The dialogue characters represent diverse behavioral scenarios, which increases the realism of the user–assistant interactions and reflects authentic user behavior. The dialogues also incorporate sociocultural features characteristic of the Russian language and culture. Information about the characters and topics is presented in Subsection \ref{subsec:characters_app} of the Appendix.

\begin{table}[htbp]
\centering
\renewcommand{\arraystretch}{1.2}%
\tiny
\begin{tabular}{@{}p{3cm}p{4.1cm}@{}}
\toprule
\textbf{Feature} & \textbf{Value / Description} \\
\midrule
Total dialogues & 85 \\
Total questions & 1,543 \\
Average dialogue length & $\approx$ 340,000 characters \\
Utterances per dialogue & 180 -- 998 \\
Session definition & Single interaction with the assistant on a specific day \\
Sessions per dialogue & 12 -- 85 \\
Evidence sessions & 52.77\% of all sessions (micro-average) \\
Questions per dialogue & 14 -- 22 \\
Number of Question Types & 17 base types \\
Language & Contemporary standard Russian (user specific) \\
Dialogue specifics & Include language errors\\
& Reflect Russian sociolinguistic worldview \\ 
Offensive content & None (no offensive, insulting, or threatening data) \\
\midrule
\multicolumn{2}{l}{\textit{Note: Dialogues are independent of each other.}} \\
\bottomrule
\end{tabular}
\captionsetup{font=small}
\caption{\small RUMBA Dataset Features and statistics.}
\label{tab:dataset_features}
\end{table}

\begin{table}[t]
\centering
\resizebox{\columnwidth}{!}{%
\begin{tabular}{lccc}
\hline
\textbf{Language} & \textbf{Dialogue text} & \textbf{+ roles} & \textbf{+ answer prompt} \\
\hline
RU & 341{,}103.13 & 344{,}758 & 346{,}751 \\
EN & 320{,}782.08 & 324{,}437 & 326{,}430 \\
\hline
\end{tabular}%
}
\caption{\small Context lengths in the Russian and English versions of RUMBA, measured in characters.}
\label{tab:avg_chars_per_dialog}
\end{table}

\subsection{Dataset creation}
\label{subsec:data_creation}

\paragraph{Russian version.}

The dataset was developed in multiple stages, including taxonomy design and dialogue construction. Initially, information extraction questions — constituting the majority of the dataset — were created, subsequently, reasoning questions and a category of temporal questions were incorporated.

The dialogue creation process involved a total of 26 contributors, comprising staff members and crowd workers (ABC Elementary\footnote{\url{https://elementary.center}} and Rambler\&Co  companies\footnote{\url{https://rambler-co.ru}}). 
Each participant received monetary compensation according to the terms and conditions specified in their respective employment or contractual agreements. Information about the participants (gender, age, and field of expertise) is presented in Fig.\ref{fig:experts} of Appendix.

Data were collected over 20 working days. The process comprised two main stages: the creation of dialogues and associated questions by the authors, followed by a validation phase conducted by other contributors to ensure quality and consistency. More detailed information about the validation process can be found in Section~\ref{sec:annotation} of Appendix.

Before the start of data collection, all participants received comprehensive instructions (See Fig. \ref{fig:guideline_authors} in Appendix), were acquainted with the question taxonomy, and were provided with the matrix.

The dialogues were created in real time. User utterances were written by human participants, while GigaChat LLM was selected to generate the assistant’s responses, as it is specifically designed for the Russian language and Russian-speaking users, and is the only model pretrained on Russian data ~\cite{gigachatteam2025gigachatfamilyefficientrussian}. The dialogue authors used the web version of GigaChat Max dated July 1–31 and October 21–November 10, 2025.

\paragraph{English version.}

We additionally provide an aligned English diagnostic split. Given the large size of the dialogue data, controlled manual translation would be too resource-intensive. Therefore, the English version of the dataset was produced using automated LLM translation, followed by human verification of the translation quality. This follows common practice in machine translation evaluation and multilingual benchmark construction, where full human assessment is costly and machine-translated data is often validated or post-edited on selected subsets~\cite{lo2023data, singh2025globalmmlu, rajaee2025empirical}.

For the assessment of translation quality, 10\% of the translated dataset (i.e., 9 dialogues or a total of 185 sessions) was uniformly sampled and evaluated by human annotators. Translation quality was assessed using the \textit{Critical} and \textit{Task-specific} criteria defined in POLLUX~\cite{martynov2025eyejudgementdissectingevaluation}, with results summarized in Table~\ref{tab:translation_criteria}. All 3,543 translated pairs RU-utterance --- EN-utterance of the 9 dialogues were validated by domain-qualified specialists. Each evaluator compared the original Russian text with its English translation according to the POLLUX criteria, and every RU-EN pair was assessed by three annotators. The overall translation quality score was estimated at 0.88 (Table~\ref{tab:translation_quality}).

The next stage involved improving the quality of the translated set through a careful validation of the question–answer pairs and the corresponding evidence sessions. A total of 1,543 question–answer pairs (3,086 utterances in total) and 2,232 evidence sessions containing responses to the questions were each validated by three annotators. Where necessary, corrections were made to the English versions of the question–answer pairs and sessions. This helped preserve the logical and semantic coherence of the question, answer and evidence sessions to the same extent as in the original Russian version.
More details on translation and its validation are provided in Appendix \ref{sec:translation}.

\section{Evaluation}
\label{sec:evaluation}

\subsection{Baselines}
\label{subsec:baselines}

We evaluate two families of baselines that correspond to two common ways of using long interaction histories in LLM-based assistants: full-context inference and memory-based retrieval-augmented inference.

\paragraph{Full-context baselines.}
The first family represents long-context models that receive the complete dialogue history directly in the input context and generate the answer from this full evidence. This setting provides a classic upper-bound-style baseline for memory use. It also tests whether the model can locate and integrate relevant evidence in very long conversational contexts. In our experiments, we selected models with context windows around or above 1M tokens: fitting the largest dialogue requires context lengths roughly above 400K tokens, while the next practical tier among available long-context models is around 1M tokens. This constraint excluded Russian-oriented models, for which we did not find suitable context lengths, as well as available open-source models from the \texttt{Qwen} and \texttt{DeepSeek} families. Although \texttt{Qwen3.5-Flash} and \texttt{Qwen3.5-Plus} nominally satisfy the context-length requirement, provider-side content inspection rejected benchmark inputs with \texttt{DataInspectionFailed} errors in our setup, so we excluded them from the final evaluation. Prior long-term memory benchmarks similarly use long-context LLMs as a natural comparison point for evaluating implicit memory mechanisms~\cite{maharana2024locomo, wu2024longmemeval}. We therefore evaluated both proprietary and open-source long-context models. The proprietary models are \texttt{gpt-4.1-mini}, \texttt{gpt-5.4}, \texttt{gemini-3.1-flash-lite}, \texttt{grok-4.1-fast},  and \texttt{claude-sonnet-4.6}; the open-weight models are \texttt{llama-4-maverick} and \texttt{minimax-01}.

\paragraph{Memory Agent/RAG baselines.}
The second family represents retrieval-based memory systems \cite{lewis2020rag}. Unlike full-context baselines, these methods construct a persistent memory representation from the dialogue history during ingestion stage. At question-answering time, a small set of relevant memories is retrieved and inserted into the answer prompt. The final answer is generated by the same fixed answering model, \texttt{openai/gpt-4.1-mini}, across all memory-based baselines. Thus, whenever we refer to an Agent/RAG or memory-based method, we compare memory layers rather than answer-generation models. Performance differences within this family primarily reflect how well each method stores, updates, structures, and retrieves memories for the fixed answering model.

Our simple RAG baseline stores original message-pair chunks from the dialogue history and retrieves them directly from a vector database. The other systems go beyond this plain setup: they use LLM-based components to extract, transform, consolidate, or structure memories before retrieval. We therefore use Agent/RAG as a compact label for retrieval-based memory systems with different degrees of LLM-mediated memory processing, even though not all of them are agents in the narrow sense. \cite{singh2025agenticrag, wang2024llmagents}.

In our experiments, this family includes simple vector-based RAG as well as dedicated memory frameworks: \texttt{mem0} (vector-based) and \texttt{mem0g} (graph-based) \cite{chhikara2025mem0}, \texttt{graphiti} (open-source \texttt{zep}) \cite{rasmussen2025zep}, \texttt{cortex} \cite{cortex_memory_system}, and \texttt{memOS} \cite{li2025memos}. We followed the setup described in the corresponding papers and repositories. For reproducibility, we integrated the open-source frameworks as \texttt{git} submodules, while \texttt{mem0(g)} was accessed through its hosted commercial API.  When a framework allowed replacing the embedding backend, we used language-specific embedding models of comparable dimension and parameter scale: \texttt{FRIDA} \cite{aiforever2024frida} for Russian and \texttt{Qwen3-Embedding-0.6B} \cite{zhang2025qwen3embedding} for English, selected as strong retrieval-oriented embedding backbones on MTEB benchmark \cite{muennighoff-etal-2023-mteb}.

\subsection{Evaluation Pipeline}
\label{subsec:setup}

In addition to the dataset, we provide an evaluation methodology and an accompanying pipeline (Fig.\ref{fig:pipeline}) as a reference protocol for validating memory-based systems. The proposed pipeline follows evaluation practices commonly used in long-context dialogue and memory benchmarks referenced in Sec.\ref{sec:related}.

RUMBA is evaluated using a two-stage pipeline: \textbf{\textit{add}} and \textbf{\textit{eval}}. In the first \textit{add} stage, the memory system processes the dialogue data sequentially as user--assistant pairs and populates its internal storage by extracting and storing memories (along with their timestamps) from the conversations. Dialogues belonging to different users are processed independently, and the resulting memories are stored in isolated user-specific memory spaces, e.g., database collections. 
The Russian and English subsets are ingested separately, resulting in distinct language-specific memory stores.
This stage is skipped in the full-context evaluation setup. 

In the \textit{eval} stage, the system is evaluated using the open-source \texttt{Lighteval} framework~\cite{lighteval}. For each benchmark question, the system first retrieves the top-$k=10$ relevant memories from the corresponding store. The retrieved timestamped\footnote{The \texttt{mem0g} API did not provide timestamps for graph memories. To evaluate the graph-based implementation as is, we did not augment it with timestamped vector memories, resulting in an atemporal graph-based setup.} memories, along with timestamped question, are then provided as context to an answer-generation model, which produces a response based on a fixed language-specific prompt (Fig. \ref{fig:answer_generation_prompts}).
The generated answer is evaluated using an LLM-as-Judge procedure that compares the predicted answer with the reference answer. Different judge models are used depending on the evaluation language. For Russian evaluation, we use \texttt{POLLUX}~\cite{martynov2025eyejudgementdissectingevaluation}, a pretrained judge model for Russian language, which was selected based on a reported correlation of $\rho = 0.704$ with expert judgments. 
For English evaluation, the same scale and criteria are applied, while \texttt{DeepSeek-R1} is used as a judge model. We discuss the choice of judges in Appendix~\ref{sec:judge_сhoice}.
The pipeline produces LLM-as-Judge accuracy as the primary metric, with F1 reported as a complementary one. The judge assigns labels 0--3; labels 0 and 1 are mapped to incorrect, while labels 2 and 3 are mapped to correct for binary accuracy. F1 is the mean per-question bag-of-words F1 between the generated and reference answers, using the built-in \texttt{Lighteval} implementation: words are split on whitespace and treated as sets, and the maximum F1 is used for questions with multiple references.
The full evaluation criteria and judge prompt templates are provided in Appendix~\ref{sec:judge_promts}.

The analysis includes: 
(1) a family-level comparison of retrieval-based memory systems against full-context baselines; and 
(2) a matched-answerer comparison between the \texttt{gpt-4.1-mini} full-context baseline and Agent/RAG systems; and
(3) within-family slice contrasts over different benchmark scopes.
Confidence intervals are estimated by bootstrap resampling over questions, and $p$-values are Holm-corrected within each comparison block. 
All results are reported separately for Russian and English as well.
Because \texttt{mem0g} is atemporal in our setup, we exclude it from Agent/RAG family aggregates for temporal, temporal-expression, and scope--temporal analyses.

\section{Results}
\label{sec:results}

\begin{table*}[t]
\centering
\scriptsize
\setlength{\tabcolsep}{2.5pt}
\begin{tabularx}{\textwidth}{>{\raggedright\arraybackslash}X
                                >{\raggedright\arraybackslash}p{0.25\textwidth}
                                cccc}
\toprule
\textbf{Method}
& 
& \multicolumn{2}{c}{\textbf{Russian}}
& \multicolumn{2}{c}{\textbf{English}} \\
\cmidrule(lr){3-4}
\cmidrule(lr){5-6}
& & \textbf{LLM-as-Judge} & \textbf{F1} & \textbf{LLM-as-Judge} & \textbf{F1} \\
\midrule
\textbf{Agent/RAG} & \textbf{Embedder} & & & & \\
\midrule
mem0g (atemporal)
& --
& \(34.93 \pm 1.21\) & \(20.52 \pm 0.81\) & \(35.26 \pm 1.22\) & \(24.27 \pm 0.83\) \\
graphiti (open-source zep)
& FRIDA \textbar{} Qwen3-Embedding-0.6B
& \(42.71 \pm 1.26\) & \(27.00 \pm 0.91\) & \(45.11 \pm 1.27\) & \(29.13 \pm 0.90\) \\
mem0
& --
& \(53.21 \pm 1.27\) & \(30.10 \pm 0.95\) & \(54.12 \pm 1.27\) & \(35.05 \pm 0.95\) \\
cortex
& FRIDA \textbar{} Qwen3-Embedding-0.6B
& \(68.24 \pm 1.19\) & \(36.32 \pm 0.99\) & \(55.80 \pm 1.26\) & \(34.26 \pm 0.94\) \\
simple RAG
& FRIDA \textbar{} Qwen3-Embedding-0.6B
& \(\mathbf{68.89 \pm 1.18}\) & \(36.17 \pm 0.98\) & \(65.46 \pm 1.21\) & \(39.00 \pm 0.96\) \\
memOS
& FRIDA \textbar{} Qwen3-Embedding-0.6B
& \(66.82 \pm 1.20\) & \(\mathbf{38.27 \pm 1.00}\) & \(\mathbf{69.99 \pm 1.17}\) & \(\mathbf{42.59 \pm 0.96}\) \\
\midrule
\textbf{Full context} & \textbf{Context size} & & & & \\
\midrule
llama-4-maverick
& 1M tokens
& \(48.35 \pm 1.27\) & \(27.18 \pm 0.97\) & \(50.81 \pm 1.27\) & \(28.58 \pm 0.95\) \\
minimax/minimax-01
& 1M tokens
& \(56.71 \pm 1.26\) & \(31.99 \pm 1.02\) & \(57.42 \pm 1.26\) & \(31.68 \pm 0.90\) \\
gpt4.1-mini
& 1M tokens
& \(60.08 \pm 1.25\) & \(27.26 \pm 0.87\) & \(62.93 \pm 1.23\) & \(33.68 \pm 0.89\) \\
gemini-3.1-flash-lite
& 1M tokens
& \(70.06 \pm 1.17\) & \(39.30 \pm 1.06\) & \(69.09 \pm 1.18\) & \(39.93 \pm 0.97\) \\
x-ai/grok-4.1-fast
& 2M tokens
& \(76.22 \pm 1.08\) & \(34.40 \pm 0.92\) & \(79.59 \pm 1.03\) & \(23.34 \pm 0.76\) \\
claude-sonnet-4.6
& 1M tokens
& \(78.61 \pm 1.04\) & \(\mathbf{43.03 \pm 1.02}\) & \(81.98 \pm 0.98\) & \(\mathbf{40.98 \pm 1.01}\) \\
gpt-5.4
& 1.05M tokens
& \(\mathbf{83.60 \pm 0.94}\) & \(40.43 \pm 0.95\) & \(\mathbf{83.99 \pm 0.93}\) & \(40.75 \pm 0.90\) \\
\bottomrule
\end{tabularx}
\caption{\small Overall RUMBA results for retrieval-based memory systems and full-context baselines. All scores are reported on a 0--100 scale; uncertainty is reported as standard error.}
\label{tab:overall_results}
\end{table*}

We report descriptive overall scores for each method in Table~\ref{tab:overall_results} and then analyze performance across the main benchmark axes: session scope, temporal reasoning, temporal-expression explicitness, and semantic type.
A detailed description of the slice-level analysis and additional results are provided in Appendix~\ref{app:detailed_analysis}. An example of a single model case study is provided in Appendix~\ref{app:case_study}.

\begin{table*}[!t]
\centering
\scriptsize
\makebox[\textwidth][c]{%
\begin{minipage}[t]{0.475\textwidth}
\vspace{0pt}
\centering
\resizebox{\linewidth}{!}{%
\begin{tabular}{@{}llrrr@{}}
\toprule
\textbf{Agent/RAG} & \textbf{Lang.} & \textbf{GPT-4.1-mini FC} & \textbf{Agent/RAG} & \textbf{$\Delta$ [95\% CI]} \\
\midrule
cortex & RU & 60.08 & 68.24 & -8.17 [-10.89, -5.44] \\
cortex & EN & 62.93 & 55.80 & +7.13 [4.41, 9.85] \\
mem0 & RU & 60.08 & 53.21 & +6.87 [4.02, 9.66] \\
mem0 & EN & 62.93 & 54.12 & +8.81 [6.03, 11.60] \\
mem0g & RU & 60.08 & 34.93 & +25.15 [22.29, 28.06] \\
mem0g & EN & 62.93 & 35.26 & +27.67 [24.89, 30.46] \\
memOS & RU & 60.08 & 66.82 & -6.74 [-9.46, -4.08] \\
memOS & EN & 62.93 & 69.99 & -7.06 [-9.66, -4.47] \\
simple RAG & RU & 60.08 & 68.89 & -8.81 [-11.47, -6.22] \\
simple RAG & EN & 62.93 & 65.46 & -2.53 [-5.25, 0.13] \\
graphiti (zep) & RU & 60.08 & 42.71 & +17.37 [14.52, 20.22] \\
graphiti (zep) & EN & 62.93 & 45.11 & +17.82 [14.91, 20.74] \\
\midrule
Overall & RU & 60.08 & 55.80 & \textbf{+4.28} [2.08, 6.43] \\
Overall & EN & 62.93 & 54.29 & \textbf{+8.64} [6.42, 10.77] \\
\bottomrule
\end{tabular}%
}
\caption{\small Direct matched-answerer comparison of the \texttt{gpt-4.1-mini} Full-context baseline with each Agent/RAG system and the overall Agent/RAG family mean. Scores and differences are reported on a 0--100 accuracy-point scale, with $\Delta=\text{\texttt{gpt-4.1-mini} FC}-\text{Agent/RAG}$; confidence intervals are paired bootstrap 95\% CIs over the same 1,543 questions.}
\label{tab:main_family_results_compact}
\end{minipage}%
\hspace{0.025\textwidth}%
\begin{minipage}[t]{0.5\textwidth}
\vspace{0pt}
\centering
\resizebox{\linewidth}{!}{%
\begin{tabular}{@{}llrrr@{}}
\toprule
\textbf{Contrast} & \textbf{Family / Language} & \textbf{A} & \textbf{B} & \textbf{$\Delta$ [95\% CI]} \\
\midrule
Single-session $-$ Multi-session & Agent/RAG Russian & 63.92 & 43.18 & +20.73 [17.71, 23.83] \\
Single-session $-$ Multi-session & Agent/RAG English & 62.92 & 40.87 & +22.06 [19.02, 25.14] \\
Single-session $-$ Multi-session & Full context Russian & 76.65 & 53.69 & \textbf{+22.96} [20.07, 25.88] \\
Single-session $-$ Multi-session & Full context English & 78.21 & 55.70 & \textbf{+22.51} [19.57, 25.50] \\
\midrule
Atemporal $-$ Temporal & Agent/RAG English & 60.02 & 52.87 & \textbf{+7.15} [3.54, 10.70] \\
Atemporal $-$ Temporal & Full context Russian & 69.39 & 62.96 & +6.43 [3.10, 9.67] \\
Atemporal $-$ Temporal & Full context English & 70.91 & 65.30 & +5.61 [2.23, 8.93] \\
\midrule
Explicit time evidence $-$ Implicit time evidence & Agent/RAG Russian & 63.33 & 44.35 & +18.99 [9.68, 28.61] \\
Explicit time evidence $-$ Implicit time evidence & Agent/RAG English & 60.20 & 38.55 & +21.65 [11.82, 31.34] \\
Explicit time evidence $-$ Implicit time evidence & Full context Russian & 76.19 & 46.17 & \textbf{+30.02} [21.90, 37.77] \\
Explicit time evidence $-$ Implicit time evidence & Full context English & 76.89 & 47.00 & \textbf{+29.89} [21.61, 38.35] \\
\midrule
Extraction $-$ Reasoning & Agent/RAG Russian & 56.12 & 41.88 & \textbf{+14.24} [10.26, 18.11] \\
Extraction $-$ Reasoning & Agent/RAG English & 53.82 & 41.67 & +12.15 [8.32, 16.02] \\
Extraction $-$ Reasoning & Full context Russian & 69.20 & 55.02 & \textbf{+14.18} [10.22, 18.19] \\
Extraction $-$ Reasoning & Full context English & 69.85 & 60.51 & +9.35 [5.48, 13.33] \\
\bottomrule
\end{tabular}%
}
{\scriptsize\emph{Note:} Agent/RAG temporal contrasts exclude \texttt{mem0g}.}
\caption{\small Main statistically significant within-family slice-difficulty contrasts. Scores are family-level means of binary LLM-as-judge correctness across methods within each family; scores and differences are reported on a 0--100 accuracy-point scale.}
\label{tab:main_slice_difficulty_compact}
\end{minipage}%
}
\end{table*}

\textbf{With the answerer controlled, the} \texttt{gpt-4.1-mini} \textbf{Full-context baseline exceeds the overall Agent/RAG} score by +4.28 points in Russian and +8.64 in English. Method-level results are mixed: cortex, memOS, and simple RAG score significantly higher in Russian; in English, memOS scores significantly higher, while simple RAG is not significantly different (Table~\ref{tab:main_family_results_compact}).

Semantic supergroup analysis shows that the \textbf{observed Full-context advantage is concentrated in Reasoning and Extraction questions}, but not in Abstention. Across both Agent/RAG and Full-context systems, Extraction questions are significantly easier than Reasoning questions in both languages, with within-family accuracy gaps ranging from $+9.35$ to $+14.24$ points (Table~\ref{tab:main_slice_difficulty_compact}).

\textbf{Among the three main benchmark axes, session scope is the strongest and most consistent difficulty driver.}
Across both method families and both languages, single-session questions are easier than multi-session questions by $20.73$--$22.96$ accuracy points, with all confidence intervals excluding zero (Table~\ref{tab:main_slice_difficulty_compact}). Full-context systems stay more accurate on multi-session questions but are not immune to this difficulty.

\textbf{Temporal questions are also harder on average}, although the effect is smaller and less uniform than the single-to-multi degradation. Among statistically significant contrasts, the atemporal-to-temporal gap ranges from $5.61$ to $7.15$ accuracy points depending on language and method family; the Russian Agent/RAG contrast is not significant after correction (Table~\ref{tab:main_slice_difficulty_compact}). In English, the Full-context--Agent/RAG gap is $+12.43$ on temporal and $+10.89$ on atemporal questions.

\textbf{Implicit temporal expressions are a particularly strong failure mode.}
Questions whose supporting evidence contains explicit temporal expressions are substantially easier for models than questions whose evidence requires resolving implicit temporal expressions, for both method families and both languages (Table~\ref{tab:main_slice_difficulty_compact}). The gap is especially large for Full-context systems: $+30.02$ accuracy points in Russian and $+29.89$ accuracy points in English.


\section{Conclusion}

We introduced RUMBA, the first benchmark for evaluating long-term conversational memory in Russian, together with an aligned English version. Unlike existing English-centric benchmarks, RUMBA accounts for the cultural and linguistic specifics of Russian, including natural interaction patterns and temporal expressions. We proposed a multilingual methodology with a fine-grained taxonomy covering contextual recall, entity tracking, temporal reasoning, and multi-step integration of evidence. We released RUMBA as an open-source resource\footnote{\url{https://github.com/ai-forever/RUMBA}} and evaluated state-of-the-art LLMs under both RAG and full-context scenarios. Our results show that current models struggle significantly with tasks requiring simultaneous memory retention, temporal reasoning, and multi-step integration in Russian. RUMBA thus serves as both a diagnostic tool and a catalyst for developing culturally aware conversational agents with robust long-term memory. Future work includes scaling the dataset, enriching reasoning-heavy examples, and probing internal memory representations beyond QA.

\section*{Limitations}
\label{sec:limitations}

\textit{Data.}
The dataset size, while adequate for initial benchmarking, is not large enough to introduce substantial difficulties for very long-context models. Scaling up both the number of dialogues and their length would further stress-test long-context capabilities. In addition, extraction-oriented questions dominate, whereas complex reasoning and temporally entangled queries are underrepresented. This imbalance reflects practical considerations and ensures a solid foundation, but it limits evaluation of higher-order memory abilities. Finally, the English version is obtained through automatic translation rather than manual rewriting, with no explicit localization applied. This may make the English split less culturally natural and introduce translation artifacts that can affect model understanding and embedding-based retrieval.

\textit{Evaluation.}
Our Russian and English evaluations use different LLM judges. Consequently, part of the observed cross-lingual difference may stem from judge-specific calibration rather than model performance alone~\cite{thakur2024judgingjudges, multilingualllmjudge2025}. RUMBA also follows a standard QA pipeline, which enables transparent and reproducible evaluation of output correctness but does not directly probe the model's internal memory representations. Nevertheless, the benchmark targets memory capabilities by construction: unlike open-domain QA, RUMBA centers on user-specific information that evolves over time, including repeated updates, corrections, and changes in personal facts.

\textit{Retrieval diagnostics.}
Our experiments include retrieval-based memory systems, but do not provide a systematic diagnostic analysis of retrieval design choices, such as the choice of embedder, ranking method, retrieval depth, etc. This is outside the main scope of the paper: our goal is to compare memory approaches under a unified pipeline and to show how RUMBA can be used for slice-level diagnostics, rather than to optimize a particular retrieval architecture.

\textit{Future work.}
The rapid development of long-term memory benchmarks and memory-augmented agents suggests several natural directions for future work. Recent benchmark extensions, such as LongMemEval-V2~\cite{wu2026longmemevalv2}, and ongoing updates to memory frameworks\footnote{\url{https://mem0.ai/research}}, appeared after RUMBA creation,  illustrate the increasing importance of larger-scale, more agentic, and more diagnostic memory evaluation. Looking ahead, future versions of RUMBA can (i) scale the dataset by adding more dialogues and extending session lengths, (ii) enrich the benchmark with a broader variety of reasoning-intensive and temporally entangled examples to improve class diversity, (iii) develop complementary diagnostics for retrieval behavior, memory retention, consistency, and update tracking across sessions, thereby further strengthening RUMBA's utility as a comprehensive evaluation framework.

\section*{Ethical consideration}

\paragraph{Data collection and annotation} All dialogues, questions, and answers were collected from scratch, without reusing any existing copyrighted or restricted datasets. Annotations were performed by native Russian speakers with prior experience in linguistic tasks. All annotators were compensated at a rate exceeding the local minimum wage, and participation was voluntary. Written informed consent was obtained from all annotators, and no personally identifiable information was collected.

\paragraph{Licensing and accessibility} The resulting benchmark is released under a public license (MIT). The Russian and English subsets are both distributed under the same terms.

\paragraph{AI assistant usage} During the preparation of this manuscript, the authors used AI assistants solely for editing and improving the English language of the text, including grammar correction, phrasing refinement, and vocabulary suggestions. 
All intellectual contributions, benchmark design decisions, and interpretations remain the sole responsibility of the authors.

\section*{Acknowledgments}
We thank all those who contributed to the RUMBA dataset. The dialogues were written, edited, and reviewed by Irina Shamova, Ksenia Kuleshova, Olga Borodina, Maria Avilova, Angelina Pryadko, Elena Dorofeeva, Tatiana Chabanyuk, Anna Plotnikova, Margarita Khurramova, Maksim Anoshin, Pavel Prosyanik, Ksenia Kondrakova, Yulia Piliguzova, Ekaterina Kuryachaya, Allam Atabaev, Inessa Kurevleva, Zakhar Chaichenko, Victoria Nikolaeva, Olga Borisova, Vsevolod Alipov, Danil Vyazovov, Evgenia Efremova, Timur Shenturk, and Konstantin Gridin. Final validation was performed by Olga Borisova, Taisiya Metelkina, Maria Saburbaeva, Yulia Piliguzova, Ekaterina Kuryachaya, and Allam Atabaev, and we are especially grateful to Artemy Stankevich for making the validation process possible. The English version was worked on by Ekaterina Kuryachaya, Allam Atabaev, Natalia Chukicheva, Ksenia Kuleshova, Dmitry Kosourov, Eleonora Petrikova, Alena Gubar, Maria Petrova, Lada Kovalchuk, and Daria Panaiotti. Finally, we appreciate the management efforts of Elvira Budaeva, Pavel Kovalev, Anna Kostikova, and Dmitry Bocharov, as well as the overall coordination by Yuliya Vasilieva.
\bibliography{custom}

@inproceedings{maharana2024locomo,
    title = "Evaluating Very Long-Term Conversational Memory of {LLM} Agents",
    author = "Maharana, Adyasha  and
      Lee, Dong-Ho  and
      Tulyakov, Sergey  and
      Bansal, Mohit  and
      Barbieri, Francesco  and
      Fang, Yuwei",
    editor = "Ku, Lun-Wei  and
      Martins, Andre  and
      Srikumar, Vivek",
    booktitle = "Proceedings of the 62nd Annual Meeting of the Association for Computational Linguistics (Volume 1: Long Papers)",
    month = aug,
    year = "2024",
    address = "Bangkok, Thailand",
    publisher = "Association for Computational Linguistics",
    url = "https://aclanthology.org/2024.acl-long.747/",
    doi = "10.18653/v1/2024.acl-long.747",
    pages = "13851--13870",
}

@inproceedings{wu2024longmemeval,
title={{LongMemEval}: Benchmarking Chat Assistants on Long-Term Interactive Memory},
author={Di Wu and Hongwei Wang and Wenhao Yu and Yuwei Zhang and Kai-Wei Chang and Dong Yu},
booktitle={The Thirteenth International Conference on Learning Representations},
year={2025},
url={https://openreview.net/forum?id=pZiyCaVuti}
}

@misc{bei2026memgallery,
      title={Mem-Gallery: Benchmarking Multimodal Long-Term Conversational Memory for {MLLM} Agents}, 
      author={Yuanchen Bei and Tianxin Wei and Xuying Ning and Yanjun Zhao and Zhining Liu and Xiao Lin and Yada Zhu and Hendrik Hamann and Jingrui He and Hanghang Tong},
      year={2026},
      eprint={2601.03515},
      archivePrefix={arXiv},
      primaryClass={cs.CL},
      url={https://arxiv.org/abs/2601.03515}, 
}

@inproceedings{fenogenova2024mera,
    title = "{MERA}: A Comprehensive {LLM} Evaluation in {R}ussian",
    author = "Fenogenova, Alena  and
      Chervyakov, Artem  and
      Martynov, Nikita  and
      Kozlova, Anastasia  and
      Tikhonova, Maria  and
      Akhmetgareeva, Albina  and
      Emelyanov, Anton  and
      Shevelev, Denis  and
      Lebedev, Pavel  and
      Sinev, Leonid  and
      Isaeva, Ulyana  and
      Kolomeytseva, Katerina  and
      Moskovskiy, Daniil  and
      Goncharova, Elizaveta  and
      Savushkin, Nikita  and
      Mikhailova, Polina  and
      Minaeva, Anastasia  and
      Dimitrov, Denis  and
      Panchenko, Alexander  and
      Markov, Sergey",
    editor = "Ku, Lun-Wei  and
      Martins, Andre  and
      Srikumar, Vivek",
    booktitle = "Proceedings of the 62nd Annual Meeting of the Association for Computational Linguistics (Volume 1: Long Papers)",
    month = aug,
    year = "2024",
    address = "Bangkok, Thailand",
    publisher = "Association for Computational Linguistics",
    url = "https://aclanthology.org/2024.acl-long.534/",
    doi = "10.18653/v1/2024.acl-long.534",
    pages = "9920--9948"
}

@inproceedings{churin2024libra,
    title = "Long Context Benchmark for the {R}ussian Language",
    author = "Churin, Igor  and
      Apishev, Murat  and
      Tikhonova, Maria  and
      Shevelev, Denis  and
      Bulatov, Aydar  and
      Kuratov, Yuri  and
      Averkiev, Sergei  and
      Fenogenova, Alena",
    editor = "Strube, Michael  and
      Braud, Chloe  and
      Hardmeier, Christian  and
      Li, Junyi Jessy  and
      Loaiciga, Sharid  and
      Zeldes, Amir  and
      Li, Chuyuan",
    booktitle = "Proceedings of the 6th Workshop on Computational Approaches to Discourse, Context and Document-Level Inferences (CODI 2025)",
    month = nov,
    year = "2025",
    address = "Suzhou, China",
    publisher = "Association for Computational Linguistics",
    url = "https://aclanthology.org/2025.codi-1.1/",
    doi = "10.18653/v1/2025.codi-1.1",
    pages = "1--13",
    ISBN = "979-8-89176-343-2",
}

@misc{martynov2025eyejudgementdissectingevaluation,
      title={Eye of Judgement: Dissecting the Evaluation of Russian-speaking {LLMs} with {POLLUX}}, 
      author={Nikita Martynov and 
      Anastasia Mordasheva and 
      Dmitriy Gorbetskiy and 
      Danil Astafurov and 
      Ulyana Isaeva and 
      Elina Basyrova and 
      Sergey Skachkov and 
      Victoria Berestova and 
      Nikolay Ivanov and 
      Valeriia Zanina and 
      Alena Fenogenova},
      year={2025},
      eprint={2505.24616},
      archivePrefix={arXiv},
      primaryClass={cs.CL},
      url={https://arxiv.org/abs/2505.24616}, 
}

@misc{lighteval,
  author = {Habib, Nathan and Fourrier, Clémentine and Kydlíček, Hynek and Wolf, Thomas and Tunstall, Lewis},
  title = {LightEval: A lightweight framework for LLM evaluation},
  year = {2023},
  version = {0.11.0},
  url = {https://github.com/huggingface/lighteval}
}

@inproceedings{vashishtha-etal-2020-temporal,
    title = "Temporal Reasoning in Natural Language Inference",
    author = "Vashishtha, Siddharth  and
      Poliak, Adam  and
      Lal, Yash Kumar  and
      Van Durme, Benjamin  and
      White, Aaron Steven",
    editor = "Cohn, Trevor  and
      He, Yulan  and
      Liu, Yang",
    booktitle = "Findings of the Association for Computational Linguistics: EMNLP 2020",
    month = nov,
    year = "2020",
    address = "Online",
    publisher = "Association for Computational Linguistics",
    url = "https://aclanthology.org/2020.findings-emnlp.363/",
    doi = "10.18653/v1/2020.findings-emnlp.363",
    pages = "4070--4078"
}

@inproceedings{chu-etal-2024-timebench,
    title = "{T}ime{B}ench: A Comprehensive Evaluation of Temporal Reasoning Abilities in Large Language Models",
    author = "Chu, Zheng  and
      Chen, Jingchang  and
      Chen, Qianglong  and
      Yu, Weijiang  and
      Wang, Haotian  and
      Liu, Ming  and
      Qin, Bing",
    editor = "Ku, Lun-Wei  and
      Martins, Andre  and
      Srikumar, Vivek",
    booktitle = "Proceedings of the 62nd Annual Meeting of the Association for Computational Linguistics (Volume 1: Long Papers)",
    month = aug,
    year = "2024",
    address = "Bangkok, Thailand",
    publisher = "Association for Computational Linguistics",
    url = "https://aclanthology.org/2024.acl-long.66/",
    doi = "10.18653/v1/2024.acl-long.66",
    pages = "1204--1228"
}

@misc{piryani2026itshightimesurvey,
      title={It's High Time: A Survey of Temporal Question Answering}, 
      author={Bhawna Piryani and Abdelrahman Abdallah and Jamshid Mozafari and Avishek Anand and Adam Jatowt},
      year={2026},
      eprint={2505.20243},
      archivePrefix={arXiv},
      primaryClass={cs.CL},
      url={https://arxiv.org/abs/2505.20243}, 
}

@inproceedings{gigachatteam2025gigachatfamilyefficientrussian,
    title = "{G}iga{C}hat Family: Efficient {R}ussian Language Modeling Through Mixture of Experts Architecture",
    author = "Mamedov, Valentin  and
      Kosarev, Evgenii  and
      Leleytner, Gregory  and
      Shchuckin, Ilya  and
      Berezovskiy, Valeriy  and
      Smirnov, Daniil  and
      Kozlov, Dmitry  and
      Averkiev, Sergei  and
      Ivan, Lukyanenko  and
      Proshunin, Aleksandr  and
      Israfilova, Ainur  and
      Baskov, Ivan  and
      Chervyakov, Artem  and
      Shakirov, Emil  and
      Kolesov, Mikhail  and
      Khomich, Daria  and
      Latortseva, Daria  and
      Porkhun, Sergei  and
      Fedorov, Yury  and
      Kutuzov, Oleg  and
      Kudriavtseva, Polina  and
      Soldatova, Sofiia  and
      Egor, Kolodin  and
      Pyatkin, Stanislav  and
      Menshykh, Dzmitry  and
      IUrevich, Grafov Sergei  and
      Damirov, Eldar  and
      Karlov, Vladimir  and
      Gaitukiev, Ruslan  and
      Shatenov, Arkadiy  and
      Fenogenova, Alena  and
      Savushkin, Nikita  and
      Minkin, Fedor",
    editor = "Mishra, Pushkar  and
      Muresan, Smaranda  and
      Yu, Tao",
    booktitle = "Proceedings of the 63rd Annual Meeting of the Association for Computational Linguistics (Volume 3: System Demonstrations)",
    month = jul,
    year = "2025",
    address = "Vienna, Austria",
    publisher = "Association for Computational Linguistics",
    doi = "10.18653/v1/2025.acl-demo.10",
    pages = "93--106",
}

@inproceedings{chervyakov2026multimodalevaluationrussianlanguagearchitectures,
    title = "Multimodal Evaluation of {R}ussian-language Architectures",
    author = "Chervyakov, Artem  and
      Isaeva, Ulyana  and
      Emelyanov, Anton  and
      Safin, Artem  and
      Tikhonova, Maria  and
      Kharitonov, Alexander  and
      Lyakh, Yulia  and
      Surovtsev, Petr  and
      Shevelev, Denis  and
      Saburov, Vildan  and
      Konovalov, Vasily  and
      Rykov, Elisei  and
      Sviridov, Ivan  and
      Miftakhova, Amina  and
      Alimova, Ilseyar  and
      Panchenko, Alexander  and
      Kapitanov, Alexander  and
      Fenogenova, Alena",
    editor = "Demberg, Vera  and
      Inui, Kentaro  and
      Marquez, Llu{\'i}s",
    booktitle = "Proceedings of the 19th Conference of the {E}uropean Chapter of the {A}ssociation for {C}omputational {L}inguistics (Volume 1: Long Papers)",
    month = mar,
    year = "2026",
    address = "Rabat, Morocco",
    publisher = "Association for Computational Linguistics",
    doi = "10.18653/v1/2026.eacl-long.94",
    pages = "2114--2161",
}

@inproceedings{lewis2020rag,
  author = {Lewis, Patrick and Perez, Ethan and Piktus, Aleksandra and Petroni, Fabio and Karpukhin, Vladimir and Goyal, Naman and K{\"u}ttler, Heinrich and Lewis, Mike and Yih, Wen-tau and Rockt{\"a}schel, Tim and Riedel, Sebastian and Kiela, Douwe},
  title = {Retrieval-Augmented Generation for Knowledge-Intensive {NLP} Tasks},
  booktitle = {Advances in Neural Information Processing Systems 33},
  year = {2020},
  url = {https://proceedings.neurips.cc/paper/2020/hash/6b493230205f780e1bc26945df7481e5-Abstract.html}
}

@article{wang2024llmagents,
  title = {A Survey on Large Language Model based Autonomous Agents},
  author = {Wang, Lei and Ma, Chen and Feng, Xueyang and Zhang, Zeyu and Yang, Hao and Zhang, Jingsen and Chen, Zhiyuan and Tang, Jiakai and Chen, Xu and Lin, Yankai and Zhao, Wayne Xin and Wei, Zhewei and Wen, Ji-Rong},
  journal = {Frontiers of Computer Science},
  volume = {18},
  number = {6},
  pages = {186345},
  year = {2024},
  doi = {10.1007/s11704-024-40231-1},
  url = {https://arxiv.org/abs/2308.11432}
}

@article{singh2025agenticrag,
  title = {Agentic Retrieval-Augmented Generation: A Survey on Agentic {RAG}},
  author = {Singh, Aditi and Ehtesham, Abul and Kumar, Saket and Khoei, Tala Talaei},
  journal = {arXiv preprint arXiv:2501.09136},
  year = {2025},
  url = {https://arxiv.org/abs/2501.09136}
}

@article{chhikara2025mem0,
  title = {{Mem0}: Building Production-Ready {AI} Agents with Scalable Long-Term Memory},
  author = {Chhikara, Prateek and Khant, Dev and Aryan, Saket and Singh, Taranjeet and Yadav, Deshraj},
  journal = {arXiv preprint arXiv:2504.19413},
  year = {2025},
  url = {https://arxiv.org/abs/2504.19413}
}

@article{rasmussen2025zep,
  title = {{Zep}: A Temporal Knowledge Graph Architecture for Agent Memory},
  author = {Rasmussen, Preston and Paliychuk, Pavlo and Beauvais, Travis and Ryan, Jack and Chalef, Daniel},
  journal = {arXiv preprint arXiv:2501.13956},
  year = {2025},
  url = {https://arxiv.org/abs/2501.13956}
}

@article{li2025memos,
  title = {{MemOS}: An Operating System for Memory-Augmented Generation ({MAG}) in Large Language Models},
  author = {Li, Zhiyu and Song, Shichao and Wang, Hanyu and Niu, Simin and Chen, Ding and Yang, Jiawei and Xi, Chenyang and Lai, Huayi and Zhao, Jihao and Wang, Yezhaohui and Ren, Junpeng and Lin, Zehao and Huo, Jiahao and Chen, Tianyi and Chen, Kai and Li, Kehang and Yin, Zhiqiang and Yu, Qingchen and Tang, Bo and Yang, Hongkang and Xu, Zhi-Qin John and Xiong, Feiyu},
  journal = {arXiv preprint arXiv:2505.22101},
  year = {2025},
  url = {https://arxiv.org/abs/2505.22101}
}

@misc{aiforever2024frida,
  title = {{FRIDA}: Full-Scale Finetuned Retrieval Model Inspired by Denoising Architecture Based on {T5}},
  author = {{ai-forever}},
  year = {2024},
  howpublished = {\url{https://huggingface.co/ai-forever/FRIDA}},
  note = {Hugging Face model card}
}

@article{zhang2025qwen3embedding,
  title = {{Qwen3} Embedding: Advancing Text Embedding and Reranking Through Foundation Models},
  author = {Zhang, Yanzhao and Li, Mingxin and Long, Dingkun and Zhang, Xin and Lin, Huan and Yang, Baosong and Xie, Pengjun and Yang, An and Liu, Dayiheng and Lin, Junyang and Huang, Fei and Zhou, Jingren},
  journal = {arXiv preprint arXiv:2506.05176},
  year = {2025},
  url = {https://arxiv.org/abs/2506.05176}
}

@misc{cortex_memory_system,
  title={Cortex: Advanced Memory System for AI Agents},
  author={Bhattacharjee, Biswaroop},
  year={2025},
  url={https://github.com/prem-research/cortex}
}

@inproceedings{muennighoff-etal-2023-mteb,
    title = "{MTEB}: Massive Text Embedding Benchmark",
    author = "Muennighoff, Niklas and Tazi, Nouamane and Lorentz, Lo{\"\i}c and Reimers, Nils",
    editor = "Rogers, Anna and Boyd-Graber, Jordan and Okazaki, Naoaki",
    booktitle = "Proceedings of the 17th Conference of the European Chapter of the Association for Computational Linguistics",
    month = may,
    year = "2023",
    address = "Dubrovnik, Croatia",
    publisher = "Association for Computational Linguistics",
    url = "https://aclanthology.org",
    pages = "2014--2037",
}

@inproceedings{thakur2024judgingjudges,
    title = "Judging the Judges: Evaluating Alignment and Vulnerabilities in {LLM}s-as-Judges",
    author = "Thakur, Aman Singh  and
      Choudhary, Kartik  and
      Ramayapally, Venkat Srinik  and
      Vaidyanathan, Sankaran  and
      Hupkes, Dieuwke",
    editor = "Arviv, Ofir  and
      Clinciu, Miruna  and
      Dhole, Kaustubh  and
      Dror, Rotem  and
      Gehrmann, Sebastian  and
      Habba, Eliya  and
      Itzhak, Itay  and
      Mille, Simon  and
      Perlitz, Yotam  and
      Santus, Enrico  and
      Sedoc, Jo{\~a}o  and
      Shmueli Scheuer, Michal  and
      Stanovsky, Gabriel  and
      Tafjord, Oyvind",
    booktitle = "Proceedings of the Fourth Workshop on Generation, Evaluation and Metrics (GEM{\texttwosuperior})",
    month = jul,
    year = "2025",
    address = "Vienna, Austria and virtual meeting",
    publisher = "Association for Computational Linguistics",
    url = "https://aclanthology.org/2025.gem-1.33/",
    pages = "404--430",
}

@inproceedings{multilingualllmjudge2025,
  title = {How Reliable is Multilingual {LLM}-as-a-Judge?},
  author = {Fu, Xiyan and Liu, Wei},
  booktitle = {Findings of the Association for Computational Linguistics: EMNLP 2025},
  pages = {11040--11053},
  year = {2025},
  address = {Suzhou, China},
  publisher = {Association for Computational Linguistics},
  doi = {10.18653/v1/2025.findings-emnlp.587},
  url = {https://aclanthology.org/2025.findings-emnlp.587/}
}

@inproceedings{tan2025judgebench,
  title = {{JudgeBench}: A Benchmark for Evaluating {LLM}-Based Judges},
  author = {Tan, Sijun and Zhuang, Siyuan and Montgomery, Kyle and Tang, William Yuan and Cuadron, Alejandro and Wang, Chenguang and Popa, Raluca Ada and Stoica, Ion},
  booktitle = {The Thirteenth International Conference on Learning Representations},
  year = {2025},
  url = {https://openreview.net/forum?id=G0dksFayVq}
}

@article{du2026memory,
  title        = {Memory for Autonomous LLM Agents: Mechanisms, Evaluation, and Emerging Frontiers},
  author       = {Du, Pengfei},
  journal      = {arXiv preprint arXiv:2603.07670},
  year         = {2026},
  eprint       = {2603.07670},
  archivePrefix = {arXiv},
  primaryClass = {cs.AI}
}

@inproceedings{hu2025memoryagentbench,
  title        = {Evaluating Memory in LLM Agents via Incremental Multi-Turn Interactions},
  author       = {Hu, Yuanzhe and Wang, Yu and McAuley, Julian},
  booktitle    = {International Conference on Learning Representations},
  year         = {2026},
  url          = {https://openreview.net/forum?id=DT7JyQC3MR}
}

@article{uddin2026memora,
  title        = {From Recall to Forgetting: Benchmarking Long-Term Memory for Personalized Agents},
  author       = {Uddin, Md Nayem and Shubham, Kumar and Blanco, Eduardo and Baral, Chitta and Wang, Gengyu},
  journal      = {arXiv preprint arXiv:2604.20006},
  year         = {2026},
  eprint       = {2604.20006},
  archivePrefix = {arXiv},
  primaryClass = {cs.CL}
}

@inproceedings{lo2023data,
  title = "Data Sampling and (In)stability in Machine Translation Evaluation",
  author = "Lo, Chi-kiu and Knowles, Rebecca",
  editor = "Rogers, Anna and Boyd-Graber, Jordan and Okazaki, Naoaki",
  booktitle = "Findings of the Association for Computational Linguistics: ACL 2023",
  month = jul,
  year = "2023",
  address = "Toronto, Canada",
  publisher = "Association for Computational Linguistics",
  url = "https://aclanthology.org/2023.findings-acl.826/",
  doi = "10.18653/v1/2023.findings-acl.826",
  pages = "13064--13074"
}

@inproceedings{singh2025globalmmlu,
  title = "Global {MMLU}: Understanding and Addressing Cultural and Linguistic Biases in Multilingual Evaluation",
  author = "Singh, Shivalika and Romanou, Angelika and Fourrier, Cl{\'e}mentine and Adelani, David Ifeoluwa and Ngui, Jian Gang and Vila-Suero, Daniel and Limkonchotiwat, Peerat and Marchisio, Kelly and Leong, Wei Qi and Susanto, Yosephine and Ng, Raymond and Longpre, Shayne and Ruder, Sebastian and Ko, Wei-Yin and Bosselut, Antoine and Oh, Alice and Martins, Andre and Choshen, Leshem and Ippolito, Daphne and Ferrante, Enzo and Fadaee, Marzieh and Ermis, Beyza and Hooker, Sara",
  editor = "Che, Wanxiang and Nabende, Joyce and Shutova, Ekaterina and Pilehvar, Mohammad Taher",
  booktitle = "Proceedings of the 63rd Annual Meeting of the Association for Computational Linguistics (Volume 1: Long Papers)",
  month = jul,
  year = "2025",
  address = "Vienna, Austria",
  publisher = "Association for Computational Linguistics",
  url = "https://aclanthology.org/2025.acl-long.919/",
  doi = "10.18653/v1/2025.acl-long.919",
  pages = "18761--18799",
  ISBN = "979-8-89176-251-0"
}

@inproceedings{rajaee2025empirical,
  title = "An Empirical Analysis of Machine Translation for Expanding Multilingual Benchmarks",
  author = "Rajaee, Sara and Choenni, Rochelle and Shutova, Ekaterina and Monz, Christof",
  editor = "Haddow, Barry and Kocmi, Tom and Koehn, Philipp and Monz, Christof",
  booktitle = "Proceedings of the Tenth Conference on Machine Translation",
  month = nov,
  year = "2025",
  address = "Suzhou, China",
  publisher = "Association for Computational Linguistics",
  url = "https://aclanthology.org/2025.wmt-1.1/",
  doi = "10.18653/v1/2025.wmt-1.1",
  pages = "1--30",
  ISBN = "979-8-89176-341-8"
}

@misc{wu2026longmemevalv2,
  title        = {LongMemEval-V2: Evaluating Long-Term Agent Memory Toward Experienced Colleagues},
  author       = {Wu, Di and Ji, Zixiang and Kawatkar, Asmi and Kwan, Bryan and Gu, Jia-Chen and Peng, Nanyun and Chang, Kai-Wei},
  year         = {2026},
  eprint       = {2605.12493},
  archivePrefix = {arXiv},
  primaryClass = {cs.CL}
}

@misc{chen2025judgelrm,
  title = {{JudgeLRM}: Large Reasoning Models as a Judge},
  author = {Chen, Nuo and Hu, Zhiyuan and Zou, Qingyun and Wu, Jiaying and Wang, Qian and Hooi, Bryan and He, Bingsheng},
  year = {2025},
  eprint = {2504.00050},
  archivePrefix = {arXiv},
  primaryClass = {cs.CL},
  url = {https://arxiv.org/abs/2504.00050}
}

\appendix

\section{Taxonomy details}
\label{sec:taxappendix}

\begin{figure*}[t!]
    \centering
    \includegraphics[width=.95\linewidth]{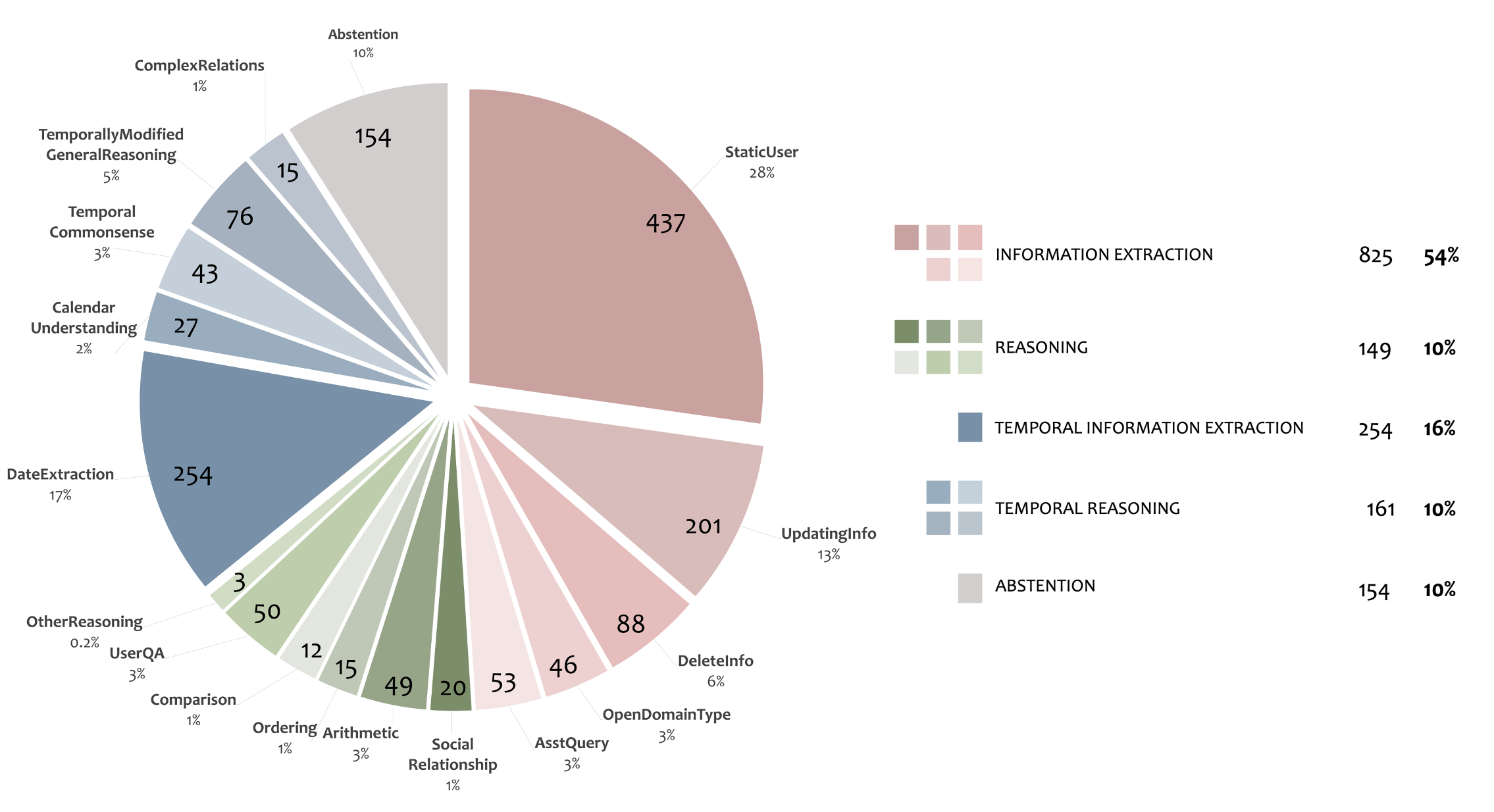}
    \caption{Overview of RUMBA Taxonomy.}
    \label{fig:taxonomy}
\end{figure*}

\paragraph {Semantic axis.} Questions are broadly classified into information-extraction and reasoning questions, based on the information available about the user. A detailed classification of question types, along with their descriptions and illustrative examples, is presented in Table \ref{tab:full_description}.

\begin{figure*}[h!]
    \centering
    \begin{minipage}{0.26\textwidth}
        \centering
        \includegraphics[width=\linewidth]{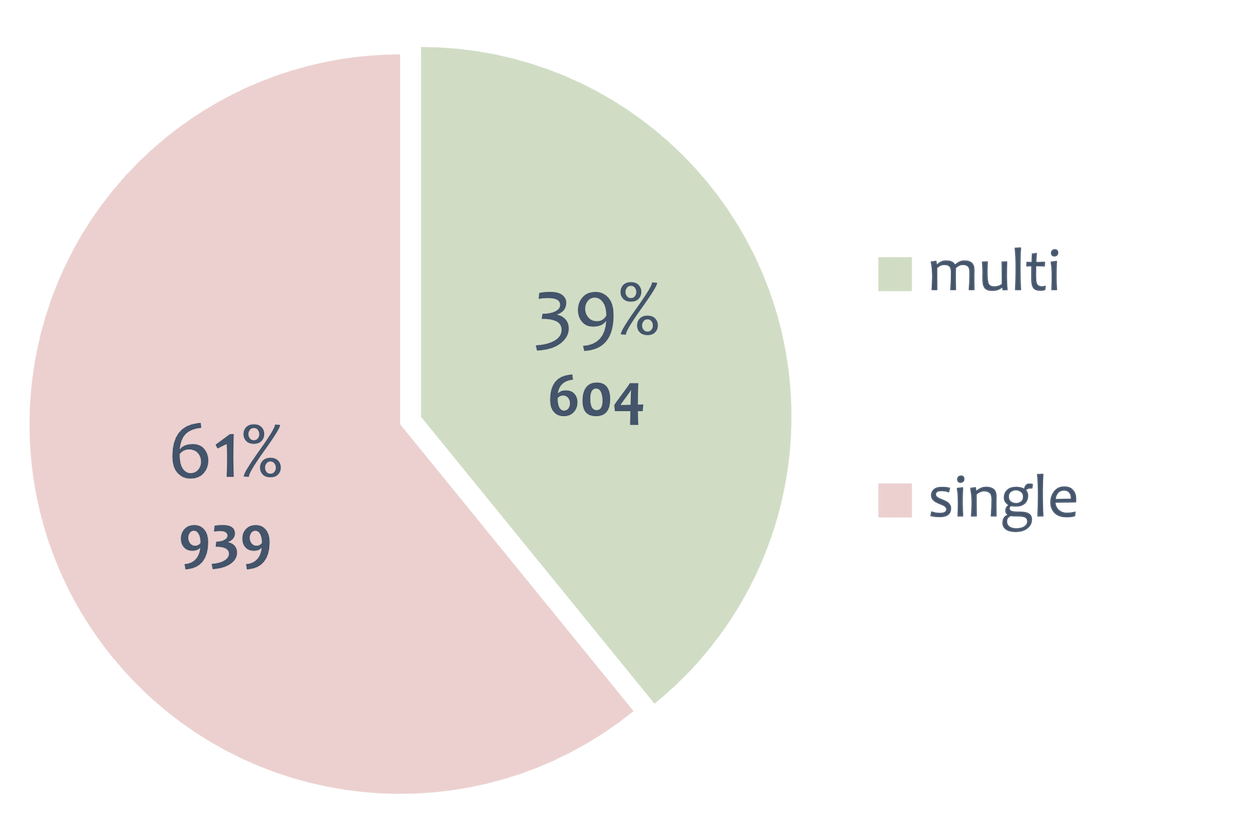}
        \captionsetup{width=.9\linewidth}
        \caption{Distribution across the quantitative axis, reflecting the ratio between single-session and multi-session questions, including both the number of questions and their percentage representation.}
        \label{fig:tax_sin_m}
    \end{minipage}%
    \begin{minipage}{0.26\textwidth}
        \centering
        \includegraphics[width=\linewidth]{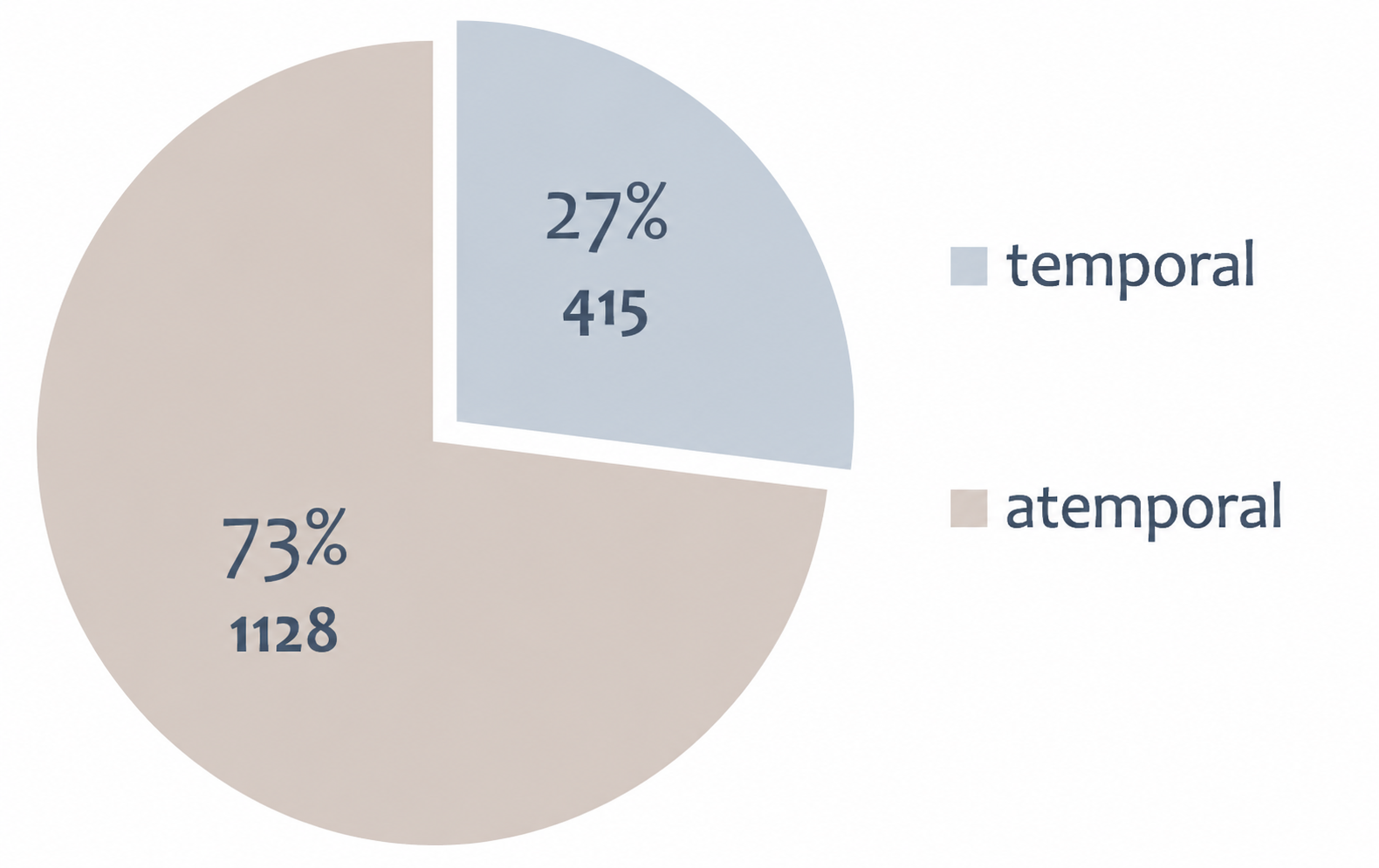}
        \captionsetup{width=0.9\linewidth}
        \caption{Distribution across the axis of temporality, reflecting the ratio between temporal and atemporal questions, including both the number of questions and their percentage representation.}
        \label{fig:tax_temp}
    \end{minipage}
    \begin{minipage}{0.27\textwidth}
        \includegraphics[width=\linewidth]{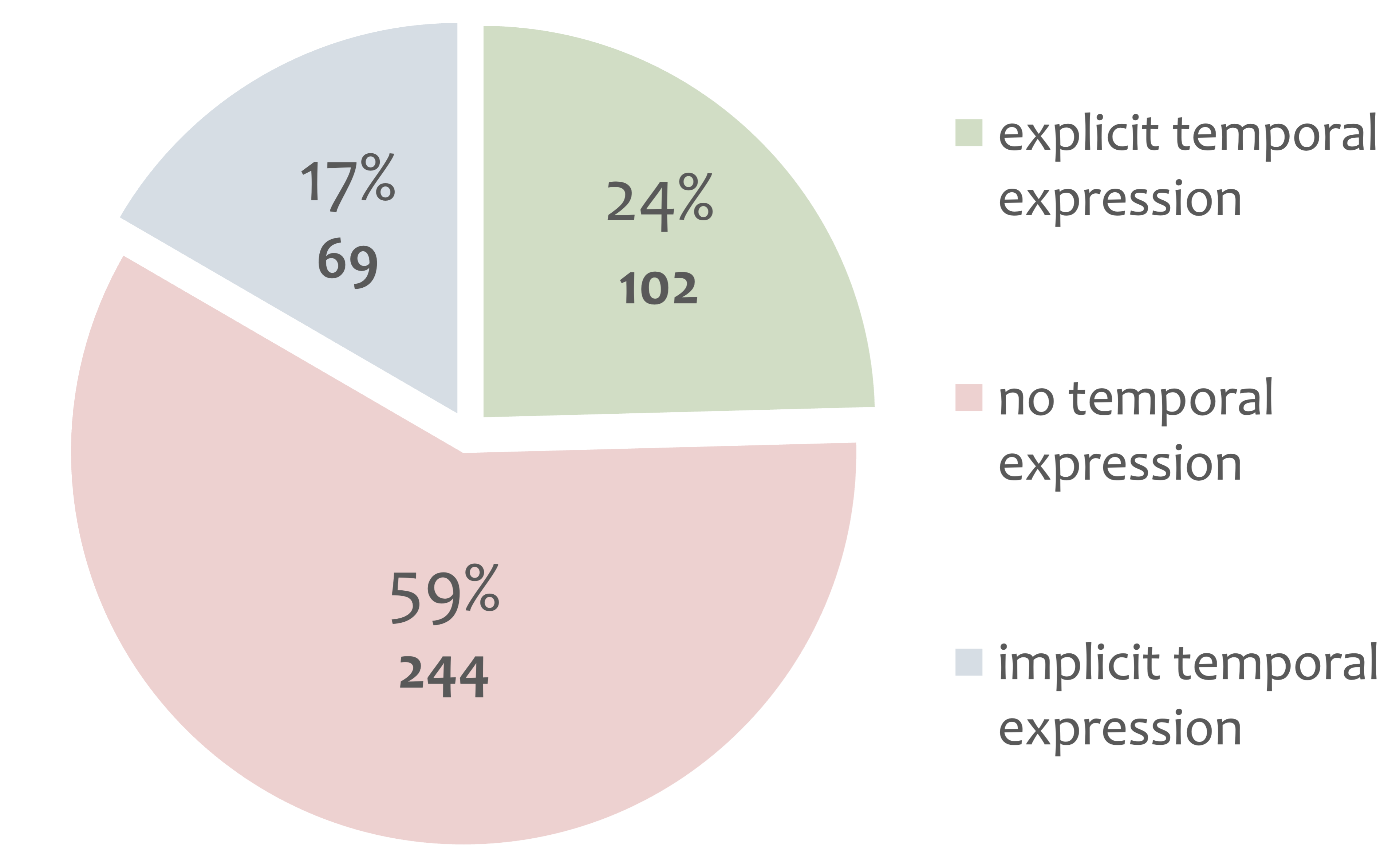}
        \captionsetup{width=0.9\linewidth}
        \caption{Temporal tags: distribution of explicit, implicit, non-expressed temporal expressions in the dialogues, including both the number of expressions and their percentage representation.}
        \label{fig:temporal_tags}
    \end{minipage}%
\end{figure*}

\paragraph {Quantitative axis.} Questions are classified into two types. \textit{Single-session questions}: the answer to the question is contained within a single session, whereas in \textit{multi-session questions}, the answer is distributed across multiple sessions. The distribution of questions on the quantitative axis is depicted in Fig.\ref{fig:tax_sin_m}.

\begin{figure*}[h!]
    \centering
    \includegraphics[width=.42\linewidth]
    {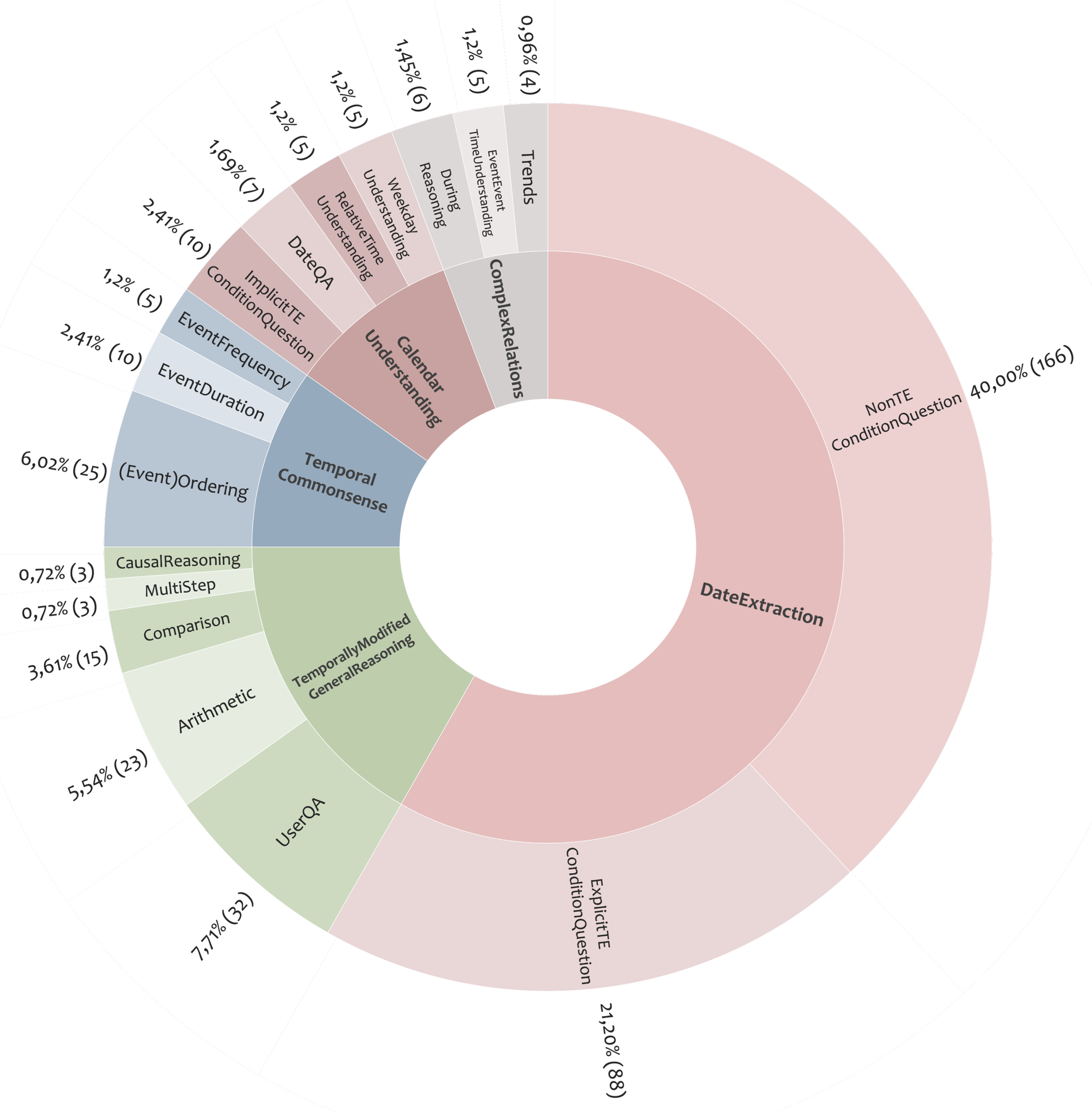}
    \caption{Overview of temporal questions Taxonomy: distribution by type, with corresponding counts and percentages.}
    \label{fig:taxonomy_temporal}
\end{figure*}

\paragraph{Axis of temporality.} Questions are classified as \textit{atemporal}, for which comprehension of the dialogue’s temporal framework is not required, or \textit{temporal}, for which such temporal understanding is essential to answer. The distribution of questions along axis of temporality is presented in Fig.\ref{fig:tax_temp}. The taxonomy of temporal questions is presented in Fig.\ref{fig:taxonomy_temporal}.

Each temporal question is accompanied by a tag indicating how temporal information was presented in the dialogue: \textit{explicit temporal expression} (time was indicated explicitly, e.g., a specific day and/or month, year, etc.), \textit{implicit temporal expression} (time was indicated implicitly, e.g., with words \textit{yesterday, 2 days ago}, etc.), or \textit{no temporal expression} (no temporal information was provided). Fig.\ref{fig:temporal_tags} presents the distribution of temporal expression tags.

\section{Annotation details}
\label{sec:annotation}

\subsection{Dataset Creation}
\label{dataset creation_app}

\begin{figure*}[!ht]
\centering
\includegraphics[width=\textwidth]{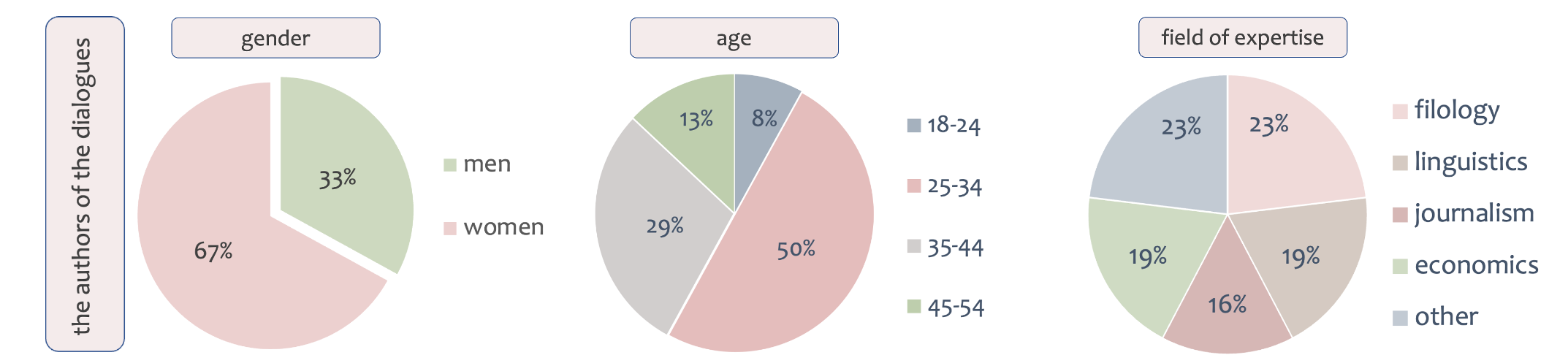} 
\caption{Information about the dialogue authors including gender, age and field of expertise distribution, reported in terms of percentages.} \label{fig:experts} 
\end{figure*}

\begin{figure*}
    \centering
    \includegraphics[width=\textwidth]{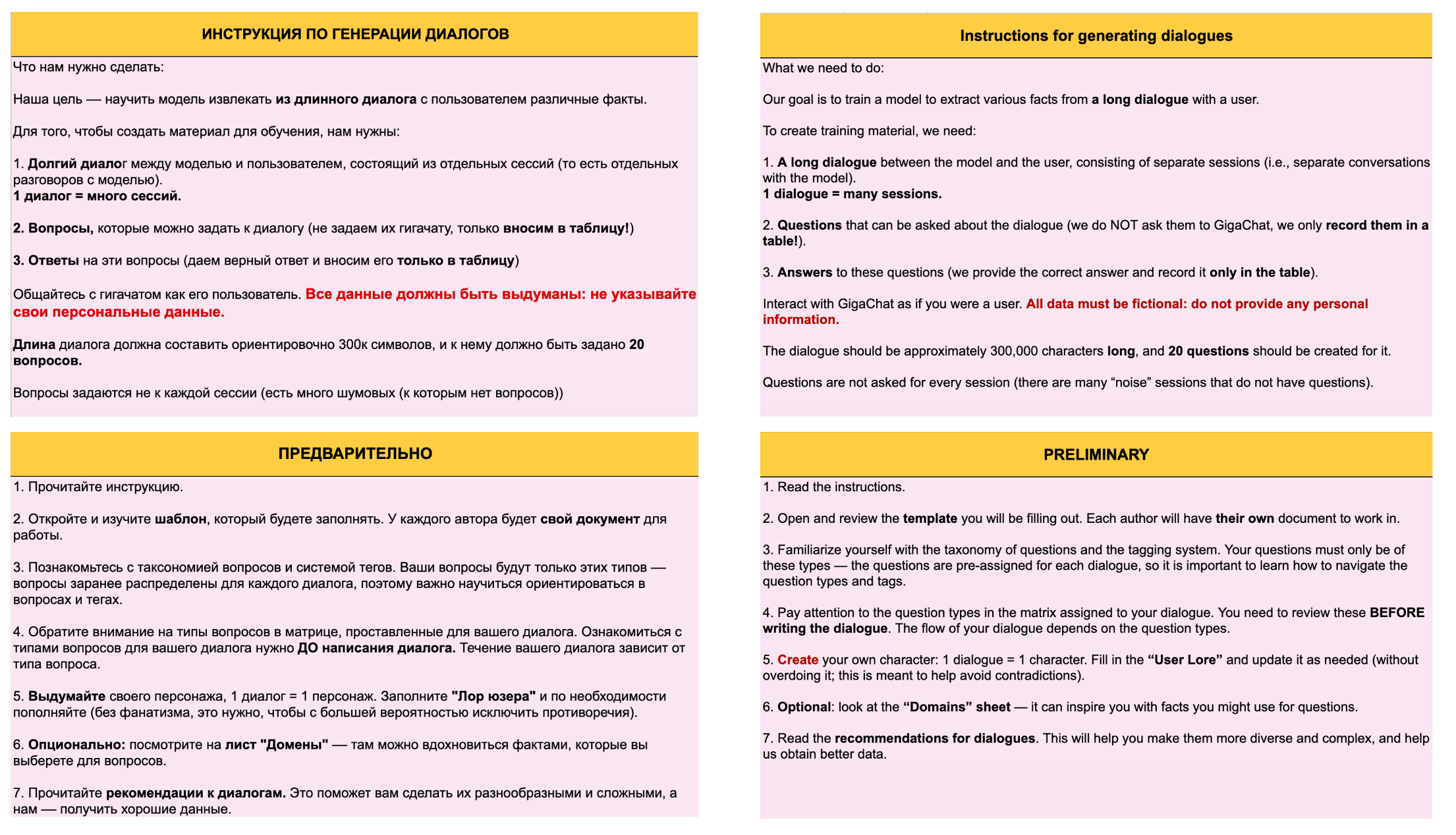}
\caption{\small Author instructions describing the required manuscript components and the requirement to avoid personal data in dialogues.}
\label{fig:guideline_authors}
\end{figure*}

\begin{figure*}
    \centering
    \includegraphics[width=\textwidth]{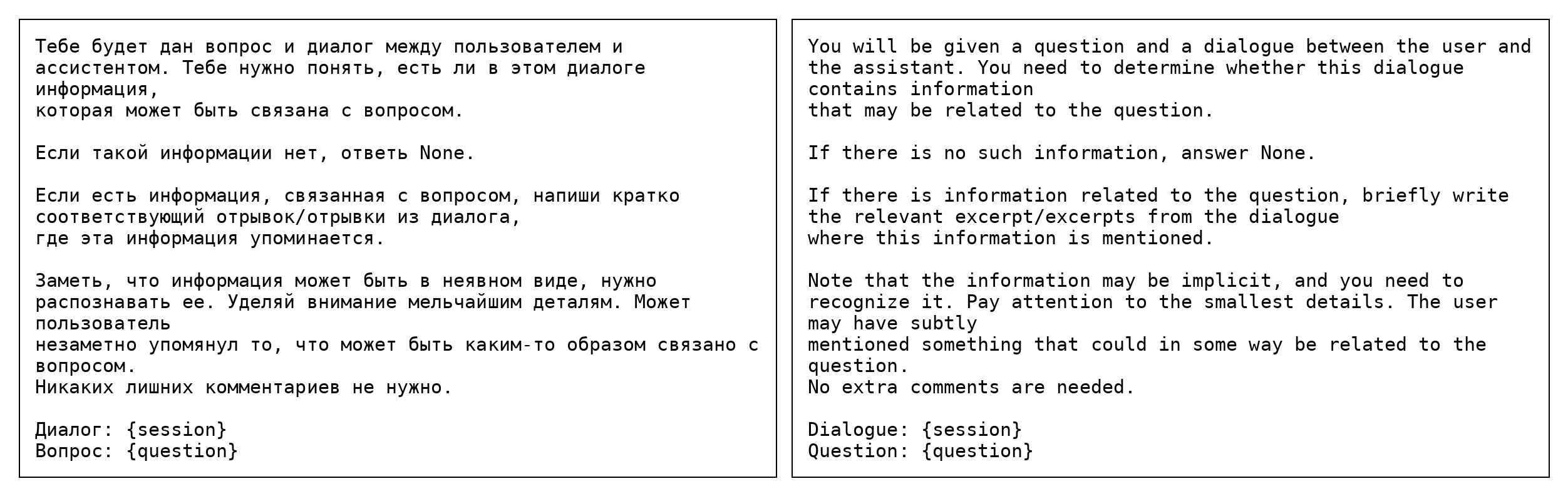}
    \caption{Evidence annotation prompt template used during validation of dialogue sessions and their associated questions. The original Russian prompt is shown together with an English translation provided for visualization in the paper.}
    \label{fig:evidence_annotation}
\end{figure*}

\begin{figure*}[t]
    \centering
    \scalebox{1.10}[0.85]{%
        \includegraphics[width=0.707\textwidth]{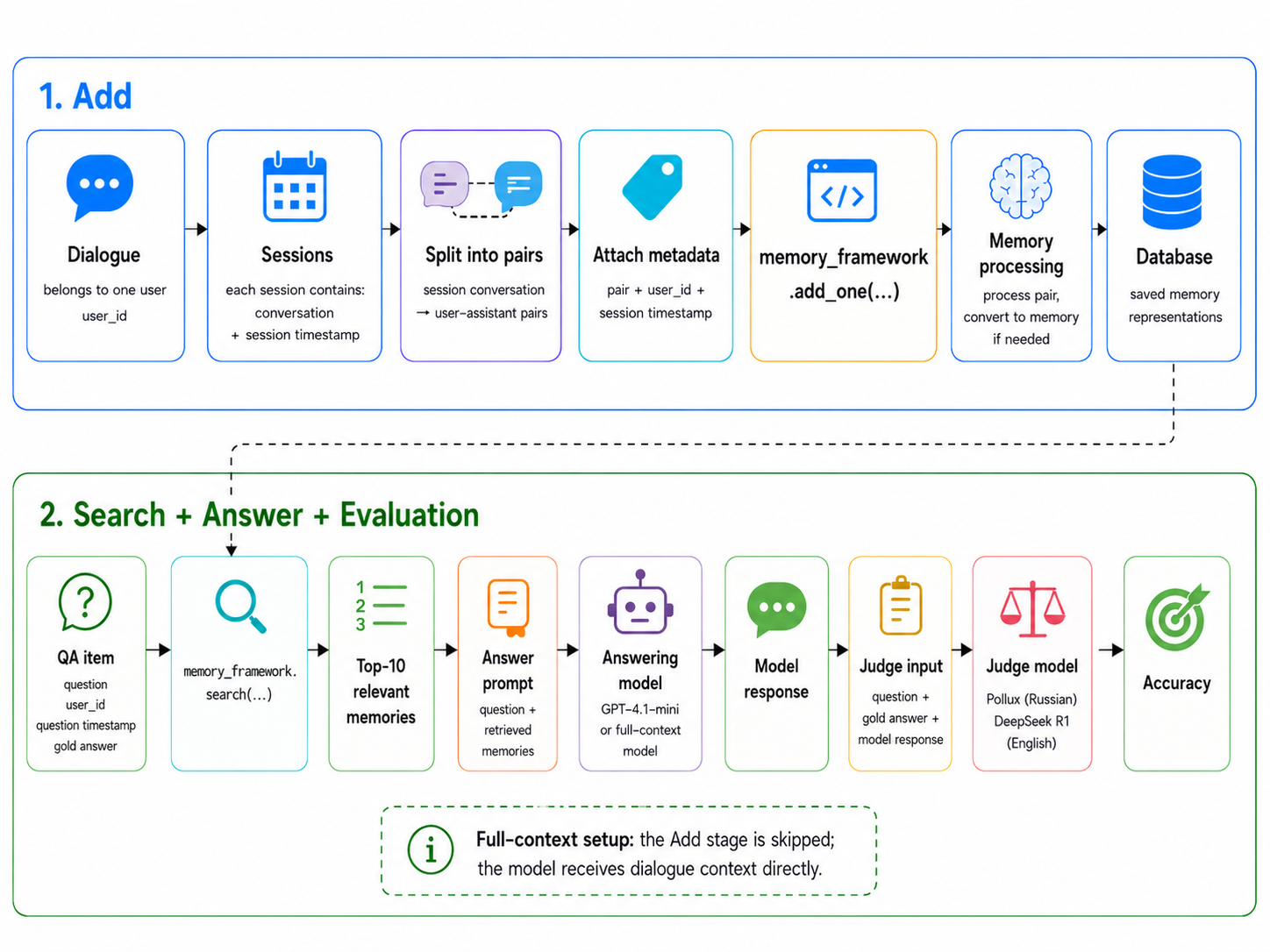}
    }
    \caption{RUMBA validation pipeline.}
    \label{fig:pipeline}
\end{figure*}

\begin{figure*}[p]
    \centering
    \includegraphics[width=\textwidth]{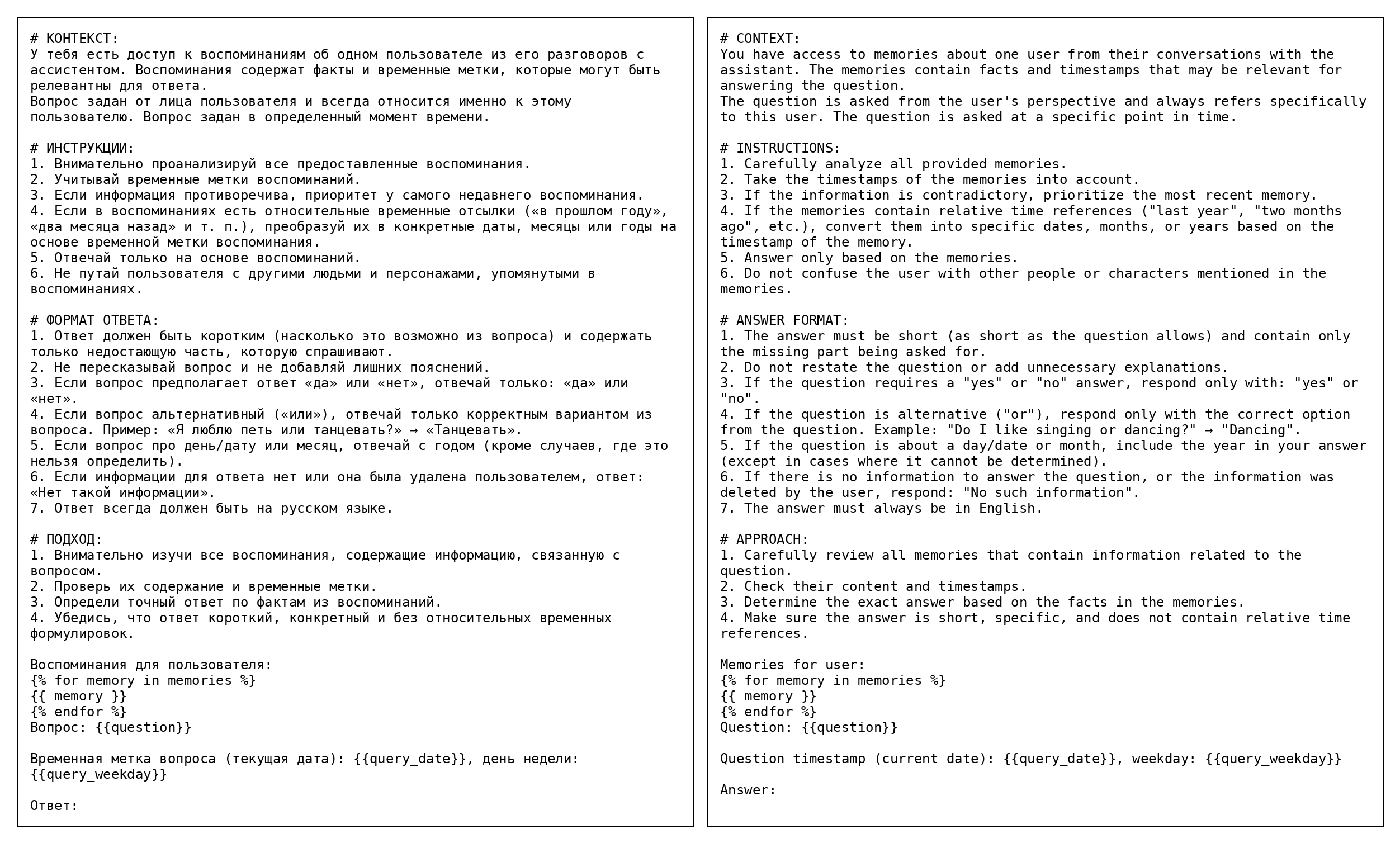}
    \caption{Answer-generation prompt templates used for Russian and English evaluation. The templates condition the retrieved memories (full dialogues in case of full context evaluation), memory timestamps, the benchmark question, and the question timestamp with its weekday.}
    \label{fig:answer_generation_prompts}
\end{figure*}

On average, the development (creation and validation) of a single dialogue and its corresponding set of questions required approximately 2 working days to complete.

To standardize the dialogue creation process, a matrix was developed containing the required question types for each dialogue. Authors used this matrix during dialogue construction, which ensured diversity of question types within each dialogue. 

\paragraph{Dialogue Validation Process.}
The validation process consisted of three stages. (1) Authors cross-validated dialogues, checking question adequacy, taxonomy alignment, and consistency of dialogue and question dates. Necessary corrections to questions and dates were introduced at this stage. (2) Questions and corresponding sessions were validated using an LLM (see the paragraph below \textit{Evidence annotation}). Based on LLM answers, validators — including the seven highest-performing authors and three independent validators — evaluated whether (a) the question matched its taxonomy category; (b) all dates were explicit and temporally consistent; (c) the answer was contained in the designated sessions and absent from unrelated ones; and (d) the dialogue was internally consistent and provided sufficient information to answer the question. Validators also revised questions and sessions when necessary and occasionally added new sessions. (3) Final validation was performed after all revisions. Each dialogue was reviewed by three validators who had not previously reviewed that dialogue.

Finally, the complete set of question–answer pairs, session dates, question dates, and taxonomy labels was independently validated by human annotators with an overlap of three.

\paragraph{Evidence annotation.}
We additionally constructed an utterance-level evidence annotation to identify where answer-relevant information appears in the dialogues. 
The input consisted of spreadsheet files with written dialogues, where each row contained a speaker role, an utterance, and a session-boundary marker, together with question-level metadata: the question, its reference answer, and its type. 
We first reconstructed session identifiers from the boundary markers and extracted the set of questions with their associated answers, types, and source sessions. 

Then, for each utterance--question pair, we queried \texttt{GigaChat-Max} and asked whether the utterance contained explicit or implicit information relevant to answering the question. 
If no relevant information was present, the model returned \texttt{None}; otherwise, it returned a concise excerpt or comment describing the relevant evidence.

The output was an augmented spreadsheet for each dialogue. 
It preserved the \texttt{session\_id}, speaker role, and utterance text, and added one evidence column per question. 
Each evidence column encoded the question, question type, reference answer, and associated source sessions in its header, while its cells contained the model-produced evidence for each utterance or an empty value when no evidence was detected. 
This produced a dense utterance-by-question evidence map for subsequent human inspection and diagnostic analysis.

The LLM validation prompt for GigaChat Max in Russian (and translated into English) is shown in Fig. \ref{fig:evidence_annotation}.

\subsection{Users characters} \label{subsec:characters_app}

\begin{figure*}[htbp]
\centering
\includegraphics[width=\textwidth]{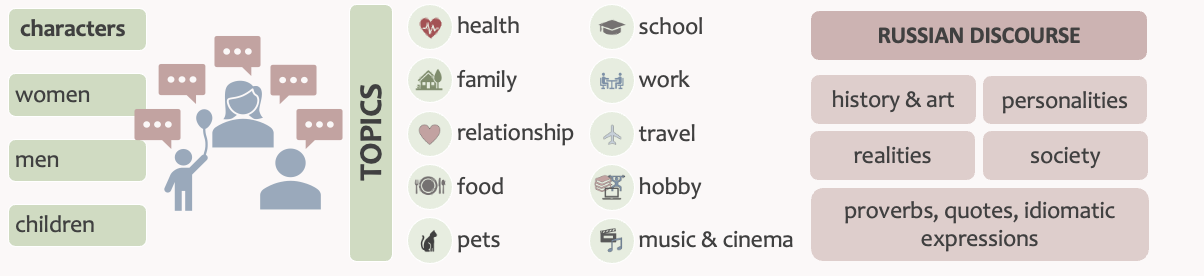}
\caption{High-level information about RUMBA dialogues. Dialogue characters include male and female adults as well as children. The top 10 conversation topics range from health to music and cinema. The dialogues also extensively reflect Russian discourse, including Russian history and art, social realities, notable figures, and broader societal characteristics.}\label{fig:characters}
\end{figure*}

\paragraph{Dialogues Characters.} Users characters represented in the dialogues are heterogeneous and multifaceted, and their interactions encompass a range of thematic domains, which enhances the realism of the constructed interactions and brings them closer to authentic user–assistant conversations. The behavioral scenarios assigned to these characters likewise reflect real-world user patterns, ranging from memory-related operations (such as storing, forgetting, and modifying information) to various strategies for expressing intent. These include explicit requests to memorize a fact, introducing a fact within a single utterance or anaphorically, and linking facts across different sessions, etc. Information about the characters and topics is presented in Fig.\ref{fig:characters}.

\paragraph{Sociolinguistic Portrait of the Characters.} The dialogues reflect sociocultural realities characteristic of Russia and Russian culture, as they include grammatical constructions typical of Russian, along with culture-specific references such as idioms, proverbs, folk sayings, anecdotes, and allusions to canonical and popular Russian literary and cultural works. Dialogues intentionally include spelling, grammar, lexical, stylistic errors to approximate authentic user-generated input.

\section{Translation}
\label{sec:translation}
Two translation granularities were used. Questions and answers were translated independently at the sentence level. Dialogue messages were translated at the session level, where the full sequence of messages within a session was provided to the model in a single request to preserve conversational context and message ordering. \texttt{gpt-4.1-mini}\footnote{\url{https://openai.com/index/gpt-4-1/}} was used as a translator model. 
Fig.~\ref{fig:translation_prompts} reports the system prompts and compact user-message schemas used for machine translation.

All remaining dataset fields were carried over unchanged, as they were either originally in English or language-independent.

\subsection{Translation validation details}
\label{sec:translation_validation}

The translation validation was carried out by specialists who possess relevant academic qualifications in the field (civil law contract workers and crowd workers (ABC Elementary\footnote{\url{https://elementary.center}} and Rambler\&Co  companies\footnote{\url{https://rambler-co.ru}}). 

\paragraph{Validation of translation quality: 10\% of the translated dataset.} Each evaluator received the original Russian text and its English translation for comparison and assessment of translation quality. Evaluators were provided with translation quality assessment criteria implemented in POLLUX~\cite {martynov2025eyejudgementdissectingevaluation}. The names of the criteria, their descriptions, and the scoring system are presented in Table \ref{tab:translation_criteria}. 

The average time required to complete the evaluation of a single RU-EN pair of utterances was approximately 3 minutes.

\paragraph{Question-Answer-Evidence Sessions Validation.} Experts evaluated a triplet: the evidence sessions and associated question–answer pairs. At this stage, it was essential to ensure that: (1) the translated question preserves the meaning and intent of the original question; (2) the translated gold answer to the question is consistent with the original answer; (3) the translated gold answer corresponds to the translated evidence session, i.e., the session indeed contains the information required to answer the question; and (4) the translated question–answer–session triplet is free of errors and contradictions and remains logically and semantically coherent to the same extent as the original Russian version. All issues were corrected.

The average time required to complete the evaluation of the triplet was approximately 5 minutes.

Non-evidence sessions (irrelevant to questions, filler sessions) were not manually validated by experts. For these sessions, we assume a translation quality score of 0.88, which was obtained through 10\%-dataset validation.

The evaluators received compensation for each completed task in accordance with the terms stipulated in their employment contracts. The validation procedure was carried out using the ABC Elementary \footnote{\url{https://elementary.center/}} platform.

\begin{table*}
\small
\centering
\caption{Translation Evaluation Criteria, their descriptions, and the scoring system}
\label{tab:translation_criteria}
\renewcommand{\arraystretch}{1.3}%
\begin{tabular}{>{\raggedright\arraybackslash}p{2cm}p{7.5cm}p{6.5cm}@{}}
\toprule
\textbf{Criteria name} & \textbf{Description} & \textbf{Scores} \\
\midrule
\multicolumn{3}{l}{Critical criteria} \\
\midrule
(a) Format violation or (b) Censor block & (a) This criterion evaluates whether it is necessary to further grade the LLM’s output based on other criteria or the LLM’s output is unreadable and cannot be graded.
\par(b) The criterion evaluates whether this is the LLM's output or there is just a placeholder with the text "Unfortunately, generative models cannot discuss such topics" or similar standard text where the LLM avoids answering. &  \textbf{(a) 0:} We read the model’s response and it can be further evaluated based on other criteria. 
\par\textbf{1:} The model’s response shows excessive looping, or there are many artifacts, or the model’s response is entirely in a different language, or the model simply copied the user’s query. It does not make sense to further evaluate the model’s response. 
\par\textbf{(b) 0:} The censor did not trigger, the LLM's output can be evaluated further based on other criteria. 
\par \textbf{1:} The censor triggered, there is nothing to evaluate. \\
\midrule
\multicolumn{3}{l}{Task-specific criteria} \\
\midrule
Original goal & This criterion evaluates whether the communicative purpose of the source text is preserved in the translation: for example, to inform, advertise, accuse, tease, and so on. & \textbf{0:} In the translation, the source text intent is completely lost. 
\par \textbf{1: (a)} The source text intent is fully preserved in the translation. \textbf{(b)} In the translation, the source text intent is preserved but slightly blurred (for example, the aim was to teasingly poke fun at the reader, but only the humorous effect remains without the mockery). \\
Original tone & This criterion evaluates whether the tone of the source text is preserved in the translation. It refers to that part of the expressive plan which consists of the connotatively marked elements of the source text. If events are described with irony, the translator should (at least partially) retain this irony. If the information in the source text is presented neutrally, the translator should convey the same tone. Conversely, if the tone of the source text is not neutral, the translation should (at least partially) convey this as well.(Source: S. V. Tyulenev “Translation Theory”) & \textbf{0: }The emotional tone of the text was completely lost in the translation. 
\par \textbf{1: (a)} The LLM completely preserved the tone of
the source text. \textbf{(b)} In general, the translation preserved the tone of the source text, but it is somewhat diluted; it could have been handled better. \\
Author viewpoint & This criterion evaluates whether the translation distorts the author’s position. The translation should not contradict the author’s position. The author’s position may include the author’s point of view (for example, they may be subtly criticizing a product between the lines); ideological stance (often, the ideological position can be inferred
from indirect signs); degree of confidence/uncertainty (if the author speaks about something uncertainly, it should not become a statement in the translation), etc. & \textbf{0:} The author’s position was completely distorted during the translation. 
\par \textbf{1: (a)} The author’s position is completely preserved, and the LLM conveyed the subtext, if there was any. \textbf{(b)} The author’s position is generally preserved, but there are some issues (for example,the LLM overlooked certain veiled messages). \\
Compliance with functional style & This criterion evaluates whether the functional style is preserved in the translation. When translating, the text should maintain its functional style: literary, journalistic, official business, scientific, or colloquial-everyday, including the linguistic features of each style. & \textbf{0: }The LLM completely changed the functional
style. 
\par \textbf{1: (a) }The translated text fully corresponds to the
functional style in which the source text is written. \textbf{(b)} The LLM retained the functional style, but some linguistic features not typical for this functional style appeared in the translation.\\
Language norms & This criterion evaluates the compliance with lexical, grammatical, syntactic and stylistic norms of the target language. The translation should “sound natural”, meaning it should contain appropriate vocabulary and grammar, syntax, and stylistic devices characteristic of the target language. & \textbf{0: }The LLM made 3 or more errors in compliance with lexical, grammatical, syntactic and stylistic norms of the target language.
\par \textbf{1: (a)} The LLM made no errors in compliance with lexical, grammatical, syntactic and stylistic norms of the target language. \textbf{(b)} The LLM made one or two errors in compliance with lexical, grammatical, syntactic and stylistic norms of the target language. \\
Factual accuracy & This criterion evaluates the accuracy of conveying the factual information presented in the source text. The LLM should not distort factual (precision) information from the source text. An exception is made for factual information that would definitely not be understood by a native speaker of the target language: in such cases, the factual information should be adapted to convey the tone of the source text and
the author’s intent. & \textbf{0: }The LLM made factual errors, incorrectly conveying precise information from the source text. 
\par\textbf{1: (a) }The LLM conveyed the facts and their related context/subtext completely accurately. \textbf{(b)} The LLM did not make any factual errors but made a slip related, for example, to the subtext concerning some phenomenon or event.\\
\bottomrule
\end{tabular}
\end{table*}

\begin{table*}
\centering
\small
\caption{Criterion-specific scores obtained from human evaluation of translation quality.}
\label{tab:translation_quality}
\begin{tabular}{@{}p{6cm}p{1.2cm}@{}}
\toprule
\textbf{Criteria} & \textbf{Value} \\
\midrule
Critical criteria (Format violation or Censor block) & 0.99 \\
\midrule
Task-specific criteria \\
\midrule
Original goal & 0.99 \\
Original tone & 0.98 \\
Author viewpoint & 0.99\\
Compliance with functional style & 0.99 \\
Language norms & 0.9 \\
Factual accuracy & 0.97 \\
\midrule
General score & 0.88 \\
\bottomrule
\end{tabular}
\end{table*}

\begin{figure*}[t]
\centering
\begin{minipage}{0.72\textwidth}
\begin{PromptBlock}[
  fontsize=\fontsize{5.5pt}{6.4pt}\selectfont,
  baselinestretch=0.95,
  framesep=2pt
]
QA-pair translation

SYSTEM:
You are a professional translator. Translate Russian to natural English.
You are given a question-answer pair. Translate both together.
Rules:
- Preserve exact meaning and the relationship between question and answer.
- Keep punctuation and sentence type where possible.
- Preserve grammatical cues affecting meaning (gender, number, etc.).
- Do not omit, generalize, or add information.
- Keep names, titles, and quoted text consistent.
- If the answer is an alphabetical list, reorder it to be correct in English while preserving the items.

USER:
{"question": question_ru, "answer": answer_ru}


Dialogue-session translation

SYSTEM:
You are a professional translator. Translate Russian to natural English.
This is a conversation session; messages are context-dependent.
Rules:
- Preserve messages order and meaning exactly.
- Preserve formatting inside each message text: keep newlines, markdown, bullets, code blocks, etc - if any.
- Translate ONLY the message text; do not change dia_id.
- Output must match the provided schema.

USER:
{"dia_id": dia_id, "messages": [{"speaker": speaker, "text": text_ru}, ...]}
\end{PromptBlock}
\end{minipage}
\caption{Translation message templates used to construct the English split. System prompts are shown verbatim; user messages are shown as compact payload schemas.}
\label{fig:translation_prompts}
\end{figure*}

\begin{figure*}[ht!]
    \centering
    \includegraphics[width=\textwidth]{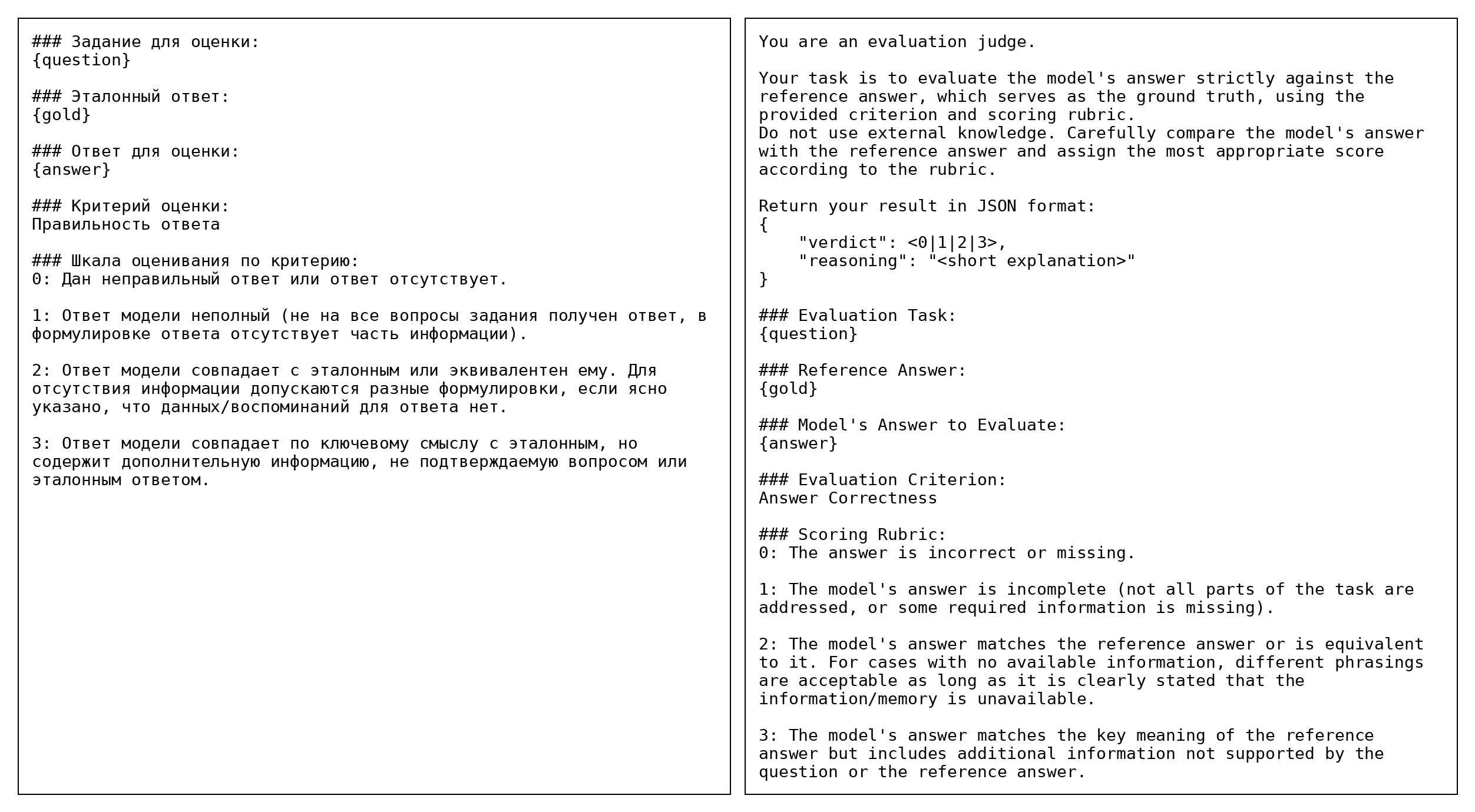}
    \caption{LLM-as-Judge prompt templates used for Russian and English evaluation. Placeholders indicate the inserted benchmark question, reference answer, and model answer.}
    \label{fig:judge_prompts}
\end{figure*}

\section{Evaluation details}
\label{sec:evaluation_details}

Fig.~\ref{fig:pipeline} shows the RUMBA evaluation pipeline. 
Language-specific answer-generation prompts are provided in Fig.~\ref{fig:answer_generation_prompts}.

\section{Judge details}
\label{sec:judge}

\subsection{English judge choice}
\label{sec:judge_сhoice}

Our goal was to select an English judge that could be used with the same rubric as \texttt{POLLUX}, so that the English scores would be produced under an evaluation procedure as close as possible to the Russian one. We based the selection on JudgeBench~\cite{tan2025judgebench} and its Hugging Face leaderboard\footnote{\url{https://huggingface.co/spaces/ScalerLab/JudgeBench}}. JudgeLRM additionally reports 78.67\% agreement between \texttt{DeepSeek-R1} and human judgments on PandaLM~\cite{chen2025judgelrm}. Higher-ranked alternatives were not suitable for this setup: several fine-tuned judge or reward models are designed for pairwise preference comparison rather than scoring a single answer against a reference; Prometheus-style judges rely on their own scoring scale; and prompted GPT/o-series judges were avoided because GPT-family models are included among our answer-generation baselines. We therefore selected \texttt{DeepSeek-R1} as a strong non-GPT prompted judge compatible with our rubric. We found no single judge with reported human alignment for both languages under comparable conditions and thus used language-specific judges.

\subsection{Judge prompts and verdicts}
\label{sec:judge_promts}

To obtain comparable scores for the Russian and English subsets, we used the same \texttt{POLLUX}-style evaluation format in both languages. 
The only difference is that \texttt{POLLUX} was evaluated with the prompt template and response format it was trained for, while \texttt{DeepSeek-R1}, being a general-purpose model, required a short task description and a JSON output format. Full templates are presented in Fig.~\ref{fig:judge_prompts}. 

The judge assigned one of 4 rubric labels. This rubric set was selected empirically. Labels {0--2} correspond to the standard \texttt{POLLUX} rubrics for the \textit{Answer Correctness} criterion: incorrect or missing answer, incomplete answer, and answer matching or equivalent to the reference answer. 
We added label {3} for a frequent borderline case in our benchmark: the model answers the question correctly in terms of the key meaning, but adds extra details that are not supported by the question or the reference answer and would require checking the original dialogue. 
For the final binary accuracy score, labels 0 and 1 were treated as incorrect, while labels 2 and 3 were treated as correct.

\subsection{Qualitative Analysis}
\begin{table}[!htbp]
\centering
\scriptsize
\setlength{\tabcolsep}{3pt}
\renewcommand{\arraystretch}{1.15}
\begin{tabularx}{\columnwidth}{@{}Xcc@{}}
\toprule
\textbf{Judge error type} & \textbf{POLLUX} & \textbf{DeepSeek-R1} \\
\midrule
Correct answer judged as over-informative & 2 (2.6\%) & 16 (16.7\%) \\
Correct answer judged as wrong & 2 (2.6\%) & 1 (1.0\%) \\
Correct answer judged as incomplete & 19 (24.7\%) & 9 (9.4\%) \\
Over-informative answer judged as correct & 8 (10.4\%) & 3 (3.1\%) \\
Over-informative answer judged as wrong & 3 (3.9\%) & 0 (0.0\%) \\
Over-informative answer judged as incomplete & 13 (16.9\%) & 1 (1.0\%) \\
Wrong answer judged as correct & 2 (2.6\%) & 21 (21.9\%) \\
Wrong answer judged as over-informative & 19 (24.7\%) & 11 (11.5\%) \\
Wrong answer judged as incomplete & 9 (11.7\%) & 30 (31.2\%) \\
Incomplete answer judged as correct & 0 (0.0\%) & 3 (3.1\%) \\
Incomplete answer judged as over-informative & 0 (0.0\%) & 1 (1.0\%) \\
\midrule
\textbf{Total} & \textbf{77} & \textbf{96} \\
\bottomrule
\end{tabularx}
\caption{\small Manual qualitative classification of LLM judge errors for POLLUX and DeepSeek-R1. The analyzed errors account for 5.0\% and 6.2\% of all 1,543 benchmark questions, respectively; row percentages are computed within each judge's error set.}
\label{tab:judge_error_types}
\end{table}

In addition to the aggregate agreement metrics, we conducted a manual qualitative analysis of judge errors. The goal of this analysis was to identify recurring failure modes of the LLM-as-a-judge setup and to better understand how these failures may affect the interpretation of the benchmark results.

A linguist manually inspected cases in which the automatic judge verdict did not match the human annotation and assigned each case to an error category. The annotation focused on the direction of the judge error: whether the judge underestimated the answer quality, overestimated it, or made a score-level distinction that did not affect the final binary correctness decision.

Table~\ref{tab:judge_error_types} summarizes the resulting error taxonomy. For \texttt{POLLUX}, the analysis includes 77 judge error cases, corresponding to 5.0\% of all questions. For \texttt{DeepSeek-R1}, the analysis includes 96 judge error cases after filtering out cases that were excluded from the qualitative analysis, corresponding to 6.2\% error rate.

The qualitative analysis shows that the two judges have different error profiles. \texttt{POLLUX} tends to be more conservative: among binary-relevant errors, 37 cases correspond to underestimation of answer quality, while 21 cases correspond to overestimation. Thus, 63.8\% of \texttt{POLLUX} binary-relevant errors are underestimations. The most frequent \texttt{POLLUX} error types are judging correct answers as incomplete and judging wrong answers as over-informative, each accounting for 19 cases. Another frequent failure mode is treating over-informative answers as incomplete, which occurred in 13 cases.

\texttt{DeepSeek-R1} shows the opposite tendency. Among binary-relevant errors, 11 cases correspond to underestimation, whereas 36 cases correspond to overestimation. Thus, 76.6\% of \texttt{DeepSeek-R1} binary-relevant errors are overestimations. The largest category is wrong answers judged as incomplete, with 30 cases. Although this category does not necessarily change the final binary decision if both wrong and incomplete answers are mapped to incorrect, it indicates that the judge often recognizes that the answer is imperfect but fails to identify that the semantic content is fundamentally wrong. More importantly, \texttt{DeepSeek-R1} also frequently assigns positive judgments to incorrect answers: 21 wrong answers were judged as correct and 11 wrong answers were judged as over-informative.

We further grouped reasoning-level failures into broader categories. For both judges, the dominant source of error is incorrect identification of the semantic core of the answer. This accounts for 50 cases for \texttt{POLLUX} and 87 cases for \texttt{DeepSeek-R1}. In these cases, the judge fails to correctly determine whether the model answer preserves the key meaning of the reference answer. \texttt{POLLUX} also shows a relatively larger share of cases where the judge reasoning is mostly plausible but the final verdict is inconsistent with that reasoning. This occurred in 21 cases. In contrast, \texttt{DeepSeek-R1} rarely distorts the rubric explicitly, but more often fails at semantic equivalence judgments.

These findings suggest that the judge errors are not random. The Russian judge is biased toward stricter scoring and tends to penalize answers for incompleteness or unsupported extra details. The English judge is more permissive and more often overestimates semantically incorrect answers. 

\section{Detailed Slice-Level Result Analysis}
\label{app:detailed_analysis}

This appendix provides the detailed slice-level analysis underlying the summary in Section~\ref{sec:results}. The purpose of this analysis is to use the full set of benchmark annotations to obtain more fine-grained insights into method behavior. These annotations capture different properties of memory-oriented question answering, including the semantic operation required by the question, the distribution of supporting evidence across sessions, and the need for temporal grounding.

Each question belongs to three main axes: semantic type, single-session versus multi-session evidence, and temporal versus atemporal setting. We additionally analyze temporal-expression explicitness, language effects, evidence-position effects, and semantic supergroups. Unless stated otherwise, all analyses use LLM-as-Judge accuracy as the primary metric.

Let $y_{m,q}^{\ell} \in \{0,1\}$ denote the binary correctness score of method $m$ on question $q$ in language $\ell$. For a method family $F$, we define the per-question family score as
\[
s_{F,q}^{\ell} = \frac{1}{|F|}\sum_{m \in F} y_{m,q}^{\ell}.
\]
For a benchmark slice $A$ with question set $Q_A$, the slice-level family score is
\[
\bar{s}_{F,A}^{\ell} = \frac{1}{|Q_A|}\sum_{q \in Q_A} s_{F,q}^{\ell}.
\]
Family-level comparisons within a slice are computed as
\[
\Delta_{\mathrm{FC}-\mathrm{Agent/RAG},A}^{\ell}
=
\bar{s}_{\mathrm{FC},A}^{\ell}
-
\bar{s}_{\mathrm{Agent/RAG},A}^{\ell},
\]
whereas slice-difficulty contrasts within a family compare two slices $A$ and $B$ as
\[
\Delta_{A-B,F}^{\ell}
=
\bar{s}_{F,A}^{\ell}
-
\bar{s}_{F,B}^{\ell}.
\]
Thus, slices of different sizes are compared through their own mean per-question scores rather than raw counts. Confidence intervals are estimated by bootstrap resampling over questions; $p$-values are Holm-corrected over planned comparisons.

Slice-level results are reported separately for Agent/RAG and full-context systems, and all analyses are reported independently for Russian and English. We also provide descriptive method-level results to compare individual baselines within and across the two method families.
For temporal, temporal-expression, and scope--temporal family aggregates, the Agent/RAG family excludes \texttt{mem0g}, whose graph-based setup is atemporal; all other aggregates retain the full family.

All code used to produce slice‑level analyses, statistical tests, tables, and figures is open source.

\subsection{Full Context vs Agent/RAG as Method Families}

Table~\ref{tab:app_family_all_slices} reports the \textit{non-direct family-level comparison} across the main benchmark slices, without matching the answer model.
Full-context scores significantly exceed Agent/RAG scores in all listed slices for both languages. The overall gap is larger in English ($+15.11$ accuracy points) than in Russian ($+11.86$ accuracy points). The difference between these two gaps is also significant: $(\Delta_{\mathrm{EN}}-\Delta_{\mathrm{RU}})=+3.25$ accuracy points, 95\% CI $[1.82, 4.72]$. The matched \texttt{gpt-4.1-mini} comparison in Table~\ref{tab:main_family_results_compact} yields smaller overall gaps.

\begin{table*}[t]
\centering
\scriptsize
\begin{tabular}{llrrrrr}
\toprule
\textbf{Slice} & \textbf{Lang.} & \textbf{Full context} & \textbf{Agent/RAG} & \textbf{$\Delta$} & \textbf{95\% CI} & \textbf{Holm $p$} \\
\midrule
Overall & RU & 67.66 & 55.80 & +11.86 & [10.56, 13.17] & <0.001 \\
Overall & EN & 69.40 & 54.29 & +15.11 & [13.80, 16.42] & <0.001 \\
\midrule
Single & RU & 76.65 & 63.92 & +12.73 & [11.12, 14.36] & <0.001 \\
Single & EN & 78.21 & 62.92 & +15.29 & [13.66, 16.91] & <0.001 \\
Multi & RU & 53.69 & 43.18 & +10.51 & [8.27, 12.75] & <0.001 \\
Multi & EN & 55.70 & 40.87 & +14.83 & [12.61, 17.10] & <0.001 \\
\midrule
Atemporal & RU & 69.39 & 60.55 & +8.84 & [7.27, 10.39] & <0.001 \\
Atemporal & EN & 70.91 & 60.02 & +10.89 & [9.32, 12.40] & <0.001 \\
Temporal & RU & 62.96 & 58.41 & +4.55 & [1.77, 7.32] & 0.001 \\
Temporal & EN & 65.30 & 52.87 & +12.43 & [9.54, 15.28] & <0.001 \\
\midrule
Single + atemporal & RU & 78.61 & 69.00 & +9.62 & [7.77, 11.50] & <0.001 \\
Single + atemporal & EN & 80.11 & 70.03 & +10.08 & [8.18, 11.91] & <0.001 \\
Multi + atemporal & RU & 55.49 & 47.82 & +7.67 & [4.91, 10.39] & <0.001 \\
Multi + atemporal & EN & 57.05 & 44.93 & +12.11 & [9.40, 14.80] & <0.001 \\
Single + temporal & RU & 71.54 & 67.20 & +4.34 & [0.83, 7.93] & 0.027 \\
Single + temporal & EN & 73.29 & 60.38 & +12.91 & [9.28, 16.53] & <0.001 \\
Multi + temporal & RU & 48.42 & 43.51 & +4.92 & [0.56, 9.30] & 0.029 \\
Multi + temporal & EN & 51.76 & 40.13 & +11.63 & [6.75, 16.57] & <0.001 \\
\bottomrule
\end{tabular}
\caption{\small Family-level comparison between Full context and Agent/RAG across the main benchmark slices. Scores are family-level means of binary LLM-as-judge correctness; scores and differences are reported on a 0--100 accuracy-point scale, with $\Delta = \text{Full context} - \text{Agent/RAG}$. Agent/RAG temporal and scope--temporal aggregates exclude \texttt{mem0g}.}
\label{tab:app_family_all_slices}
\end{table*}

\subsection{Single-Session vs Multi-Session Difficulty}

Table~\ref{tab:app_scope_difficulty} shows that multi-session questions are consistently harder than single-session questions. The single-to-multi degradation ranges from $+20.73$ to $+22.96$ accuracy points across languages and families. This makes session multiplicity the strongest difficulty axis in the benchmark.

Importantly, the multi-session penalty is not specific to retrieval-based systems. Full-context systems also show a large degradation from single-session to multi-session questions. Additional interaction tests show that the degradation difference between full-context and Agent/RAG families is not statistically significant: in Russian, the difference is $+2.23$ accuracy points with 95\% CI $[-0.69, 5.13]$; in English, it is $+0.46$ accuracy points with 95\% CI $[-2.42, 3.42]$.

\begin{table*}[t]
\centering
\small
\begin{tabular}{llrrrrr}
\toprule
\textbf{Family} & \textbf{Lang.} & \textbf{Single} & \textbf{Multi} & \textbf{$\Delta$} & \textbf{95\% CI} & \textbf{Holm $p$} \\
\midrule
Agent/RAG & RU & 63.92 & 43.18 & +20.73 & [17.71, 23.83] & <0.001 \\
Agent/RAG & EN & 62.92 & 40.87 & +22.06 & [19.02, 25.14] & <0.001 \\
Full context & RU & 76.65 & 53.69 & +22.96 & [20.07, 25.88] & <0.001 \\
Full context & EN & 78.21 & 55.70 & +22.51 & [19.57, 25.50] & <0.001 \\
\bottomrule
\end{tabular}
\caption{\small Single-session versus multi-session difficulty within method families. Scores are family-level means of binary LLM-as-judge correctness; scores and differences are reported on a 0--100 accuracy-point scale, with $\Delta = \text{Single} - \text{Multi}$.}\label{tab:app_scope_difficulty}
\end{table*}

\subsection{Temporal vs Atemporal Difficulty}

Table~\ref{tab:app_temporal_difficulty} shows that temporal questions are harder on average, but the effect is smaller than the single-to-multi degradation. The atemporal-to-temporal gap ranges from $+2.14$ to $+7.15$ accuracy points. The largest temporal penalty appears for English Agent/RAG systems, while the Russian Agent/RAG contrast is not significant after correction.

In English, the Full-context--Agent/RAG gap is $+12.43$ accuracy points on temporal questions and $+10.89$ on atemporal questions. In Russian, the corresponding gaps are $+4.55$ and $+8.84$, respectively.

\begin{table*}[t]
\centering
\small
\begin{tabular}{llrrrrr}
\toprule
\textbf{Family} & \textbf{Lang.} & \textbf{Atemporal} & \textbf{Temporal} & \textbf{$\Delta$} & \textbf{95\% CI} & \textbf{Holm $p$} \\
\midrule
Agent/RAG & RU & 60.55 & 58.41 & +2.14 & [-1.55, 5.74] & 0.259 \\
Agent/RAG & EN & 60.02 & 52.87 & +7.15 & [3.54, 10.70] & <0.001 \\
Full context & RU & 69.39 & 62.96 & +6.43 & [3.10, 9.67] & <0.001 \\
Full context & EN & 70.91 & 65.30 & +5.61 & [2.23, 8.93] & 0.004 \\
\bottomrule
\end{tabular}
\caption{\small Atemporal versus temporal difficulty within method families. Scores are family-level means of binary LLM-as-judge correctness; scores and differences are reported on a 0--100 accuracy-point scale, with $\Delta = \text{Atemporal} - \text{Temporal}$. Agent/RAG aggregates exclude \texttt{mem0g}.}
\label{tab:app_temporal_difficulty}
\end{table*}

\subsection{Joint Scope--Temporal Intersections}

Table~\ref{tab:app_scope_temporal_intersections} reports the four intersections of the scope and temporal axes. The hardest setting is the multi-session + temporal intersection. The multi-session penalty is significant within both atemporal and temporal questions. Within single-session questions, the temporal penalty is significant for English Agent/RAG and both full-context languages, but not for Russian Agent/RAG. Within multi-session questions, it is not significant for Agent/RAG in either language or for English full context after correction.

\begin{table*}[t]
\centering
\small
\begin{tabular}{llrrrr}
\toprule
\textbf{Family} & \textbf{Lang.} & \textbf{Single + Atemporal} & \textbf{Multi + Atemporal} & \textbf{Single + Temporal} & \textbf{Multi + Temporal} \\
\midrule
Agent/RAG & RU & 69.00 & 47.82 & 67.20 & 43.51 \\
Agent/RAG & EN & 70.03 & 44.93 & 60.38 & 40.13 \\
Full context & RU & 78.61 & 55.49 & 71.54 & 48.42 \\
Full context & EN & 80.11 & 57.05 & 73.29 & 51.76 \\
\bottomrule
\end{tabular}
\caption{\small Family-level accuracy on the 4 scope--temporal intersections. Scores are family-level means of binary LLM-as-judge correctness and are reported on a 0--100 accuracy-point scale. The table shows how session scope and temporal reasoning jointly affect performance within each method family and language. Agent/RAG aggregates exclude \texttt{mem0g}.}\label{tab:app_scope_temporal_intersections}
\end{table*}

\begin{table*}[t]
\centering
\scriptsize
\begin{tabular}{llrrrrr}
\toprule
\textbf{Contrast} & \textbf{Family / Lang.} & \textbf{A} & \textbf{B} & \textbf{$\Delta$} & \textbf{95\% CI} & \textbf{Holm $p$} \\
\midrule
Single+Atemporal $-$ Multi+Atemporal & Agent/RAG RU & 69.00 & 47.82 & +21.17 & [17.18, 25.03] & <0.001 \\
Single+Atemporal $-$ Multi+Atemporal & Agent/RAG EN & 70.03 & 44.93 & +25.10 & [21.13, 28.97] & <0.001 \\
Single+Atemporal $-$ Multi+Atemporal & Full context RU & 78.61 & 55.49 & +23.12 & [19.62, 26.42] & <0.001 \\
Single+Atemporal $-$ Multi+Atemporal & Full context EN & 80.11 & 57.05 & +23.06 & [19.53, 26.54] & <0.001 \\
\midrule
Single+Temporal $-$ Multi+Temporal & Agent/RAG RU & 67.20 & 43.51 & +23.70 & [17.67, 29.77] & <0.001 \\
Single+Temporal $-$ Multi+Temporal & Agent/RAG EN & 60.38 & 40.13 & +20.25 & [14.10, 26.33] & <0.001 \\
Single+Temporal $-$ Multi+Temporal & Full context RU & 71.54 & 48.42 & +23.12 & [17.50, 28.64] & <0.001 \\
Single+Temporal $-$ Multi+Temporal & Full context EN & 73.29 & 51.76 & +21.53 & [15.69, 27.33] & <0.001 \\
\midrule
Single+Atemporal $-$ Single+Temporal & Agent/RAG RU & 69.00 & 67.20 & +1.79 & [-2.30, 6.02] & 0.406 \\
Single+Atemporal $-$ Single+Temporal & Agent/RAG EN & 70.03 & 60.38 & +9.65 & [5.33, 14.05] & <0.001 \\
Single+Atemporal $-$ Single+Temporal & Full context RU & 78.61 & 71.54 & +7.08 & [3.32, 10.67] & <0.001 \\
Single+Atemporal $-$ Single+Temporal & Full context EN & 80.11 & 73.29 & +6.82 & [3.29, 10.47] & <0.001 \\
\midrule
Multi+Atemporal $-$ Multi+Temporal & Agent/RAG RU & 47.82 & 43.51 & +4.32 & [-1.58, 10.15] & 0.207 \\
Multi+Atemporal $-$ Multi+Temporal & Agent/RAG EN & 44.93 & 40.13 & +4.80 & [-0.96, 10.57] & 0.207 \\
Multi+Atemporal $-$ Multi+Temporal & Full context RU & 55.49 & 48.42 & +7.07 & [1.61, 12.38] & 0.045 \\
Multi+Atemporal $-$ Multi+Temporal & Full context EN & 57.05 & 51.76 & +5.29 & [-0.46, 10.84] & 0.206 \\
\bottomrule
\end{tabular}
\caption{\small Within-family contrasts over the scope--temporal intersections. Scores are family-level means of binary LLM-as-judge correctness; scores and differences are reported on a 0--100 accuracy-point scale, with $\Delta$ computed as the first slice minus the second slice in each contrast. Agent/RAG aggregates exclude \texttt{mem0g}.}\label{tab:app_scope_temporal_contrasts}
\end{table*}

\subsection{Temporal-Expression Explicitness}

Table~\ref{tab:app_temporal_expression_fc_vs_agent} first reports the Full context versus Agent/RAG comparison (in a non-answerer-fixed setting) within each temporal expression tag. The gap is significant for explicit expressions in both languages and for all English tags; the Russian implicit and no-expression gaps are small and not significant after correction.

\begin{table*}[t]
\centering
\scriptsize
\begin{tabular}{llrrrrr}
\toprule
\textbf{Temporal-expression tag} & \textbf{Lang.} & \textbf{Full context} & \textbf{Agent/RAG} & \textbf{$\Delta$} & \textbf{95\% CI} & \textbf{Holm $p$} \\
\midrule
Explicit & RU & 76.19 & 63.33 & +12.86 & [8.07, 17.79] & <0.001 \\
Explicit & EN & 76.89 & 60.20 & +16.69 & [11.93, 21.60] & <0.001 \\
Implicit & RU & 46.17 & 44.35 & +1.82 & [-3.60, 7.16] & 0.692 \\
Implicit & EN & 47.00 & 38.55 & +8.45 & [1.41, 15.40] & 0.017 \\
No temporal expression & RU & 62.18 & 60.33 & +1.85 & [-1.89, 5.70] & 0.692 \\
No temporal expression & EN & 65.63 & 53.85 & +11.78 & [7.88, 15.71] & <0.001 \\
\bottomrule
\end{tabular}
\caption{\small Family-level comparison between Full context and Agent/RAG by temporal expression tags. Scores are family-level means of binary LLM-as-judge correctness; scores and differences are reported on a 0--100 accuracy-point scale, with $\Delta = \text{Full context} - \text{Agent/RAG}$. Agent/RAG aggregates exclude \texttt{mem0g}.}
\label{tab:app_temporal_expression_fc_vs_agent}
\end{table*}

Table~\ref{tab:app_temporal_expression} shows that implicit temporal expressions are much harder than explicit temporal expressions. This is one of the strongest slice-level findings. For full-context systems, the explicit-to-implicit gap is approximately $30.02$ accuracy points in Russian and $29.89$ accuracy points in English. Thus, simply providing the full dialogue does not eliminate the difficulty of implicit temporal grounding.

Questions with no explicit temporal expression occupy an intermediate regime: they are significantly easier than questions with implicit temporal evidence, but for full-context systems they are significantly harder than  questions with explicit temporal evidence.

\begin{table*}[t]
\centering
\scriptsize
\begin{tabular}{llrrrrr}
\toprule
\textbf{Contrast} & \textbf{Family / Lang.} & \textbf{A} & \textbf{B} & \textbf{$\Delta$} & \textbf{95\% CI} & \textbf{Holm $p$} \\
\midrule
Explicit $-$ Implicit & Agent/RAG RU & 63.33 & 44.35 & +18.99 & [9.68, 28.61] & <0.001 \\
Explicit $-$ Implicit & Agent/RAG EN & 60.20 & 38.55 & +21.65 & [11.82, 31.34] & <0.001 \\
Explicit $-$ Implicit & Full context RU & 76.19 & 46.17 & +30.02 & [21.90, 37.77] & <0.001 \\
Explicit $-$ Implicit & Full context EN & 76.89 & 47.00 & +29.89 & [21.61, 38.35] & <0.001 \\
\midrule
No expression $-$ Explicit & Agent/RAG RU & 60.33 & 63.33 & -3.01 & [-9.69, 3.82] & 0.387 \\
No expression $-$ Explicit & Agent/RAG EN & 53.85 & 60.20 & -6.34 & [-13.05, 0.61] & 0.153 \\
No expression $-$ Explicit & Full context RU & 62.18 & 76.19 & -14.01 & [-19.94, -7.71] & <0.001 \\
No expression $-$ Explicit & Full context EN & 65.63 & 76.89 & -11.26 & [-17.02, -5.20] & <0.001 \\
\midrule
No expression $-$ Implicit & Agent/RAG RU & 60.33 & 44.35 & +15.98 & [7.33, 24.95] & <0.001 \\
No expression $-$ Implicit & Agent/RAG EN & 53.85 & 38.55 & +15.30 & [6.73, 24.30] & <0.001 \\
No expression $-$ Implicit & Full context RU & 62.18 & 46.17 & +16.01 & [8.75, 23.20] & <0.001 \\
No expression $-$ Implicit & Full context EN & 65.63 & 47.00 & +18.63 & [11.05, 26.59] & <0.001 \\
\bottomrule
\end{tabular}
\caption{\small Temporal-expression explicitness analysis within method families. Scores are family-level means of binary LLM-as-judge correctness; scores and differences are reported on a 0--100 accuracy-point scale, with $\Delta$ computed as the first tag minus the second tag in each contrast. Agent/RAG aggregates exclude \texttt{mem0g}.}\label{tab:app_temporal_expression}
\end{table*}

\subsection{Language Effects}

Table~\ref{tab:app_language_effects} reports English--Russian differences. \textit{These comparisons should be interpreted cautiously because Russian and English runs use language-specific judges.} At the family level, language effects are statistically detectable but small: Agent/RAG systems perform slightly better in Russian overall, whereas full-context systems perform slightly better in English overall. The clearest language-conditioned difference for Agent/RAG appears on temporal questions and implicit/no-expression temporal tags, where Russian scores are higher. For full-context systems, the largest language effect appears in the Reasoning semantic supergroup.

\begin{table*}[t]
\centering
\scriptsize
\begin{tabular}{llrrrrr}
\toprule
\textbf{Slice} & \textbf{Family} & \textbf{English} & \textbf{Russian} & \textbf{EN--RU} & \textbf{95\% CI} & \textbf{Holm $p$} \\
\midrule
Overall & Agent/RAG & 54.29 & 55.80 & -1.51 & [-2.78, -0.24] & 0.019 \\
Overall & Full context & 69.40 & 67.66 & +1.74 & [0.72, 2.74] & 0.001 \\
\midrule
Single & Agent/RAG & 62.92 & 63.92 & -0.99 & [-2.59, 0.57] & 0.217 \\
Single & Full context & 78.21 & 76.65 & +1.57 & [0.27, 2.85] & 0.042 \\
Multi & Agent/RAG & 40.87 & 43.18 & -2.32 & [-4.42, -0.19] & 0.040 \\
Multi & Full context & 55.70 & 53.69 & +2.01 & [0.35, 3.62] & 0.040 \\
\midrule
Atemporal & Agent/RAG & 60.02 & 60.55 & -0.53 & [-2.16, 1.06] & 0.524 \\
Atemporal & Full context & 70.91 & 69.39 & +1.52 & [0.35, 2.66] & 0.025 \\
Temporal & Agent/RAG & 52.87 & 58.41 & -5.54 & [-8.05, -3.08] & <0.001 \\
Temporal & Full context & 65.30 & 62.96 & +2.34 & [0.31, 4.41] & 0.023 \\
\midrule
Implicit time & Agent/RAG & 38.55 & 44.35 & -5.80 & [-10.72, -0.87] & 0.049 \\
Implicit time & Full context & 47.00 & 46.17 & +0.83 & [-3.73, 5.38] & 0.762 \\
No temporal expression & Agent/RAG & 53.85 & 60.33 & -6.48 & [-9.75, -3.20] & <0.001 \\
No temporal expression & Full context & 65.63 & 62.18 & +3.45 & [0.88, 6.09] & 0.010 \\
\midrule
Extraction & Agent/RAG & 53.82 & 56.12 & -2.30 & [-3.83, -0.77] & 0.005 \\
Extraction & Full context & 69.85 & 69.20 & +0.65 & [-0.62, 1.89] & 0.325 \\
Reasoning & Agent/RAG & 41.67 & 41.88 & -0.22 & [-3.12, 2.63] & 0.884 \\
Reasoning & Full context & 60.51 & 55.02 & +5.48 & [3.18, 7.83] & <0.001 \\
Abstention & Agent/RAG & 83.01 & 81.60 & +1.41 & [-1.62, 4.44] & 0.382 \\
Abstention & Full context & 84.14 & 82.28 & +1.86 & [-0.56, 4.36] & 0.294 \\
\bottomrule
\end{tabular}
\caption{\small Language comparison. Scores are family-level means of binary LLM-as-judge correctness; scores and differences are reported on a 0--100 accuracy-point scale, with $\Delta = \text{English} - \text{Russian}$. Agent/RAG temporal and temporal-expression aggregates exclude \texttt{mem0g}. Note that Russian and English evaluations use different LLM judges.}\label{tab:app_language_effects}
\end{table*}

\subsection{Semantic Supergroups}

We collapse the first category axis into three semantic "supergroups": Extraction, Reasoning, and Abstention (Table \ref{tab:taxonomy_summary}). 

Table~\ref{tab:app_semantic_supergroups} shows that Abstention is comparatively saturated: Agent/RAG systems and full-context systems perform similarly (in a setting without a matched answer model), and the gap is not statistically significant in either language. In contrast, full-context systems significantly outperform Agent/RAG systems on both Reasoning and Extraction. Reasoning is the hardest supergroup for both families.

\begin{table*}[t]
\centering
\small
\begin{tabular}{llrrrrr}
\toprule
\textbf{Semantic supergroup} & \textbf{Lang.} & \textbf{Full context} & \textbf{Agent/RAG} & \textbf{$\Delta$} & \textbf{95\% CI} & \textbf{Holm $p$} \\
\midrule
Abstention & RU & 82.28 & 81.60 & +0.68 & [-2.69, 4.13] & 0.694 \\
Abstention & EN & 84.14 & 83.01 & +1.13 & [-1.73, 4.00] & 0.455 \\
\midrule
Reasoning & RU & 55.02 & 41.88 & +13.14 & [10.21, 16.12] & <0.001 \\
Reasoning & EN & 60.51 & 41.67 & +18.84 & [15.71, 22.07] & <0.001 \\
\midrule
Extraction & RU & 69.20 & 56.12 & +13.09 & [11.45, 14.70] & <0.001 \\
Extraction & EN & 69.85 & 53.82 & +16.04 & [14.43, 17.60] & <0.001 \\
\bottomrule
\end{tabular}
\caption{\small Family-level comparison between Full context and Agent/RAG by semantic supergroup from Figure~\ref{fig:taxonomy}. Scores are family-level means of binary LLM-as-judge correctness; scores and differences are reported on a 0--100 accuracy-point scale, with $\Delta = \text{Full context} - \text{Agent/RAG}$.}\label{tab:app_semantic_supergroups}
\end{table*}

Table~\ref{tab:app_semantic_supergroup_contrasts} reports the within-family semantic-supergroup difficulty contrasts. Extraction is consistently easier than Reasoning. Abstention is easier than both Extraction and Reasoning, especially relative to Reasoning. These contrasts show that the semantic grouping captures a robust difficulty gradient rather than only a difference between method families.

\begin{table*}[t]
\centering
\scriptsize
\begin{tabular}{llrrrrr}
\toprule
\textbf{Contrast} & \textbf{Family / Lang.} & \textbf{A} & \textbf{B} & \textbf{$\Delta$} & \textbf{95\% CI} & \textbf{Holm $p$} \\
\midrule
Extraction $-$ Reasoning & Agent/RAG RU & 56.12 & 41.88 & +14.24 & [10.26, 18.11] & <0.001 \\
Extraction $-$ Reasoning & Agent/RAG EN & 53.82 & 41.67 & +12.15 & [8.32, 16.02] & <0.001 \\
Extraction $-$ Reasoning & Full context RU & 69.20 & 55.02 & +14.18 & [10.22, 18.19] & <0.001 \\
Extraction $-$ Reasoning & Full context EN & 69.85 & 60.51 & +9.35 & [5.48, 13.33] & <0.001 \\
\midrule
Extraction $-$ Abstention & Agent/RAG RU & 56.12 & 81.60 & -25.48 & [-29.65, -21.31] & <0.001 \\
Extraction $-$ Abstention & Agent/RAG EN & 53.82 & 83.01 & -29.19 & [-33.52, -24.73] & <0.001 \\
Extraction $-$ Abstention & Full context RU & 69.20 & 82.28 & -13.08 & [-17.17, -9.01] & <0.001 \\
Extraction $-$ Abstention & Full context EN & 69.85 & 84.14 & -14.28 & [-18.47, -9.95] & <0.001 \\
\midrule
Reasoning $-$ Abstention & Agent/RAG RU & 41.88 & 81.60 & -39.72 & [-44.85, -34.48] & <0.001 \\
Reasoning $-$ Abstention & Agent/RAG EN & 41.67 & 83.01 & -41.34 & [-46.58, -36.15] & <0.001 \\
Reasoning $-$ Abstention & Full context RU & 55.02 & 82.28 & -27.26 & [-32.25, -22.08] & <0.001 \\
Reasoning $-$ Abstention & Full context EN & 60.51 & 84.14 & -23.63 & [-28.71, -18.46] & <0.001 \\
\bottomrule
\end{tabular}
\caption{\small Within-family difficulty contrasts between semantic supergroups from Figure~\ref{fig:taxonomy}. Scores are family-level means of binary LLM-as-judge correctness; scores and differences are reported on a 0--100 accuracy-point scale, with $\Delta$ computed as the first supergroup minus the second supergroup in each contrast.}\label{tab:app_semantic_supergroup_contrasts}
\end{table*}

\subsection{Reasoning vs Temporal Reasoning}

As a more focused reasoning analysis, we compare non-temporal Reasoning categories against Temporal Reasoning categories from Fig.\ref{fig:taxonomy}. Table~\ref{tab:app_reasoning_vs_temporal_reasoning_contrast} reports the planned contrasts.

The observed differences are small and not statistically significant after Holm correction. Moreover, the direction of the difference is not consistent across languages and method families: Temporal Reasoning is slightly higher for Agent/RAG in Russian, whereas general Reasoning is slightly higher in the other three settings. 

Although we do not observe a significant aggregate effect in this comparison, the two groups differ conceptually and may still reveal model-specific patterns for individual model/methods.

\begin{table*}[t]
\centering
\small
\begin{tabular}{llrrrrr}
\toprule
\textbf{Contrast} & \textbf{Family / Lang.} & \textbf{A} & \textbf{B} & \textbf{$\Delta$} & \textbf{95\% CI} & \textbf{Holm $p$} \\
\midrule
Reasoning $-$ TR & Agent/RAG RU & 39.49 & 44.10 & -4.61 & [-11.54, 2.35] & 0.792 \\
Reasoning $-$ TR & Agent/RAG EN & 42.62 & 40.79 & +1.83 & [-4.86, 8.68] & 1.000 \\
Reasoning $-$ TR & Full context RU & 55.99 & 54.13 & +1.87 & [-5.16, 8.95] & 1.000 \\
Reasoning $-$ TR & Full context EN & 62.42 & 58.74 & +3.68 & [-2.94, 10.18] & 0.868 \\
\bottomrule
\end{tabular}
\caption{\small General Reasoning versus Temporal Reasoning (TR) categories defined in Figure~\ref{fig:taxonomy}. Scores are family-level means of binary LLM-as-judge correctness; scores and differences are reported on a 0--100 accuracy-point scale, with $\Delta$ computed as General Reasoning minus Temporal Reasoning.}
\label{tab:app_reasoning_vs_temporal_reasoning_contrast}
\end{table*}

\subsection{Semantic-Type Heatmaps}

\begin{figure*}[t]
\centering
\includegraphics[width=\textwidth]{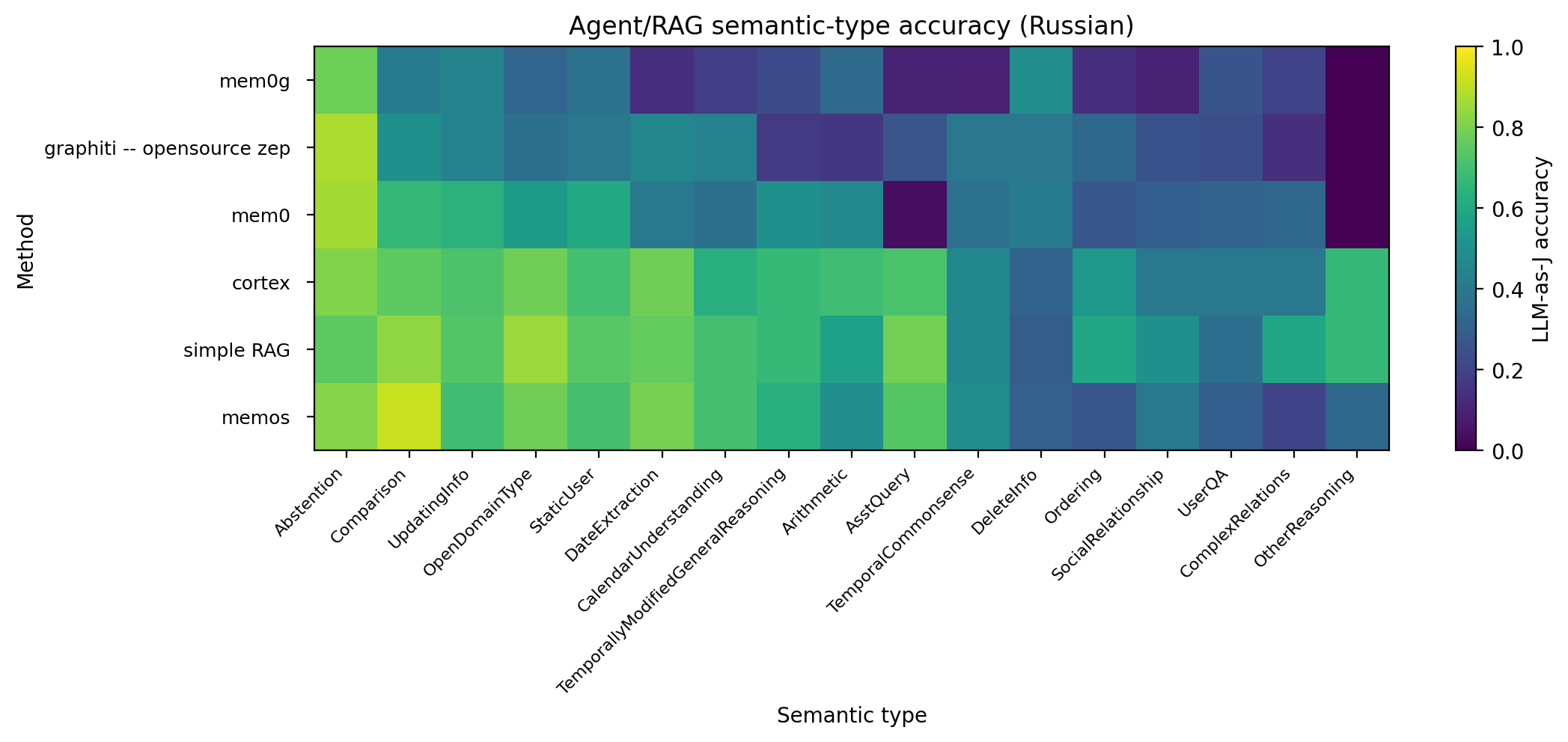}

\vspace{0.6em}
\vspace{1.0em}

\includegraphics[width=\textwidth]{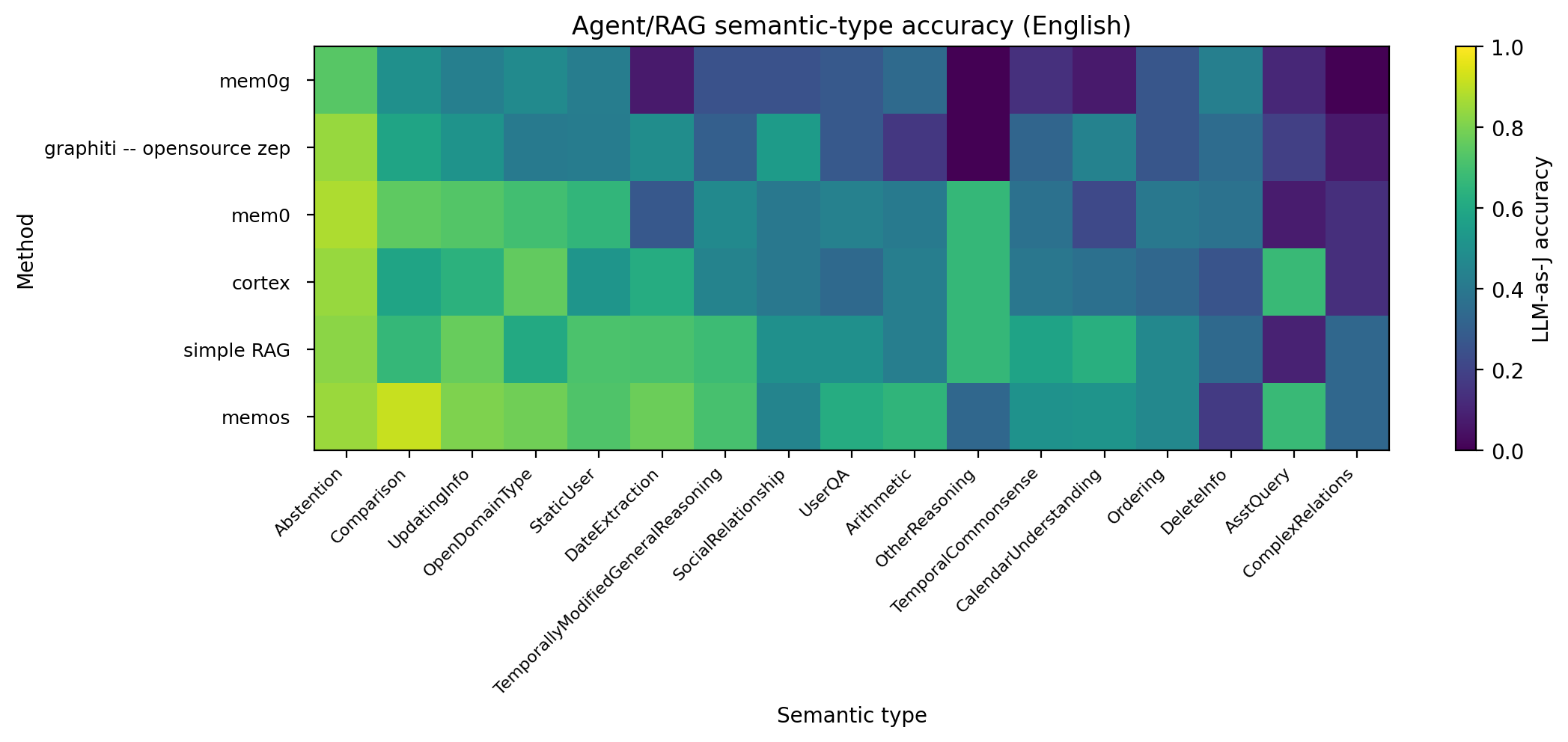}

\vspace{0.6em}

\caption{Agent/RAG heatmaps by semantic type.}
\label{fig:app_heatmap_agent_rag}
\end{figure*}

\begin{figure*}[t]
\centering
\includegraphics[width=\textwidth]{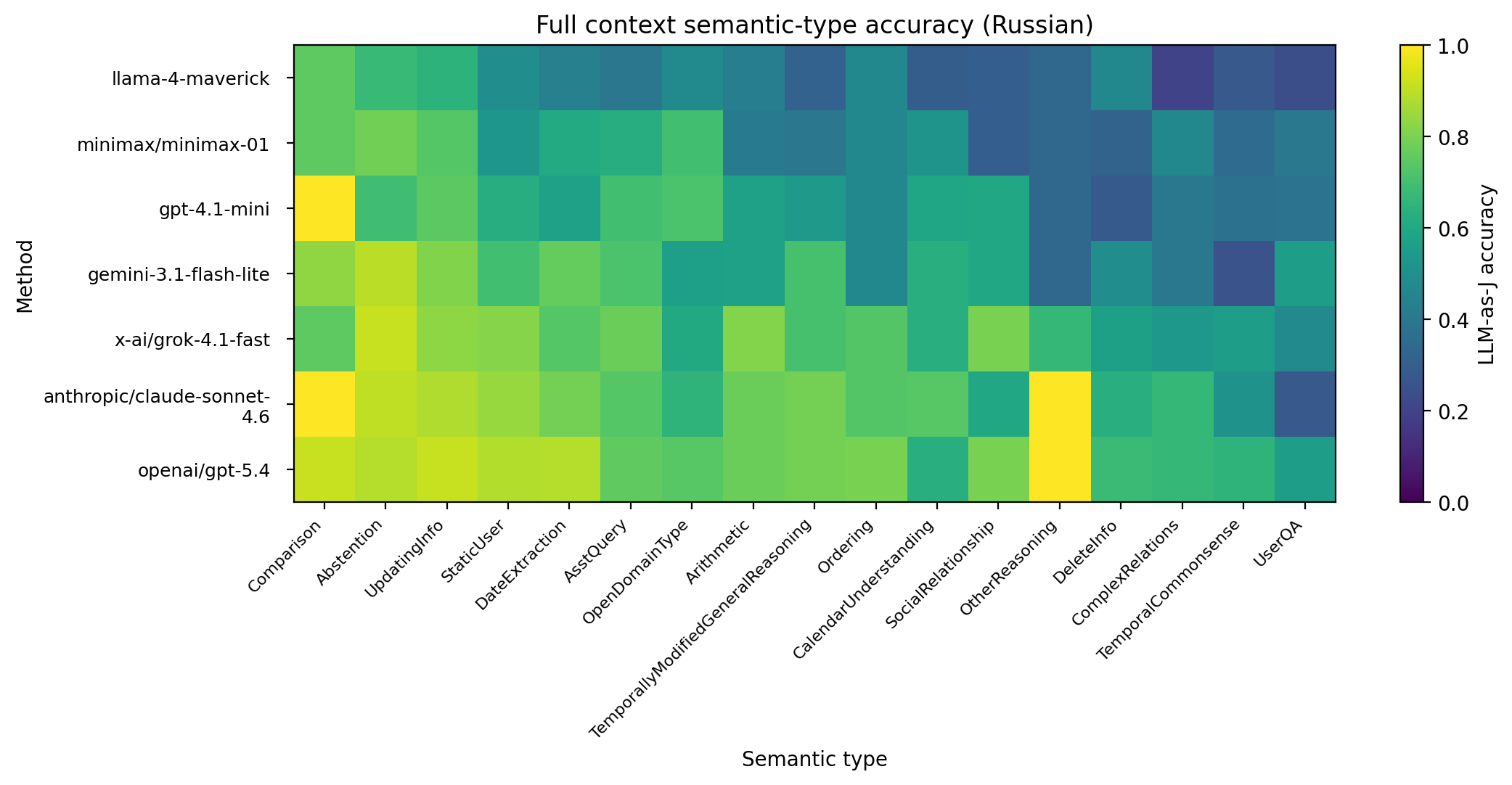}

\vspace{0.6em}
\vspace{1.0em}

\includegraphics[width=\textwidth]{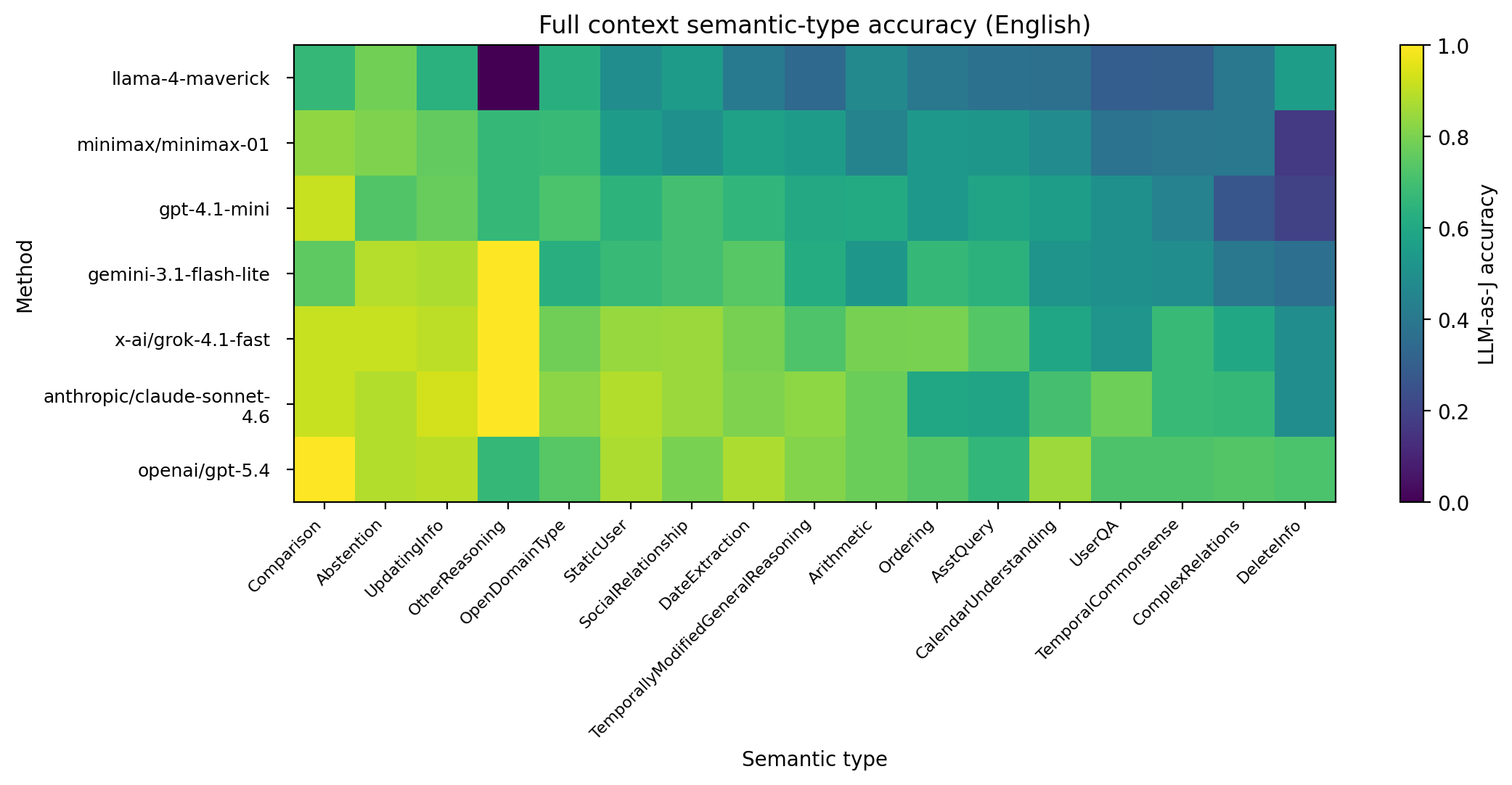}

\vspace{0.6em}

\caption{Full-context heatmaps by semantic type.}
\label{fig:app_heatmap_full_context}
\end{figure*}

Figures~\ref{fig:app_heatmap_agent_rag} and~\ref{fig:app_heatmap_full_context} show method-level heatmaps by semantic type. These figures are intended as descriptive visualizations rather than statistical comparisons, since semantic categories vary in sample size.

\subsection{Lost-in-the-Middle Proxy}

We also test whether performance drops when the evidence is located in the middle of the dialogue. Since exact token positions in the final prompt are not used here, this is a session-level proxy (thus, it cannot adjudicate token-level lost-in-the-middle effects): questions with evidence near the beginning or end are grouped as edge, and compared against questions with evidence in the middle.

Table~\ref{tab:app_lost_middle} shows no statistically significant edge-versus-middle effect for either method family or language. Thus, we do not detect a lost-in-the-middle pattern under this proxy.

\begin{table*}[t]
\centering
\small
\begin{tabular}{llrrrrr}
\toprule
\textbf{Family} & \textbf{Lang.} & \textbf{Edge} & \textbf{Middle} & \textbf{Edge--Middle} & \textbf{95\% CI} & \textbf{Holm $p$} \\
\midrule
Agent/RAG & RU & 55.62 & 56.13 & -0.51 & [-3.68, 2.66] & 0.986 \\
Agent/RAG & EN & 55.37 & 52.36 & +3.01 & [-0.20, 6.26] & 0.270 \\
Full context & RU & 68.08 & 66.90 & +1.18 & [-1.96, 4.30] & 0.986 \\
Full context & EN & 69.97 & 68.37 & +1.60 & [-1.60, 4.83] & 0.986 \\
\bottomrule
\end{tabular}
\caption{\small Lost-in-the-middle proxy analysis by evidence position in the dialogue. Scores are family-level means of binary LLM-as-judge correctness; scores and differences are reported on a 0--100 accuracy-point scale, with $\Delta$ computed as Edge minus Middle. Edge combines questions whose evidence is located near the beginning or end of the dialogue.}
\label{tab:app_lost_middle}
\end{table*}

Because evidence-position effects can be confounded by multi-session questions, we also repeat the same edge-versus-middle analysis on single-session questions only. Table~\ref{tab:app_lost_middle_single_only} shows the same conclusion: no statistically significant lost-in-the-middle effect is detected after restricting to single-session questions.

\begin{table*}[t]
\centering
\small
\begin{tabular}{llrrrrr}
\toprule
\textbf{Family} & \textbf{Lang.} & \textbf{Edge} & \textbf{Middle} & \textbf{Edge--Middle} & \textbf{95\% CI} & \textbf{Holm $p$} \\
\midrule
Agent/RAG & RU & 63.87 & 64.01 & -0.14 & [-4.04, 3.64] & 1.000 \\
Agent/RAG & EN & 64.11 & 60.41 & +3.70 & [-0.28, 7.72] & 0.289 \\
Full context & RU & 76.11 & 77.79 & -1.68 & [-5.13, 1.90] & 1.000 \\
Full context & EN & 77.83 & 79.02 & -1.19 & [-4.73, 2.47] & 1.000 \\
\bottomrule
\end{tabular}
\caption{\small Lost-in-the-middle proxy analysis restricted to single-session questions. Scores are family-level means of binary LLM-as-judge correctness; scores and differences are reported on a 0--100 accuracy-point scale, with $\Delta$ computed as Edge minus Middle.}
\label{tab:app_lost_middle_single_only}
\end{table*}

\subsection{Best-Observed Methods}

For practical interpretation, we also report the best observed method in each family. This is an upper-bound comparison and is not used as the main statistical evidence for family-level claims.

Across most full-context slices, the strongest observed model is GPT-5.4. For Agent/RAG systems, the strongest Russian results are typically obtained by simple RAG, cortex, or memOS depending on the slice; in English, memOS is the strongest Agent/RAG method across most major slices. These comparisons characterize the strongest observed upper bounds, not a controlled effect of the memory-access mechanism; the matched-answerer results in Table~\ref{tab:main_family_results_compact} show that individual Agent/RAG systems can exceed the \texttt{gpt-4.1-mini} full-context baseline.

\section{Case study: Claude Sonnet 4.6}
\label{app:case_study}

In this section, we demonstrate how RUMBA can be used to profile a single selected model rather than for benchmarking different methods.
For this case study, we choose \texttt{anthropic/claude-sonnet-4.6}, the second-best system in the full-context setting in both language versions of our evaluation, reaching 78.61 accuracy points in Russian and 81.98 accuracy points in English. 
Using the same diagnostic axes and statistical procedure as in the preceding appendix analyses, we examine where this model is robust, where it degrades, and which benchmark dimensions explain its main failure modes.

Table~\ref{tab:case_study_claude_profile} reports descriptive slice-level scores. 
Claude Sonnet 4.6 is strongest on Abstention and Extraction-oriented questions, while its weaker regions are multi-session questions, implicit temporal expressions, and Reasoning-oriented questions. 
This profile is broadly consistent across languages, although English is generally stronger than Russian.

\begin{table*}[t]
\centering
\scriptsize
\setlength{\tabcolsep}{4pt}
\begin{tabular}{llrrr}
\toprule
\textbf{Axis} & \textbf{Slice} & \textbf{\(n\)} & \textbf{Russian} & \textbf{English} \\
\midrule
Overall & All questions & 1543 & 78.61 & 81.98 \\
\midrule
Scope & Single-session & 939 & 83.49 & 88.60 \\
      & Multi-session & 604 & 71.03 & 71.69 \\
\midrule
Temporal grounding & Atemporal & 1128 & 79.79 & 83.24 \\
                   & Temporal & 415 & 75.42 & 78.55 \\
\midrule
Scope $\times$ time & Single + atemporal & 678 & 84.66 & 89.68 \\
                    & Multi + atemporal & 450 & 72.44 & 73.56 \\
                    & Single + temporal & 261 & 80.46 & 85.82 \\
                    & Multi + temporal & 154 & 66.88 & 66.23 \\
\midrule
Temporal expression & Explicit & 102 & 85.29 & 84.31 \\
                    & Implicit & 69 & 57.97 & 60.87 \\
                    & No temporal expression & 244 & 76.23 & 81.15 \\
\midrule
Semantic supergroup & Extraction & 1079 & 80.82 & 82.58 \\
                    & Reasoning & 310 & 65.16 & 76.77 \\
                    & Abstention & 154 & 90.26 & 88.31 \\
\midrule
Reasoning supergroup & General Reasoning & 149 & 60.40 & 78.52 \\
                  & Temporal Reasoning & 161 & 69.57 & 75.16 \\
\bottomrule
\end{tabular}
\caption{\small Single-model diagnostic profile for Claude Sonnet 4.6 across benchmark axes and slices. Values are LLM-as-judge accuracy points on a 0--100 scale; \(n\) denotes the number of questions in each slice.}
\label{tab:case_study_claude_profile}
\end{table*}

Table~\ref{tab:case_study_claude_significant} summarizes the statistically significant contrasts. 
All deltas are reported in accuracy points on a 0--100 scale. 
We omit non-significant contrasts from the table to keep the case study focused on the model's main diagnostic failure modes.

\begin{table*}[t]
\centering
\scriptsize
\setlength{\tabcolsep}{3.5pt}
\begin{tabular}{llcrcc}
\toprule
\textbf{Block} & \textbf{Contrast} & \textbf{Lang.} & \textbf{\(\Delta\)} & \textbf{95\% CI} & \textbf{Holm \(p\)} \\
\midrule
Scope 
& Single $-$ multi 
& RU & 12.47 & [8.21, 16.77] & $<0.001$ \\
& Single $-$ multi 
& EN & 16.92 & [12.82, 21.02] & $<0.001$ \\
\midrule
Scope $\times$ time
& Single+atemporal $-$ multi+atemporal 
& RU & 12.22 & [7.40, 17.02] & $<0.001$ \\
& Single+atemporal $-$ multi+atemporal 
& EN & 16.12 & [11.47, 20.71] & $<0.001$ \\
& Single+temporal $-$ multi+temporal 
& RU & 13.58 & [4.93, 22.39] & 0.003 \\
& Single+temporal $-$ multi+temporal 
& EN & 19.59 & [11.27, 28.12] & $<0.001$ \\
\midrule
Temporal expression
& Explicit $-$ implicit 
& RU & 27.32 & [13.68, 40.79] & $<0.001$ \\
& Explicit $-$ implicit 
& EN & 23.44 & [9.85, 36.62] & 0.001 \\
& No expression $-$ implicit 
& RU & 18.26 & [5.43, 31.30] & 0.005 \\
& No expression $-$ implicit 
& EN & 20.28 & [7.67, 32.91] & 0.003 \\
\midrule
Semantic supergroup
& Extraction $-$ Reasoning 
& RU & 15.65 & [9.84, 21.62] & $<0.001$ \\
& Extraction $-$ Reasoning 
& EN & 5.80 & [0.55, 11.06] & 0.030 \\
& Extraction $-$ Abstention 
& RU & -9.44 & [-14.55, -3.97] & 0.003 \\
& Reasoning $-$ Abstention 
& RU & -25.10 & [-32.20, -17.66] & $<0.001$ \\
& Reasoning $-$ Abstention 
& EN & -11.54 & [-18.33, -4.42] & 0.002 \\
\midrule
English $-$ Russian
& Overall 
& -- & 3.37 & [1.23, 5.44] & $<0.001$ \\
& Single-session 
& -- & 5.11 & [2.77, 7.56] & $<0.001$ \\
& Atemporal 
& -- & 3.46 & [1.15, 5.76] & 0.006 \\
& Single+atemporal 
& -- & 5.01 & [2.21, 7.82] & $<0.001$ \\
& Single+temporal 
& -- & 5.36 & [0.38, 10.34] & 0.049 \\
& Reasoning supergroup 
& -- & 11.61 & [6.77, 16.45] & $<0.001$ \\
& General Reasoning 
& -- & 18.12 & [10.07, 25.50] & $<0.001$ \\
\bottomrule
\end{tabular}
\caption{\small Significant single-model diagnostic contrasts for Claude Sonnet 4.6 across benchmark axes. Differences are reported as LLM-as-judge accuracy points on a 0--100 scale; \(\Delta\) is computed as the first slice minus the second slice in each contrast, except for the language block where \(\Delta = \text{English} - \text{Russian}\).}\label{tab:case_study_claude_significant}
\end{table*}

The clearest degradation is caused by session scope. 
As shown in Table~\ref{tab:case_study_claude_profile}, accuracy drops from 83.49 to 71.03 in Russian and from 88.60 to 71.69 in English when moving from single-session to multi-session questions. 
The corresponding contrasts in Table~\ref{tab:case_study_claude_significant} are significant in both languages, with gaps of 12.47 and 16.92 accuracy points. 
This degradation also remains significant inside both atemporal and temporal subsets. 
The hardest scope--temporal intersection is multi-session temporal questioning, where the model reaches 66.88 in Russian and 66.23 in English.
This suggests that the main structural weakness is not temporal grounding alone, but the combination of temporal grounding with evidence distributed across multiple sessions.

Temporal questions as a broad category are not the strongest explanation of the model's failures. 
The atemporal--temporal contrast is not significant after Holm correction in either language. 
However, the temporal-expression analysis reveals a more specific failure mode. 
Claude handles explicit temporal expressions well, with accuracy of 85.29 in Russian and 84.31 in English, but drops sharply on implicit temporal expressions, reaching only 57.97 and 60.87. 
The explicit--implicit gaps are large and significant in both languages: 27.32 accuracy points in Russian and 23.44 in English. 
Implicit temporal expressions are also significantly harder than questions with no temporal expression.

The semantic profile shows another major weakness. 
Claude is strong on Abstention, especially in Russian, where it reaches 90.26 accuracy points, and it is also relatively strong on Extraction. 
Reasoning is substantially weaker: 65.16 in Russian and 76.77 in English. 
The Extraction--Reasoning contrast is significant in both languages, while Reasoning--Abstention is even larger, especially in Russian, where the gap is -25.10 accuracy points. 
This indicates that the model is reliable when it must retrieve or reject unknown user information, but less reliable when the answer requires combining, comparing, or transforming it.

Temporal Reasoning is not the uniquely problematic group for Claude Sonnet 4.6. In Russian, Temporal Reasoning is numerically higher than general Reasoning, 69.57 versus 60.40 accuracy points. 
In English, the direction reverses: General Reasoning reaches 78.52, while Temporal Reasoning reaches 75.16. 
The within-language Reasoning--Temporal Reasoning contrast is not significant after correction in either language. 
However, the English--Russian gap is significant for general Reasoning, with English outperforming Russian by 18.12 accuracy points, while the corresponding language gap for Temporal Reasoning is not significant. 
This suggests that Claude's Russian weakness is concentrated more in general memory reasoning operations than in Temporal Reasoning specifically.

The language comparison also reveals a targeted rather than uniform gap. 
English is significantly better overall by 3.37 accuracy points, but the largest language differences occur in Reasoning-heavy slices. 
The English--Russian gap is 11.61 accuracy points for the Reasoning supergroup and 18.12 accuracy points for General Reasoning. 
By contrast, multi-session questions, temporal questions as a whole, implicit temporal expressions, Abstention, and Temporal Reasoning do not show significant English--Russian differences. 
Therefore, Claude's Russian weakness is concentrated primarily in General Reasoning rather than in memory retrieval or temporal grounding overall.

Finally, the evidence-position analysis does not support a clear lost-in-the-middle explanation for this model. 
The edge--middle contrast is not significant in either language, including when restricting the analysis to single-session questions. 
For all questions, edge--middle differences are -0.30 accuracy points in Russian and 3.54 in English; for single-session questions only, they are -2.78 and 0.83 accuracy points, respectively. 
Consequently, the dominant failure modes are better explained by multi-session integration, implicit temporal inference, and Reasoning operations than by the absolute position of the supporting evidence in the dialogue.                                                                                                                                                                                                                                                                                                                             \onecolumn
\begingroup
\small
\renewcommand{\arraystretch}{1.5}
\begin{longtable}{p{2cm}p{7cm}p{7cm}}%
\caption{Question taxonomy: detailed description of question types and subtypes according to the semantic axis, with illustrative examples.}
\label{tab:full_description}\\
\hline
\textbf{Category\&Type} & \textbf{Type Description} & \textbf{Example} \\
\endfirsthead
\textbf{Category\&Type} & \textbf{Type Description} & \textbf{Example} \\
\midrule
\endhead
\multicolumn{3}{r}{Continuation} \\
\endfoot
\endlastfoot
\midrule
\rowcolor{gray!20}
\multicolumn{3}{p{16cm}}{NON-TEMPORAL} \\
\midrule
\rowcolor{gray!10}
\multicolumn{3}{p{16cm}}{INFORMATION EXTRACTION} \\
\midrule
\textbf{StaticUser} & \textbf{(a)} The fact to which the question refers remains unchanged across all sessions, both qualitatively and quantitatively. 
\par \textbf{(b)} The facts to which the question refers are complementary and do not constitute a qualitative or quantitative change of the same fact across sessions. & \textbf{User:} Our dog’s name is Losyash. 
\par \textbf{Question:} What’s my dog’s name? 
\par \textbf{Answer:} Losyash. \\
\textbf{UpdatingInfo} & The fact to which the question refers undergoes qualitative or quantitative changes within one or more sessions. & \textbf{Condition:} Pick the most up-to-date answer within one or across two or more sessions. 
\par \textbf{User:} I love roses.
\par [...]
\par \textbf{User:} I really love cacti now, I’m tired of roses.
\par [...]
\par \textbf{User:} I don’t like cacti anymore, lilies are my favorite now.
\par \textbf{Question:} What are my favorite plants?
\par \textbf{Answer:} Lilies. \\
\textbf{DeleteInfo} & The fact to which the question refers is deleted by the user within one or more sessions. & \textbf{Condition:} Pick the most up-to-date answer within one or across two or more sessions.
\par \textbf{AnswerStandard:} The correct answer is “No such information” (or similar).
\par \textbf{User:} I don’t want to be a poor peasant woman, I want to be a noble lady.
...
\par \textbf{User:} Delete the information about what I want to be.
\par \textbf{Question:} What do I want to be?
\par \textbf{Answer:} No such information. \\
\textbf{AsstQuery} & The fact to which the question refers is contained in the assistant’s response. In this case, the response constitutes advice, a recommendation, or an answer to an instructional request. 
\par Important: the fact does not concern the assistant’s persona and does not disclose attributes such as age, gender, personality, name, etc. & \textbf{User:} Who should I read from the 20th century?
\par \textbf{Assistant:} [List]... Bulgakov, Pasternak, Dovlatov, Orwell, Joyce...
\par \textbf{Question:} Which Russian 20th-century writers did you recommend?
\par \textbf{Answer:} Bulgakov, Pasternak, Dovlatov. \\
\textbf{OpenDomain Type} & The fact to which the question refers pertains to widely known entities (events—past or recent—objects, public figures, etc.) and is characterized by the fact that a correct or approximate answer can be obtained from publicly available sources without consulting the session history. & \textbf{User:} In Attack on Titan, my favorite pairing is Jean and Mikasa.
\par \textbf{Question:} What’s my favorite Attack on Titan pairing?
\par \textbf{Answer: }Jean and Mikasa. \\
\midrule
\rowcolor{gray!10}
\multicolumn{3}{p{16cm}}{REASONING} \\
\midrule
\textbf{Social}\par\textbf{Relationship} & The fact to which the question refers reveals the nature of the relationship between individuals. 
\par Important: the social role is not explicitly stated in the dialogue and must be inferred through reasoning. & \textbf{User:} I don’t have many relatives, my mom only has one sister, Lera.
\par \textbf{Question:} Who is Lera to me?
\par \textbf{Answer:} Your aunt. \\
\textbf{Ordering} & The entities to which the question refers share a common attribute. To answer the question, these entities (objects, features, etc.) must be arranged in a specific order according to a given parameter. & \textbf{User:} My sisters’ names are Ira, Vasilisa, and Alisa.
\par \textbf{Question:} List my sisters in alphabetical order.
\par \textbf{Answer:} Alisa, Ira, Vasilisa. \\
\textbf{Arithmetic} & The fact to which the question refers requires arithmetic computation:
\par (a) sum — calculation of the total;
\par (b) less / more — calculation of the difference;
\par (c) average — calculation of the mean value;
\par (d) other — other mathematical operations. & AnswerStandard: The answer should include a number.
\par \textbf{User:} I have a guitar, a domra, and my grandma’s piano at home. But I only play the guitar and domra.
\par \textbf{Question:} How many musical instruments do I have?
\par \textbf{Answer:} Three. \\
\textbf{Comparison} & The fact to which the question refers requires qualitative or quantitative comparison of two or more entities. The question reflects an understanding of comparative, superlative, and equivalence relations between entities:
\par (a) comparative — direct comparison (A is more X than B);
\par (b) superlative — identification of the most X within a set;
\par (c) equivalence — A is equal to B. & \textbf{User:} We have two cats: Musya (5 years old) and Kusya (5 months old).
\par \textbf{Question:} Which pet is older?
\par \textbf{Answer:} Musya. \\
\textbf{UserQA} & The fact to which the question refers requires an understanding not only of the information about the user contained in the dialogue but also of specific or common world knowledge. & \textbf{User:} I professionally study German folklore.
\par \textbf{Question:} Am I a Germanist?
\par \textbf{Answer:} Yes. \\
\rowcolor{magenta!5}
\multicolumn{3}{p{16cm}}
{\textbf{OtherReasoning}  Other categories of reasoning tasks.} \\
\textbf{Causal}\par \textbf{Reasoning} & The fact to which the question refers requires an understanding of causal relationships: why / due to what / what was the cause / what was the effect. 
\par Important: both the cause and the effect are present in the text across different sessions and must be correlated.
\par Condition: To answer a question about the user, it is necessary to integrate information from multiple sessions. &  \textbf{User:} I got into a car accident, the car is wrecked.
\par [...]
\par \textbf{User:} I’m in the hospital, good thing the airbag worked.
\par \textbf{Question:} Why was I in the hospital?
\par \textbf{Answer:} Because you got into a car accident. \\
\textbf{MultiStep} & The fact to which the question refers requires multi-step reasoning. & \textbf{User:} I bought my first houseplants — four violets.
\par [...]
\par \textbf{User: }One violet died, Lena gave me her ficus (she had two), and my grandma gave me one big aloe and one small one.
\par [...]
\par \textbf{User:} My grandma gave Lena two snake plants because she only had ficuses.
\par \textbf{Question:} How many more plants do I have than Lena?
\par \textbf{Answer:} By 3. \\
\midrule
\rowcolor{gray!20}
\multicolumn{3}{p{16cm}}{TEMPORAL} \\
\midrule
\rowcolor{gray!10}
\multicolumn{3}{p{16cm}}{INFORMATION EXTRACTION} \\
\midrule
\multicolumn{3}{p{16cm}}{\textbf{DateExtraction}  Types of questions involving the extraction of an event date and/or questions containing a specific date.} \\
\textbf{ExplicitTE} \par \textbf{Condition} \par \textbf{Question} & \textbf{(a)} The user’s utterance explicitly contains a date, month, year, or time interval related to an event. 
\par \textbf{(b)} The question either includes a specific date (day, month, year) and/or refers to such a date. The corresponding subsubtypes: \textit{date-question, month/year-question, span-question, date-answer, month/year-answer, span-answer} & [16.09.2025] \textbf{User:} I worked at a swimming pool on July 15.
\par [20.12.2025] \textbf{Question:} Where did I work on July 15?
\par [20.12.2025] \textbf{Answer:} At a swimming pool. \\
\textbf{NonTE} \par \textbf{Condition} \par \textbf{Question }& \textbf{(a)} The user’s utterance does not explicitly mention the time of the event and contains no temporal reference. 
\par \textbf{(b)} The question either includes a specific date (day, month, year) and/or refers to such a date. The corresponding subsubtypes: \textit{date-question, month/year-question, span-question, date-answer, month/year-answer, span-answer} & [01.03.2025] \textbf{User:} Come up with a rap nickname for me. 
\par [01.03.2025] \textbf{Assistant:} [List]... Vupsen...
\par [06.06.2025] \textbf{Question:} Do you remember what nickname you gave me in March?
\par \textbf{Answer:} Vupsen. \\
\midrule
\rowcolor{gray!10}
\multicolumn{3}{p{16cm}}{REASONING} \\
\midrule
\rowcolor{magenta!5}
\multicolumn{3}{p{16cm}}
{\textbf{CalendarUnderstanding} Types of questions in which it is necessary to correctly understand and interpret calendrical information: terminology and structure, as well as the units used in calendars.} \\
\textbf{ImplicitTE} \par \textbf{Condition} \par \textbf{Question} & \textbf{(a)} The user’s utterance contains an implicit temporal reference (\textit{yesterday, today, this month, recently, the other day,} etc.).
\par \textbf{(b)}  A question aimed at identifying and extracting the time of an event. The corresponding subsubtypes: \textit{date-question, month/year-question, span-question, date-answer, month/year-answer, span-answer} & [23.11.2025] \textbf{User:} Yesterday I had a singing lesson, today I have a domra lesson.
\par [24.11.2025] \textbf{Question:} What lesson did I have on November 23?
\par \textbf{Answer:} Domra. \\ 
\textbf{RelativeTime} \par \textbf{Understanding} & A question aimed at understanding relative time, expressed in relation to the moment of speaking (i.e., when the question is asked). & [23.11.2025] \textbf{User:} I’m baking a charlotte, but it fell apart.
\par [24.11.2025] \textbf{Question:} What did I bake yesterday?
\par \textbf{Answer}: Charlotte. \\
\textbf{Weekday} \par \textbf{Understanding} & A question concerning the correspondence between a date and a day of the week. & [24.11.2025] \textbf{User:} I baked a charlotte on November 20.
\par [25.11.2025] \textbf{Question:} What did I bake on Thursday?
\par \textbf{Answer:} Charlotte \\
\textbf{DateQA} & A question concerning calendar periods, production calendars, and the properties of the calendar as a system. & [23.11.2025] \textbf{User:} I baked a charlotte on November 13, and today I'm making muffins.
\par [05.12.2025] \textbf{Question:} What did I bake in the second half of November?
\par \textbf{Answer:} Muffins. \\
\rowcolor{magenta!5}
\multicolumn{3}{p{16cm}}
{\textbf{TemporalCommonsense} A class of questions that fundamentally requires an understanding of the temporal structure of the world. } \\
\textbf{(Event)Ordering} & \textbf{(a)} A question concerning the sequence of events: what occurred before/after/between, as well as ordinal position: first/second/last. 
\par \textbf{(b)} A question involving entities sharing a common attribute, based on an understanding of event timing and/or the time of questioning. To answer, these entities must be ordered according to a specified parameter. & [21.05.2025] \textbf{User:} I have graduation coming up, and my thesis defense already happened.
\par [...]
\par [11.09.2025] \textbf{User:} I started working as a waitress before my thesis defense.
\par [11.12.2025] \textbf{Question:} Did I start working as a waitress before graduation?
\par \textbf{Answer:} Yes. \\
\textbf{EventFrequency} & A question concerning event frequency: how often does an event occur? The minimum unit of measurement is one day. & [06.04.2025] \textbf{User:} Cucumber for breakfast.
\par [...]
\par [17.04.2025] \textbf{User:} Salad with cucumbers.
\par [...]
\par [25.04.2025] \textbf{User:} Smashed cucumbers for dinner.
\par [01.05.2025] \textbf{Question:} How often do I eat cucumbers?
\par \textbf{Answer:} 3 times a month. \\
\textbf{EventDuration} & A question concerning event duration: how long did X last? how long before did Y occur? how many days after did A take place? The minimum unit of measurement is one day. & [20.10.2025] \textbf{User:} I started painting the walls.
\par [...]
\par [25.10.2025] \textbf{User:} I got divorced.
\par [30.12.2025] \textbf{Question:} How many days before the divorce did I start painting?
\par \textbf{Answer:} 5 days. \\
\rowcolor{magenta!5}
\multicolumn{3}{p{16cm}}
{\textbf{TemporallyModifiedGeneralReasoning} Reasoning questions complicated by temporal constraints.} \\
\textbf{Arithmetic} & The fact to which the question refers requires arithmetic computation based on an understanding of event timing and/or the time of questioning. & [11.09.2025] \textbf{User:} I weigh 65 kg.
\par [11.10.2025] \textbf{Question:} How much do I weigh now if last month I weighed 5 kg less?
\par \textbf{Answer:} 70 kg. \\
\textbf{Comparison} & The fact to which the question refers requires qualitative or quantitative comparison based on an understanding of event timing and/or the time of questioning. & [11.09.2025] \textbf{User:} Today I leg pressed 145 kg, last week it was 148 and 150 kg.
\par [30.10.2025] \textbf{Question:} What was my best result that month?
\par \textbf{Answer:} 150 kg. \\
\textbf{UserQA }& The fact to which the question refers requires an understanding not only of dialogue-based information but also of general or domain-specific knowledge, in the context of event timing and/or the time of questioning. & [15.03.2025] \textbf{User:} It’s my birthday today.
\par [...]
\par [15.07.2025] \textbf{User:} My husband’s 40th birthday is tomorrow.
\par [25.10.2025] \textbf{Question:} What zodiac signs are me and my husband?
\par \textbf{Answer:} Pisces and Cancer. \\
\textbf{Causal} \par \textbf{Reasoning} & The fact to which the question refers requires an understanding of causal relationships in the context of event timing and/or the time of questioning. Condition: To answer a question about the user, it is necessary to synthesize information from multiple sessions. & [15.10.2025] \textbf{User:} I argued with my mom, I’m so mad, how do I leave home?
\par [...]
\par [22.10.2025] \textbf{User}: We argued again about homework, same as last week.
\par [22.12.2025] \textbf{Question:} Why did we argue on October 15?
\par \textbf{Answer:} Because of homework. \\
\textbf{MultiStep} & The fact to which the question refers requires multi-step reasoning in the context of event timing and/or the time of questioning. & [01.09.2025] \textbf{User:} I got back from vacation, I was there from the 15th.
\par [...]
\par [15.11.2025] \textbf{User:} I’ll have a short vacation next month (5 days), and a small one this month from the 1st to the 4th.
\par [30.11.2025] \textbf{Question:} How long was my second vacation?
\par \textbf{Answer:} 4 days. \\
\rowcolor{magenta!5}
\multicolumn{3}{p{16cm}}{ComplexRelations} \\
\textbf{EventEvent} \par \textbf{Time} \par \textbf{Understanding} & The fact to which the question refers is revealed only through another event and requires an understanding of event timing and/or the time of questioning. At least two interrelated events are required. & [15.11.2025] \textbf{User:} Vasya and Petya came over.
\par [...]
\par [17.11.2025] \textbf{User:} My birthday was the day before yesterday.
\par [30.11.2025]\textbf{ Question:} Who came over on my birthday?
\par \textbf{Answer:} Vasya and Petya. \\
\textbf{During} \par \textbf{Reasoning} & A question concerning the overlap of events. At least two events are required. & [20.10.2025] \textbf{User:} I started painting the walls.
\par [...]
\par [21.10.2025] \textbf{User:} I went for a walk with my nephew Vanya.
\par [...]
\par [22.10.2025] \textbf{User:} I’m hanging out with Vanya.
\par [...]
\par [25.10.2025] \textbf{User:} I finished painting the walls.
\par [...]
\par [30.10.2025] \textbf{User:} I’m with my nephew again, everything’s good.
\par [20.11.2025] \textbf{Question:} How many times did I see my nephew while painting the walls?
\par \textbf{Answer:} 2 times. \\
\textbf{Trends} & The fact to which the question refers changes over the course of the dialogue. The question is posed to determine how a particular event or fact evolved over time. & [20.01.2025] \textbf{User:} I smoke 15 cigarettes a day.
\par [...]
\par [21.02.2025] \textbf{User:} I smoke 10 cigarettes a day.
\par [...]
\par [22.03.2025] \textbf{User:} I smoke 5 cigarettes a day.
\par [...]
\par [25.04.2025] \textbf{User:} I smoke 1 cigarette a day.
\par [...]
\par [30.05.2025] \textbf{User: }I smoke 15 cigarettes a day.
\par [21.09.2025] \textbf{Question:} How did my daily smoking change?
\par \textbf{Answer:} First it decreased from 15 cigarettes to 1, then increased sharply back to 15 cigarettes. \\
\midrule
\rowcolor{gray!20}
\multicolumn{3}{p{16cm}}{OTHER TYPES} \\
\midrule
\textbf{Abstention} & The dialogue does not contain an answer to this question. This is a control (trick) question intended to prevent hallucinations. The answer to the question is: No such information and similar. & \textbf{User:} I’m 17 and I have a boyfriend.
\par \textbf{Question:} How old is my boyfriend?
\par\textbf{Answer:} There’s no information about that. \\
\bottomrule
\end{longtable}
\endgroup
\twocolumn
                                                                                                                                                                                                                                                                                                                                                                                                                                                                                                                                                                                                                                                                                                                                                                                                                                                                                                                                                                                                                                                                                                                                                                                                                                                                                                                                                                                                                                                                                                                                                                                                                                                                                                                                                                                                                                                                                                                                                                                                                                                                                                                                                                                                                                                                                                                                                                                                                                                                                                                                                                                                                                                                                                                                                                                                                                                                                                                                                                                                                                                                                                                                                                                                                                                                                                                                                                                                                                                                                                                                                                                                                                                                                                                                                                                                                                                                                                                                                                                                                                                                                                                                                                                                                                                                                                                                                                                                                                                                                                                                                                                                                                                                                                                                                                                                                                                                                                                                                                                                                                                                                                                                                                                                                                                                                                                                                                                                                                                                                                                                                                                                                                                                                                                                                                                                                                                                                                                                                                                                                                                                                                                                                                                                                                                                                                                                                                                                                                                                                                                                                                                                                                                                                                                                                                                                                                                                                                                                                                                                                                                                                                                                                                                                                                                                                                                                                                                                                                                                                                                                                                                                 
\end{document}